\documentclass[12pt]{article}

\usepackage{amsthm,amsmath,amssymb}
\usepackage{multirow}
\usepackage[pdftex]{graphicx}
\usepackage{graphics}
\usepackage{subfigure}
\usepackage{makecell}
\usepackage{booktabs}
\usepackage{array}
\usepackage{fullpage}
\usepackage{url}
\usepackage{bm}
\usepackage{smile}
\usepackage{wrapfig}
\usepackage{lipsum}
\usepackage{mathrsfs}
\usepackage{dsfont}
\usepackage{titling}
\usepackage{epstopdf}
\usepackage{multirow}
\usepackage{natbib}
\usepackage{multirow}
\usepackage{float}
\usepackage{rotating}
\usepackage{tabularx}
\usepackage{algorithmic}

\usepackage[utf8]{inputenc}
\usepackage[english]{babel}
\usepackage[dvipsnames]{xcolor}
\usepackage{geometry}
\usepackage{mathrsfs}
\usepackage{amssymb}
\usepackage{amsmath}
\usepackage{amsthm}
\usepackage{bbm}
\usepackage{multirow}
\usepackage{graphicx}
\usepackage[ruled,vlined]{algorithm2e}
\usepackage{dirtytalk}
\usepackage{caption}
\usepackage{sidecap}
\usepackage{subfigure}

\usepackage{natbib}
\usepackage{hyperref}
\hypersetup{colorlinks=true,
            linkcolor=blue,
            urlcolor=blue,
            citecolor=blue}

\newcommand{\HH}{\mathbb{H}}
\newcommand{\GG}{\mathbb{G}}

\newcommand{\MCOD}{\textsc{mcod}}
\newcommand{\COD}{\textsc{cod}}
\newcommand{\NAIVE}{\textsc{naive}}
\newcommand{\ONESTEPCOD}{\textsc{1-step cod}}
\newcommand{\TWOSTEPCOD}{\textsc{2-step cod}}

\makeatletter

\newcommand{\Rmnum}[1]{\uppercase\expandafter{\romannumeral #1}}
\makeatother

\newcommand{\cov}{\mathrm{Cov}}

\begin{document}

\def\spacingset#1{\renewcommand{\baselinestretch}{#1}\small\normalsize}

\title{Optimal Variable Clustering for\\ High-Dimensional Matrix Valued Data}

\author{Inbeom Lee\thanks{ \fontsize{9pt}{0.4cm}\selectfont Department of Statistics and Data  Science, Cornell University, Ithaca, NY. E-mail: \texttt{il279@cornell.edu}.}~~~~~Siyi Deng\thanks{\fontsize{9pt}{0.4cm}\selectfont TikTok, Mountainview, CA. E-mail: \texttt{dsyly@gmail.com}.}~~~~~Yang Ning\thanks{\fontsize{9pt}{0.4cm}\selectfont Department of Statistics and Data Science, Cornell University, Ithaca, NY. E-mail: \texttt{yn265@cornell.edu}. }}

\date{}

\maketitle

\begin{abstract}
\noindent Matrix valued data has become increasingly prevalent in many applications. Most of the existing clustering methods for this type of data are tailored to the mean model and do not account for the dependence structure of the features, which can be very informative, especially in high-dimensional settings or when mean information is not available. To extract the information from the dependence structure for clustering, we propose a new latent variable model for the features arranged in matrix form, with some unknown membership matrices representing the clusters for the rows and columns. Under this model, we further propose a class of hierarchical clustering algorithms using the difference of a weighted covariance matrix as the dissimilarity measure. Theoretically, we show that under mild conditions, our algorithm attains clustering consistency in the high-dimensional setting. While this consistency result holds for our algorithm with a broad class of weighted covariance matrices, the conditions for this result depend on the choice of the weight. To investigate how the weight affects the theoretical performance of our algorithm, we establish the minimax lower bound for clustering under our latent variable model in terms of some cluster separation metric. Given these results, we identify the optimal weight in the sense that using this weight guarantees our algorithm to be minimax rate-optimal. The practical implementation of our algorithm with the optimal weight is also discussed. Simulation studies show that our algorithm performs better than existing methods in terms of the adjusted Rand index (ARI). The method is applied to a genomic dataset and yields meaningful interpretations.
\end{abstract}
		
{\small \noindent \textit{Keywords}: Clustering, matrix data, high dimensional estimation, minimax optimality, latent variable model, hierarchical algorithm}	

\spacingset{1.13}

\section{Introduction}
Cluster analysis is one of the most important unsupervised learning techniques and has been widely used to discover the underlying group structure in data, arising in many applications including economics, image analysis, psychology and the biomedical sciences  \citep{everitt2011cluster,kogan2007introduction}. In these applications, matrix valued data is becoming increasingly prevalent. For example, in genetic studies, one may observe data matrices of dimension $p \times q$ from $n$ subjects, where the $(j,k)$th entry corresponds to the expression value of the $j$th gene at the $k$th tissue \citep{AGEMAP}. The biologist is often interested in identifying clusters of genes that share similar biological functions and also clusters of similar tissues. Similarly, in functional magnetic resonance imaging (fMRI) studies, to understand how the brain connectivity structure changes under different tasks/stimuli, researchers can measure the blood oxygen level (BOLD) within each region of interest (ROI) from the brain under a variety of conditions \citep{mitchell2008predicting}. The data points from each participant can be stacked as a matrix, in which the rows and columns correspond to different ROIs and tasks/stimuli, respectively. Given the data from $n$ participants, it is of interest to simultaneously cluster the ROIs and tasks/stimuli. Such clustering results can be used as a dimension reduction step to further investigate brain connectivity networks \citep{eisenach2020high}. However, in many cases such as when doing genetic studies, the data is already pre-processed and centered, and this loss of mean information prevents the use of existing mean-based clustering methods. Driven by these applications and limitations of existing methods, the goal of this paper is to develop a statistical clustering framework and a feasible algorithm for matrix valued data that utilizes covariance information with good theoretical guarantees. In particular, assuming that $n$ i.i.d. samples $X^{(1)},\ldots,X^{(n)}$ of a random matrix $X\in \mathbb{R}^{p\times q}$ are observed, we aim to recover the cluster membership of the rows and columns of the feature matrix $X$.

In the literature, clustering a data matrix is often known as biclustering \citep{hartigan1972direct,madeira2004biclustering}. Most of the existing biclustering methods can be classified into the following categories: (1) hierarchical approaches based on the dendrogram \citep{Dendro}; (2) extensions of K-means \citep{fraiman2020biclustering}; (3) the penalized likelihood approach \citep{sparseBC}; (4) convex clustering via fused lasso \citep{convexbiclustering}; and (5) clustering based on the singular value decomposition \citep{sill2011robust,lee2010biclustering}. In these existing works, the goal is to simultaneously cluster $n$ samples and $d$ features stacked as an $n\times d$ data matrix, whereas in our problem we are interested in clustering a $p\times q$ feature matrix. Since the features are often correlated, the $p\times q$ feature matrix induces a sophisticated but informative dependence structure that is not accounted for in the existing biclustering literature. When also considering that there are instances where the data lacks mean information completely as in the case of centered gene-tissue data, it is apparent that there is a gap between existing biclustering methods and methods needed in various applications. More recently, multiway clustering, also known as tensor clustering, is attracting increasing attention  \citep{mankad2014biclustering,zhao2016identifying,TensorClustering,sun2019dynamic,wang2019three,wang2019multiway}. Since we observe $n$ i.i.d $p\times q$ feature matrices, we can view our data as a 3-way tensor with dimension $p\times q\times n$. The existing tensor clustering methods such as the tensor block model  \citep{wang2019multiway,TensorClustering} aim to  cluster each mode of the tensor from the mean model. While these approaches enjoy great success in many applications and can be adapted to our setting, they may not perform well when the dependence structure of the features holds more information than the mean structure. There are also several recent works on tensor clustering that utilize some sort of dependence structure (\cite{deng2022tensor}, \cite{mai2022doubly}), but these methods focus on clustering over the observations and not over the features (i.e., the rows and the columns of $X$) and thus is not directly applicable to our setting.

In this paper, our first contribution is to propose a new latent variable model for clustering matrix valued data. Assume the features are stacked as a random matrix $X\in\mathbb{R}^{p\times q}$, which follows
\begin{equation}\label{eq_model}
X=AZB^T+\Gamma,
\end{equation}
where without loss of generality, $Z\in\mathbb{R}^{K_1\times K_2}$ is a latent variable matrix with $E(Z)=0$. $A\in \mathbb{R}^{p\times K_1}$ and $B\in \mathbb{R}^{q\times K_2}$ are the unknown binary membership matrices for the rows and columns, respectively, and $\Gamma\in\mathbb{R}^{p\times q}$ represents the random noise matrix with entries that have mean 0 and variance $\sigma_{ab}^2=\Var(\Gamma_{ab})$. We assume that the entries in the noise matrix $\Gamma$ are mutually independent and are also independent of the entries in $Z$. Entries of the membership matrix $A$ take values in $\{0,1\}$, such that $A_{ak}=1$ if row $a$ belongs to row cluster $k$ and $A_{ak}=0$ otherwise. In this paper, we focus on the non-overlapping and exhaustive clustering scenario. That is, for each row $a$, there exists one and only one cluster $k$ with $A_{ak}=1$. The same requirement holds for the  membership matrix $B$. To see why $A$ and $B$ are interpreted as membership matrices, we note that, for any feature $X_{ab}$, if $A_{ak}=1$ and $B_{b\ell}=1$ for some $k$ and $\ell$, then model (\ref{eq_model}) implies $X_{ab}=Z_{k\ell}+\Gamma_{ab}$. That is, it implies that the feature $X_{ab}$ is associated with the latent variable $Z_{k\ell}$. For this reason, we say that $X_{ab}$ belongs to row cluster $k$ and column cluster $\ell$. Under model (\ref{eq_model}), we can formally define the row clusters as the partition 
\begin{equation}\label{eq_cluster}
\cG^{(r)}=\{G_1^{(r)},...,G_{K_1}^{(r)}\}~~\textrm{of}~~\{1,...,p\}, ~~\textrm{where}~~ G_k^{(r)}=\{a: A_{ak}=1\}
\end{equation} 
for any $1\leq k\leq K_1$, where $K_1$ is the unknown number of row clusters. When the context is clear, $\cG$ will be used for notational simplicity. Without loss of generality, we focus on how to recover the unknown membership matrix $A$ for the rows (or equivalently $\cG^{(r)}$) up to label switching. 

Model (\ref{eq_model}) can be viewed as the extension of the G-block model for clustering a random vector in \cite{clustering} to matrix valued data. Indeed, if we vectorize the matrix $X$, model (\ref{eq_model}) is equivalent to $\textrm{vec}(X)=M\textrm{vec}(Z)+\textrm{vec}(\Gamma)$ with $M=B\otimes A$, where $\textrm{vec}(X)$ denotes the vectorization of $X$, formed by stacking the columns of $X$ into a single column vector, and $\otimes$ denotes the Kronecker product. Thus, compared to \cite{clustering} which allows $M$ to be any unstructured $(pq)\times (K_1K_2)$ membership matrix, we impose the Kronecker product structure to the membership matrix $M$. While our model for $\textrm{vec}(X)$ is more restrictive than the model in \cite{clustering}, it actually comes with two advantages for matrix clustering. First, as seen above, $A$ and $B$  are interpreted as the membership matrices for the rows and columns. Ignoring the Kronecker product structure and directly applying the model in \cite{clustering} would no longer produce interpretable results for matrix clustering, as shown in Figure \ref{comparison_cord} in Section \ref{vectorized} in the Supplementary Material. Second, the Kronecker product of $A$ and $B$ provides a more parsimonious parametrization for the unknown membership matrix $M$, leading to stronger theoretical guarantees on clustering. %In particular, we show that under our model (\ref{eq_model}) the cluster separation condition required for clustering consistency is weaker than that required in \cite{clustering}.

%We note that although our model and method share similarities with some existing works, there are also key differences. For example, although the dimensions of the data of interest $X$ may be the same, unlike the aforementioned works on tensor clustering, our model (\ref{eq_model}) implies $\EE(X)=0$ and the cluster structure is characterized entirely by the dependence structure of $X$ due to the latent variable matrix $Z$. While there are recent works on tensor clustering that utilize the dependence structure (\cite{deng2022tensor}, \cite{mai2022doubly}), these methods focus on clustering over the observations and not over the features (i.e., the rows and the columns). 

%We can also generalize our proposed method to accommodate both mean and covariance information in clustering for the case where both components are expected to be meaningful. This will be discussed further in Section \ref{sec_extension} in the paper and Section \ref{supp_extension} in the Supplementary Material.

It is also worth mentioning that our proposed latent variable model shares similarities with the stochastic block model (SBM) widely used in community detection in network analysis. First introduced in \cite{holland1983stochastic}, the stochastic block model assumes that nodes of a network are partitioned into subgroups called blocks and  the distribution of the ties between nodes is dependent on the blocks to which the nodes belong. More specifically, they imposed the model $\EE(A) = ZBZ^T$ where $A$ is the $n \times n$ adjacency matrix, $Z$ is the $n \times K$ binary block membership matrix and $B$ is the $K \times K$ matrix whose element $B_{ij}$ represents the probability that a node from block $i$ is connected to a node in block $j$. We defer the details on further development of the method to \cite{airoldi2008mixed,zhang2020detecting,abbe2017community} and the references therein. It must be noted, however, that community detection with stochastic block type models is inherently different from our setting since the former is focused on clustering the nodes (i.e., the samples), whereas our matrix clustering setting is focused on clustering the features (the rows and the columns of $X$).

Our second contribution is to propose a class of hierarchical clustering methods based on the weighted covariance matrix $\Sigma_{p,W}=\EE(XWX^T)$ for some pre-specified positive semi-definite matrix $W\in\mathbb{R}^{q\times q}$ with the aim to recover the unknown membership matrix $A$ (and similarly for $B$ as well). We use the difference in entries of this weighted covariance matrix to define the dissimilarity measure in our hierarchical algorithm to recover the membership matrix $A$. To establish the theoretical guarantees of our algorithm, we introduce the metric $\MCOD(\Sigma_{p,W})/\|X\|_W$ to quantify how well the clusters are separated. The precise definition is detailed in Section \ref{sec_consistency}. Theoretically, we develop a general result on the clustering consistency of our hierarchical algorithm that holds for a broad class of weight matrices $W$. To attain clustering consistency for the rows, we require the cluster separation metric $\MCOD(\Sigma_{p,W})/\| X \|_W$ to be no smaller than the order of $\{(\log p) / (nK_2)\}^{1/2}$, where $K_2$ is the unknown number of column clusters. The implication is that with the help of a larger $K_2$, clustering the rows of $X$ becomes easier. 

To investigate the optimality of our algorithm, we establish the minimax lower bound for clustering under our latent variable model. While the clustering consistency property holds for a broad class of weight matrices $W$, our algorithm with a generic weight $W$ may not be minimax optimal. To derive an optimal clustering algorithm for the rows, the key is to account for the information in the column clusters. The intuition is that with a more accurate column cluster result, we can decorrelate the dependence structure of $X$ and reduce the noise induced by $\Gamma$, which in turn improves the clustering accuracy for the rows. Following this argument, we define the optimal weight $W_O$, and propose to estimate it by $\hat W_O=\hat{B}(\hat{B}^T\hat{B})^{-2}\hat B^T/s$, where $\hat{B}$ is an estimate of the column membership matrix $B$ and $s$ denotes the estimated number of clusters obtained from $\hat B$. Under mild conditions, we show that the proposed algorithm with the estimated optimal weight  attains the minimax lower bound, and therefore is rate-optimal for clustering. To the best of our knowledge, our paper is the first to formally establish minimax optimality for clustering matrix valued data. From a technical perspective, to show the optimality of our algorithm, the main challenge is to quantify the stability of the algorithm with respect to an imperfect estimate $\hat B$ as the weight matrix. In Section \ref{sec_example} we provide sufficient conditions to show the stability of the algorithm under some additional modeling assumptions. In practice, the hierarchical algorithm can be applied iteratively to cluster the rows and columns for increased accuracy. Finally, we conduct extensive numerical studies to support our theoretical results.

The rest of the paper is organized as follows. In Section \ref{sec_method}, we propose the hierarchical algorithm and discuss the advantage of using the optimal weight, $W_O$. In Section \ref{sec_theory}, we define a notion of cluster separation and use it to establish the clustering consistency of our algorithm and to derive the minimax lower bound. Since the algorithm using the optimal weight depends on the initial estimate $\hat B$, in Section \ref{sec_example}, we verify the cluster separation condition and stability condition required in Theorem \ref{thm_upper} for clustering consistency in matrix normal models. The practical implementation of the algorithm is discussed in Section \ref{sec_practical}. Three further extensions of our method - the dependent noise model, the nested clustering method to incorporate both mean and covariance information, and higher order tensor models - are discussed in Section \ref{sec_extension}. The simulation results and real data analysis are presented in Sections \ref{sec_simu} and \ref{sec_real}, respectively. The paper concludes with a discussion in Section \ref{sec_discussion}.

{\bf Notation}. 
For any $1\leq a\neq b\leq p$, we write $a \sim_{\cG} b$ if $a$ and $b$ are in the same cluster (i.e., $a, b\in G_k$ for some $k$). Otherwise, we write $a \nsim_{\cG} b$. 
For a matrix $X$, we use the following norms: $||X||_{\mathrm{max}}=\max_{i,j}|X_{ij}|$,  $||X||_{\infty} = \max_{i} \sum_{j=1}^q |X_{ij}|$,  $||X||_{F}=(\sum_{i,j} X_{ij}^2)^{1/2}$. $||X||_{\text{op}}$ denotes the largest singular value of $X$. The largest and smallest eigenvalues are denoted by $\lambda_{\max}(\cdot)$ and $\lambda_{\min}(\cdot)$. We use $X_{\cdot j}$ and $X_{j\cdot}$ to denote the $j$th column and row of $X$, respectively. For two positive sequences $a_n$ and $b_n$, we write $a_n\asymp b_n$ if $C\leq a_n/b_n\leq C'$ for all $n$ for some constant $C,C'>0$. Similarly, we use $a_n\lesssim b_n$ ($a_n\gtrsim b_n$) to denote $a_n\leq Cb_n$ ($a_n\geq Cb_n$) for all $n$ for some constant $C>0$. 

\section{Methodology}\label{sec_method}

\subsection{Hierarchical Clustering via Weighted Covariance Differences}\label{sec_alg}

Recall that the random matrix $X$ follows model (\ref{eq_model}).  In this section, we propose a class of clustering methods for the rows of $X$ based on the weighted covariance matrix $\Sigma_{p,W}=\EE(XWX^T)$, where $W\in\RR^{q\times q}$ is some positive semi-definite matrix to be chosen. We add a subscript $p$ to indicate $\Sigma_{p,W}$ is a $p\times p$ matrix corresponding to the rows. The same type of method can be used to cluster the columns of $X$. 

In the following, we first outline how to identify the unknown membership matrix $A$ from the weighted covariance matrix $\Sigma_{p,W}$ on the population level. Under model (\ref{eq_model}), by the independence between $Z$ and $\Gamma$, we obtain that 
\begin{equation}\label{eq_sigma}
\Sigma_{p,W}=A\EE(ZB^TWBZ^T)A^T+\EE(\Gamma W\Gamma^T).
\end{equation}
To recover the membership matrix $A$ from $\Sigma_{p,W}$, one needs to first separate the two matrices $A\EE(ZB^TWBZ^T)A^T$ and $\EE(\Gamma W\Gamma^T)$. Noting that $\EE(\Gamma W\Gamma^T)$ is a diagonal matrix as the elements in $\Gamma$ are mutually independent and the weight matrix $W$ is deterministic, we therefore focus on the non-diagonal entries of $\Sigma_{p,W}$, that is $[\Sigma_{p,W}]_{ac}=[A\EE(ZB^TWBZ^T)A^T]_{ac}$ for any $1\leq a\neq c\leq p$. By the definition of the membership matrix $A$, for any $a\in G_k$, $c\in G_\ell$ and $a\neq c$, we have
\begin{equation}\label{eq_sigma_ac}
\big(\Sigma_{p,W}\big)_{ac}=\Big[A\EE(ZB^TWBZ^T)A^T\Big]_{ac}=\Big[\EE(ZB^TWBZ^T)\Big]_{k\ell}.
\end{equation}
In view of (\ref{eq_sigma_ac}), the within-cluster covariance difference $(\Sigma_{p,W})_{ac}-(\Sigma_{p,W})_{bc}$ with $a \sim_{\cG} b$ is always 0 for any $c\neq a,b$. In other words, as long as $(\Sigma_{p,W})_{ac}-(\Sigma_{p,W})_{bc}$ is nonzero for some $c$, it indicates that $a$ and $b$ are not in the same cluster. Thus, the covariance difference is  indicative of the clustering structure. Following \cite{clustering}, we formally define the covariance difference ($\COD$) as
\begin{align}\label{eq_COD}
    \mathrm{\COD}_{\Sigma_{p,W}}(a,b) &:= \underset{c \neq a,b}{\text{max}}~ \Big|\big(\Sigma_{p,W}\big)_{ac}-\big(\Sigma_{p,W}\big)_{bc}\Big|.
\end{align}
From the above argument, we have $\COD_{\Sigma_{p,W}}(a,b)=0$ if $a \sim_{\cG} b$. Moreover, if $\COD_{\Sigma_{p,W}}(a,b)>0$ holds for all  $a \nsim_{\cG} b$, we are able to identify all the clusters. Let 
\begin{align}\label{eq_MCOD}
\MCOD(\Sigma_{p,W}) &:=\min_{a\underset{\cG}{\nsim}_{\cG} b}~\COD_{\Sigma_{p,W}}(a,b)
\end{align}
denote the minimum $\COD$ value over all possible $a\nsim_{\cG} b$. On the population level, provided $\MCOD(\Sigma_{p,W})>0$, the membership matrix $A$ (or equivalently $\cG^{(r)}$ in (\ref{eq_cluster})) is identifiable from the weighted covariance matrix $\Sigma_{p,W}$ up to label switching. 

Based on the above results on the population level, we will develop a hierarchical clustering algorithm to estimate the membership matrix $A$ (or equivalently $\cG^{(r)}$ in (\ref{eq_cluster})). Given $n$ i.i.d. samples $X^{(1)},...,X^{(n)}$, we first estimate $\Sigma_{p,W}$ with
$$
\hat{\Sigma}_{p,W} = \frac{1}{n}\sum_{i=1}^n X^{(i)}WX^{(i)T},
$$
and then plug this into (\ref{eq_COD}) to form a dissimilarity measure $\COD_{\hat\Sigma_{p,W}}(a,b)$ for any $1\leq a,b\leq p$. The hierarchical algorithm starts with every variable representing a singleton cluster. At each step, the closest two clusters are merged into one single cluster based on the following dissimilarity measure between two sets $I$ and $J$
\begin{equation}\label{eq_CODAB}
\COD_{\hat\Sigma_{p,W}}(I,J)=\max_{a\in I, b\in J}\COD_{\hat\Sigma_{p,W}}(a,b).
\end{equation} 
We refer to \cite{Dendro} for alternative definitions of dissimilarity between two clusters and further discussions. Finally, we terminate this process and report the clusters when the dissimilarity measure $\COD_{\hat\Sigma_{p,W}}(I,J)$ exceeds a threshold value $\alpha>0$. This hierarchical algorithm, summarized in Algorithm \ref{alg_1}, improves the existing $\COD$ algorithm proposed by \cite{clustering}. First, our algorithm satisfies the so-called monotonicity property, which states that merged clusters always have smaller values of $\COD_{\hat\Sigma_{p,W}}(I,J)$ than unmerged ones \citep{Dendro}. However, the algorithm in \cite{clustering} may merge indices $a$ and $b$ into one cluster even if there exists another index $c$ with  $\COD_{\Sigma_{p,W}}(a,c)$ smaller than $\COD_{\Sigma_{p,W}}(a,b)$. We include in Section \ref{app_compare} in the Supplementary Material a toy example that illustrates this point. Empirically, we find that our hierarchical algorithm produces more stable clustering results than the algorithm in \cite{clustering}. Second, the hierarchical algorithm is more flexible in incorporating side information, such as the number of row clusters $K_1$. While our Algorithm \ref{alg_1} does not require the user to know $K_1$, with such information from domain knowledge or existing literature, the algorithm is expected to yield more reliable clustering results.

\begin{algorithm}
\caption{Hierarchical Algorithm with $\COD$}
    \textbf{INPUT}: The estimated weighted covariance matrix $\hat\Sigma_{p,W}$ and a threshold $\alpha>0$.
    \begin{itemize}
    \item[(1)] Calculate $\COD_{\hat\Sigma_{p,W}}(a,b)$ for $1\leq a,b \leq p$.
    \item[(2)] Create a hierarchical tree based on the value of \begin{itemize}
    \item[] $\COD_{\hat\Sigma_{p,W}}(I,K) =\underset{a \in I,~b \in K}{\max}\COD_{\hat\Sigma_{p,W}}(a,b)$ for sets $I$ and $K$.
    \end{itemize}
    \item[(3)] Use the threshold value $\alpha$ to cut the tree and obtain the estimated row clustering $\hat \cG^{(r)}$. More precisely, we use the following rule to find the clusters:
    \begin{itemize}
    \item[] For any two sets of candidate clusters $I$ and $K$ from the hierarchical tree, \\ merge them into one cluster if and only if $\COD_{\hat\Sigma_{p,W}}(I,K)\leq \alpha$.
    \end{itemize}
    \end{itemize}
\label{alg_1}
\end{algorithm}

\subsection{Optimal Choice of $W$} \label{sec_optimal}
While our Algorithm \ref{alg_1} can be applied with any weight matrix $W$ in $\hat\Sigma_{p,W}$, the empirical and theoretical performance of the algorithm critically depends on the choice of $W$. In practice, the simplest choice of $W$ could be $W_I=I_q/q$, where $I_q$ is a $q\times q$ identity matrix. With this choice of $W_I$, the weighted covariance matrix $\Sigma_{p,W_I}=\frac{1}{q}\sum_{j=1}^{q}\EE(X_{\cdot j}X_{\cdot j}^T)$ can be interpreted as the average of the second order moment of the columns of $X$. This ``naive" weight $W_I$ can be used directly in our Algorithm \ref{alg_1} (named $\NAIVE$ $\COD$) or it can be used as an initial value in a multi-step iterative algorithm ($\ONESTEPCOD$, $\TWOSTEPCOD$), which will be further discussed in Section \ref{sec_practical}. %Generally, NAIVE COD is not as competitive as the aforementioned iterative methods and the performance of the algorithm can be improved if more information on the column cluster structure can be taken into account. 

To motivate the development of the optimal choice of $W$, we temporarily assume that the true column cluster structure (i.e., the membership matrix $B$) is known up to label switching. We define $X^{*}=XB(B^TB)^{-1}\in\RR^{p\times K_2}$, which can be interpreted as the average of $X$  over columns in the same column cluster. To see this, let us consider a toy example. Assume that $X$ has $q=4$ columns with $K_2=2$ column clusters, where the first two columns belong to cluster 1 and the last two columns belong to cluster 2. In this case, the membership matrix $B$ can be written as  
$$
B=\begin{bmatrix}
1 & 0\\
1 & 0\\
0 & 1\\
0 & 1
\end{bmatrix}\in \RR^{4\times 2}.
~~\textrm{Then we have}~~ 
X^{*}=
\begin{bmatrix}
\frac{X_{11}+X_{12}}{2}&\frac{X_{13}+X_{14}}{2}\\
:&:\\
\frac{X_{p1}+X_{p2}}{2}&\frac{X_{p3}+X_{p4}}{2}\\
\end{bmatrix}\in\RR^{p\times 2}.
$$
Clearly, the two columns of $X^{*}$ represent the averages of $X$ in the same column cluster. Inspired by the interpretation of $\Sigma_{p,W_I}$, we now compute the average of the second order moment of the columns of $X^{*}$ as follows:
\begin{equation}\label{eq_sigma}
\frac{1}{K_2}\sum_{j=1}^{K_2}\EE(X^*_{\cdot j}X_{\cdot j}^{*T})=\frac{1}{K_2}\EE(X^{*}X^{*T})=\EE(XW_OX^T)=\Sigma_{p,W_O},
\end{equation}
where we set $W_O=B(B^TB)^{-2}B^T/K_2$ by the definition of $X^*$. 
This matrix $W_O$ is the optimal weight for reasons that will be explained in Section \ref{sec_theory}. Intuitively, the weighted covariance matrix $\Sigma_{p,W_O}$ with $W_O$ is more informative for clustering than $\Sigma_{p,W_I}$, as the random noise in $\Gamma$ is reduced when we construct $X^*$ by aggregating the columns of $X$ in the same column cluster. To better illustrate this point, we consider a special case. Assume that the columns of $X$ have $K_2$ clusters with equal size $q/K_2$. After some algebra, it is shown that 
$$
\Sigma_{p,W_I}=\frac{1}{K_2}A\EE(ZZ^T)A^T+\frac{1}{q}\EE(\Gamma\Gamma^T), 
$$
and
$$
\Sigma_{p,W_O}=\frac{1}{K_2}A\EE(ZZ^T)A^T+\frac{K_2}{q}\bigg\{\frac{1}{q}\EE(\Gamma\Gamma^T)\bigg\}. 
$$
Clearly, both $\Sigma_{p,W_I}$ and $\Sigma_{p,W_O}$ contain the same amount of row cluster information via the term $A\EE(ZZ^T)A^T/K_2$. However, compared to $\Sigma_{p,W_I}$, the error matrix induced by the covariance of $\Gamma$ is further reduced by a factor of $K_2/q$ in $\Sigma_{p,W_O}$. Therefore, we expect the clustering algorithm using $\Sigma_{p,W_O}$ to outperform the naive method using $\Sigma_{p,W_I}$ for recovering the membership matrix $A$, which is indeed the case in simulations; see Section \ref{sec_simu}.

Since $W_O=B(B^TB)^{-2}B^T/K_2$ depends on the unknown membership matrix $B$, Algorithm \ref{alg_1} is not directly applicable. In principle, if an initial estimate of $B$, say $\hat B$, is available, we can plug in the estimator $\hat B$ and apply Algorithm \ref{alg_1} with $\hat\Sigma_{p,\hat W_O}$, where $\hat W_O=\hat B(\hat B^T\hat B)^{-2}\hat B^T/s$ and $s$ denotes the estimated number of clusters from $\hat B$. Theoretically, in the next section, we will establish a general result on the clustering consistency of Algorithm \ref{alg_1} with a data dependent weight matrix $\hat W$, which covers the case with $\hat W_O$. In practice, we recommend using an iterative hierarchical algorithm to repeatedly cluster the rows and columns of $X$. The detailed implementation of the algorithm using the optimal weight is discussed in Section \ref{sec_practical}.

\section{Theoretical Guarantees}\label{sec_theory}

In this section we establish the theoretical results for our proposed clustering method. In Section \ref{sec_consistency}, we present a general result on the clustering consistency of Algorithm \ref{alg_1} with a data dependent weight matrix $\hat W$. Subsequently, we develop the minimax lower bound for the matrix clustering problem in Section \ref{sec_minimax}. In particular, these  results imply that Algorithm \ref{alg_1} with the estimated optimal weight matrix $\hat W_{O}$ defined in Section \ref{sec_optimal} is minimax optimal for clustering.

\subsection{Clustering Consistency of Algorithm \ref{alg_1}}\label{sec_consistency}

To formally study the clustering consistency property, we need to first define a proper notion of cluster separation distance. Recall from the argument in Section \ref{sec_alg} that $\MCOD(\Sigma_{p,W})>0$ implies the identifiability of $A$ up to label switching. One may attempt to use $\MCOD(\Sigma_{p,W})$ to measure cluster separation. But, using $\MCOD(\Sigma_{p,W})$ alone is not ideal, as $\MCOD(\Sigma_{p,W})$ is not invariant to the scale of $W$. To be specific, we have $\MCOD(\Sigma_{p,tW})=t\cdot\MCOD(\Sigma_{p,W})$ for any $t>0$, implying that the MCOD value can be arbitrarily large by rescaling $W$. 

In order to resolve this issue, we define $||X||_W$ as follows:
$$
\|X\|_W=K_2^{1/2}\max_{1\leq a\leq p}\big\|L^T\Var(X_{a\cdot})L\big\|_F,
$$
where $L$ is a matrix satisfying $LL^T=W$. We note that $\|X\|_W$ can be interpreted as the amount of the variance of the row vector in $X$ reweighted by $L$. From a technical perspective, such a quantity plays a natural role when applying the concentration inequality to control  $\hat\Sigma_{p,W}-\Sigma_{p,W}$. For convenience, we also include a $K_2^{1/2}$ factor in $\|X\|_W$ to rescale the quantity to be of constant order. For example, under our model (\ref{eq_model}) and some mild conditions, we show in Section \ref{app_norm} of the Supplementary Material that $\|X\|_W=O(1)$ holds for a general class of weight matrices. 

In this paper, we define the cluster separation distance as  $\MCOD(\Sigma_{p,W})/\|X\|_W$. First, we can view $\MCOD(\Sigma_{p,W})/\|X\|_W$ as a standardized distance, which measures cluster separation per unit ``variance" of $X$. Second, $\MCOD(\Sigma_{p,W})/\|X\|_W$ is invariant to the scale of $W$ and also the scale of $X$ (e.g., transform $X$ to $tX$ for any $t\in\RR$). Finally, we note that our cluster separation metric depends on the choice of the weight matrix $W$ since we use a weighted covariance distance to construct the $\MCOD$ in (\ref{eq_MCOD}). Thus, even if we consider the same data generating model (\ref{eq_model}), the value of the cluster separation metric may differ depending on the choice of $W$. This has important implications for clustering consistency (see Remark 3.2 and Section \ref{app_remark} in the Supplementary Material) and the minimax lower bound (see Section \ref{sec_minimax}). 

If the weight $W$ is known, we can directly apply our Algorithm \ref{alg_1} with the input $\hat\Sigma_{p,W}$. However, if $W$ depends on unknown parameters (e.g., the weight $W_O = B(B^TB)^{-2}B^T/K_2$), we need to estimate $W$ and use a data dependent weight in Algorithm \ref{alg_1}. To study clustering consistency of the algorithm with a data dependent weight, we assume that there exists an initial estimator $\hat W$ of a deterministic weight matrix $W$.  To simplify the theoretical analysis, we focus on analyzing Algorithm \ref{alg_1} with sample splitting. Specifically, we randomly divide the data into two folds, $\{X^{(i)}: i\in D_1\}$ and $\{X^{(i)}: i\in D_2\}$, where $D_1\cap D_2=\emptyset$ and $D_1\cup D_2=\{1,...,n\}$. The estimator $\hat W$ is constructed using the data in $D_1$, and then we apply Algorithm \ref{alg_1} with the input $\hat{\Sigma}_{p,\hat W} = \sum_{i\in D_2} X^{(i)}\hat W X^{(i)T}/|D_2|$, where $\hat{\Sigma}_{p,\hat W}$ is the weighted sample covariance matrix using the data in $D_2$. By using this simple procedure, we can remove the dependence of $\hat W$ and the data in $D_2$. 

Let us denote $\Sigma_{p,\hat W}=\EE(X\hat WX^T|\hat W)$, where the expectation is taken with respect to $X$ and is independent of $\hat W$. The following main theorem in this section shows the clustering consistency of our algorithm in the non-asymptotic regime. 
\begin{theorem}
\label{thm_upper}
Under model (\ref{eq_model}), assume that $\textrm{vec}(X)$ is multivariate Gaussian, $\log p=o(n)$ and the following two conditions hold: 
\begin{itemize}
    \item[(A1)] Cluster separation condition: $\MCOD(\Sigma_{p,W})/\|X\|_W>c_0\eta$, where $\eta\geq c_1 \sqrt{\frac{\log p}{nK_2}}$ for a universal constant $c_1 > 0$ and an arbitrary constant $c_0 \geq 4$.
    \item[(A2)] Stability condition: $\Big\{\MCOD(\Sigma_{p,\hat W})/\|X\|_{\hat W}\Big\}\Big/\Big\{\MCOD(\Sigma_{p,W})/\|X\|_W\Big\}>4/c_0$, where $c_0$ is defined in (A1).
\end{itemize}
Then using our Algorithm \ref{alg_1} with $\widehat{\Sigma}_{p,\hat W}$ and the threshold  $\alpha=2\eta \cdot \|X\|_{\hat W}$, we obtain perfect row cluster recovery (i.e., $\hat\cG^{(r)}=\cG^{(r)}$) with probability greater than $1-\frac{c_2}{p}$ for some constant $c_2>0$.
\end{theorem}

In the following, we start from the discussion on the conditions in Theorem \ref{thm_upper}. The Gaussian assumption is imposed in order to derive a sharp bound for $(\hat\Sigma_{p,\hat W}-\Sigma_{p,\hat W})_{jk}$ when applying the Hanson-Wright inequality, as we can decorrelate two dependent Gaussian variables to make them independent. Recently, \cite{hwd} derived a variant of the Hanson-Wright inequality for dependent data with the so-called convex concentration property. 
Using this new inequality, we can relax the Gaussian assumption to more general distributions (e.g., sub-Gaussian) with the convex concentration property. The conclusion in Theorem \ref{thm_upper} remains the same. However, to keep our presentation focused, we impose the Gaussian assumption in this theorem.

The condition $\log p=o(n)$ is standard for high-dimensional data. We further assume two major conditions, (A1) and (A2). Recall from the previous discussion that the cluster separation is measured by  $\MCOD(\Sigma_{p,W})/\|X\|_W$. Condition (A1) implies that in the ideal case (i.e., the weight matrix $W$ is known), the clusters must be separated by a factor of  $\sqrt{\frac{\log p}{nK_2}}$. For this reason, we call (A1) the cluster separation condition. On top of (A1), we also need condition (A2), because in some cases the target weight matrix $W$ (e.g., the optimal weight $W_O$) needs to be estimated. Condition (A2) quantifies the stability of the cluster separation metric $\MCOD(\Sigma_{p,W})/\|X\|_W$ with respect to the perturbation of $W$. Essentially, condition (A2) guarantees that  $\MCOD(\Sigma_{p,\hat W})/\|X\|_{\hat W}$ with the estimate $\hat W$ is still beyond the order $\sqrt{\frac{\log p}{nK_2}}$. When $W$ does not need to be estimated, we can simply use $\hat\Sigma_{p,W}$ in Algorithm \ref{alg_1}, and in this case (A2) holds trivially. It is important to note that the arbitrary constant $c_0$ in condition (A2) is $\geq$ 4, which allows the separation based on $\hat W$ to be smaller than that on $W$. Indeed, the interplay between the two conditions (A1) and (A2) is characterized by $c_0$. A larger $c_0$ can relax the stability condition (A2), whereas the cluster separation condition (A1) becomes more stringent. Lastly, we note that (A1) and (A2) are both high-level technical conditions, which will be further explored in Section \ref{sec_example} and shown to always hold under additional specific modeling assumptions.   

One important implication of Theorem \ref{thm_upper} is that the minimum cluster separation  for clustering consistency is of order $\sqrt{\frac{\log p}{nK_2}}$, which decreases as the number of column clusters $K_2$ grows. In other words, clustering the rows of $X$ becomes easier if the columns of $X$ have more column clusters. Indeed, this phenomenon is reasonable as the data from two different column clusters show weaker dependence and therefore improve the convergence rate of $(\hat\Sigma_{p,\hat W}-\Sigma_{p,\hat W})_{jk}$ in the Hanson-Wright inequality. This result clearly demonstrates the benefit of clustering the matrix $X$ over vector clustering. %To be precise, consider the following simple alternative. We just take any column of $X$ and apply our algorithm to cluster this $p\times 1$ vector. This corresponds to a special case of our framework with $q=K_2=1$. From Theorem \ref{thm_upper}, we can see that the minimum cluster separation for clustering any column of $X$ is of order $\sqrt{\frac{\log p}{n}}$, which is much stronger than the condition (A1) when $K_2$ diverges. 
Finally, we note that in the special case $q=K_2=1$, the order of our cluster separation metric matches the existing result for vector clustering in \cite{clustering}. \\
\\
\noindent \textbf{Remark 3.2.}\label{rem_consistency}
The results in Theorem \ref{thm_upper} are generally applicable to our clustering algorithm with any positive semi-definite matrix $W$, provided (A1) and (A2) hold. Recall that in Section \ref{sec_optimal}, we consider two specific weights, $W_I=I_q/q$ and $W_O=B(B^TB)^{-2}B^T/K_2$, where the latter was called the optimal weight. We can apply our algorithm with either $\hat\Sigma_{p,W_I}$ or $\hat\Sigma_{p,\hat W_O}$, where $\hat W_O$ is an estimate of $W_O$ defined in Section \ref{sec_optimal}. %In this remark, we will discuss the implication of  Theorem \ref{thm_upper} on our algorithm with these two weights. 
Since the rate of the cluster separation for consistency is always $\sqrt{\frac{\log p}{nK_2}}$, which does not depend on $W$, one might be tempted to conclude that there is no benefit of using the optimal weight $W_O$ over $W_I$ for clustering consistency. However, this conclusion is imprecise, as the value of the cluster separation metric $\MCOD(\Sigma_{p,W})/\|X\|_W$ depends on $W$ and may differ substantially. When $q\gg K_2$, our algorithm with the  optimal weight attains clustering consistency in the presence of a larger noise level $\sigma^2$ compared to when using $W_I$. A more detailed illustration of this point can be found in Section \ref{app_remark} in the Supplementary Material.

\subsection{Minimax Optimality} \label{sec_minimax}

In this section, we establish the minimax lower bound for the matrix clustering problem. %Since the cluster separation distance $\MCOD(\Sigma_{p,W})/\|X\|_W$ depends on $W$, we can separately consider the lower bound results for the two cases $W=W_I$ and $W=W_O$.  
In this paper, we will focus on the optimal weight $W_O$ and from it construct an appropriate parameter space. %and defer the results for $W_I$ to Section \ref{supp_W_I} of the Supplementary Material. 
Assume that $\textrm{vec}(X)\sim N(0,\Sigma)$, where $\Sigma\in \RR^{pq\times pq}$. We define the relevant parameter space as
\begin{equation*}
    M_O(p,q,K_1,K_2,\eta)\\=\big\{ \Sigma\in \RR^{pq\times pq}\big|~X~\textrm{satisfies model}~(\ref{eq_model}), ~~\MCOD(\Sigma_{p,W_O})/\|X\|_{W_O}\geq \eta\big\},
\end{equation*}
where the cluster separation metric is defined based on $W_O$. The following theorem provides the lower bound for clustering over the parameter space $M_O:=M_O(p,q,K_1,K_2,\eta)$. 

\begin{theorem} \label{thm_lowerbound_2}
For $K_1 \geq 3$, there exists a positive constant $c$ such that, for any $\eta$ that satisfies
\begin{align*}
    0 \leq \eta < c\sqrt{\frac{\log p}{nK_2}}~,
\end{align*}
we have
\begin{align*}
\underset{\widehat{\cG}}{\inf}~\underset{\Sigma \in M_O}{\sup}~\PP_{\Sigma}(\widehat{\cG}\neq \cG) ~\geq~ \frac{1}{7},
\end{align*}
where the infimum is taken over all possible estimators of $\cG$.
\end{theorem}

Theorem \ref{thm_lowerbound_2} shows that it is impossible to attain clustering consistency uniformly over the parameter space $M_O$ when the minimum cluster separation value $\eta$ is below the threshold $\sqrt{\frac{\log p}{n K_2}}$, which matches the rate of $\eta$ in condition (A1) in Theorem \ref{thm_upper}. Thus, provided the stability condition (A2) in Theorem \ref{thm_upper} holds, our Algorithm \ref{alg_1} using $\widehat{\Sigma}_{p,\hat W_O}$ and $\alpha=2\eta \cdot \|X\|_{\hat{W}_O}$ is minimax optimal for clustering. 

We also have a minimax lower bound result in which the cluster separation metric is defined with an arbitrary column membership matrix $\bar B$. The setup of the appropriate parameter space, the presentation of the theorem, its derivation and discussion can be found in Section \ref{general_lower_bound} in the Supplementary Material.

\section{Applications to Matrix Normal Models}\label{sec_example}

As seen from the previous section, our Algorithm \ref{alg_1} using the weighted covariance matrix  $\widehat{\Sigma}_{p,\hat W_O}$ is minimax optimal, provided conditions (A1) and (A2) in Theorem \ref{thm_upper} hold. In this section, we will verify conditions (A1) and (A2) with $\hat W=\hat W_O$. To make the analysis of the cluster separation metric $\MCOD(\Sigma_{p,W})/\|X\|_W$ tractable, we will make some additional modeling assumptions. 

On top of our model (\ref{eq_model}), we further assume that the latent variable $Z$ follows the matrix normal distribution. That is $Z \sim \mathrm{MN}(0,U,V)$, where $U\in\RR^{K_1\times K_1}$ and $V\in\RR^{K_2\times K_2}$ are symmetric positive definite matrices. Note that this is equivalent to saying that $\mathrm{vec}(Z)\sim \mathrm{MVN}(\mathrm{vec}(0),V\otimes U)$, which gives us  $\EE(Z_{jk}Z_{j'k'})=U_{jj'}V_{kk'}$ and $\EE(Z_{jk})=0$ for any $1\leq j,j'\leq K_1$ and $1\leq k, k'\leq K_2$. 

Recall that $\hat W_O=\hat B(\hat B^T\hat B)^{-2}\hat B^T/s$, where $\hat{B}$ is an initial estimator of the column membership matrix $B$ and $s$ is the estimated number of clusters. To facilitate the analysis, we use $\cG^{(c)} = \{G_1^{(c)}, ... ,G_{K_2}^{(c)}\}$ and $\hat\cG^{(c)} = \{\hat{G}_1^{(c)}, ... ,\hat{G}_{s}^{(c)}\}$ to denote the true column cluster structure and the estimated column structure, respectively. We define a $K_2 \times s$ matrix $G$ that carries information about the clustering accuracy of the initial estimator $\hat B$:
\[G=B^T\hat{B}(\hat{B}^T\hat{B})^{-1}=\begin{bmatrix}
\frac{|G_1^{(c)} \cap \widehat{G}_{1}^{(c)}|}{|\widehat{G}_{1}^{(c)}|}&\frac{|G_1^{(c)} \cap \widehat{G}_{2}^{(c)}|}{|\widehat{G}_{2}^{(c)}|}&...&\frac{|G_1^{(c)} \cap \widehat{G}_{s}^{(c)}|}{|\widehat{G}_{s}^{(c)}|}\\
:&:&&:\\
\frac{|G_{K_2}^{(c)} \cap \widehat{G}_{1}^{(c)}|}{|\widehat{G}_{1}^{(c)}|}&\frac{|G_{K_2}^{(c)} \cap \widehat{G}_{2}^{(c)}|}{|\widehat{G}_{2}^{(c)}|}&...&\frac{|G_{K_2}^{(c)} \cap \widehat{G}_{s}^{(c)}|}{|\widehat{G}_{s}^{(c)}|}\\
\end{bmatrix}.
\]
Note that the columns of $G$ sum to $1$, and in the ideal case of $\cG^{(c)} = \hat\cG^{(c)} \text{ (i.e., } \hat{B}=B)$, we have that $G=I_{K_2}$. Denote by 
\begin{equation}\label{eq_CK}
C_K=\max_{1\leq a\leq p}\frac{1}{K_2}\sum_{t=1}^{K_2}\frac{(\sum_{j \in [t]}\sigma^2_{aj})^2}{|[t]|^4}  
\end{equation}
the maximum weighted average of the error variances over $K_2$ column clusters, where $[t]$ denotes the $t$-th column cluster. Similarly, we define 
\begin{equation}\label{eq_CS}
C_s=\max_{1\leq a\leq p}\frac{1}{s}\sum_{t=1}^{s}\frac{(\sum_{j \in [\hat t]}\sigma^2_{aj})^2}{|[\hat t]|^4}
\end{equation} 
where $[\hat t]$ denotes the estimated $t$-th column cluster. The following proposition shows the conditions under which (A1) and (A2) in Theorem \ref{thm_upper} hold in the matrix normal model. The proof can be found in Section \ref{matnor} of the Supplementary Material.

\begin{proposition}\label{matrixnormal}
Under model (\ref{eq_model}) and the above matrix normal model assumptions, if we assume:
\begin{enumerate}
    \item[(P0)] \hspace{0.01cm} $C_{\min} ~\leq~ \lambda_{\min}(V) ~\leq~ \lambda_{\max}(V) ~\leq~ C_{\max} $ for two positive constants $C_{\min}$ and $C_{\max}$. 
    \item[(P1)] \textit{(Cluster Separation)} For any $1 \leq j \neq k \leq K_1$, there exists $1 \leq l \leq K_1$ such that
    \begin{align*}
    \big|U_{jl}-U_{kl}\big| ~&>~ c_0 \cdot c_1\sqrt{\frac{\log(p)}{nK_2}} \cdot  \frac{K_2}{\text{tr}(V)} \cdot \Bigg\{||\text{diag}(U)||_{\max}\cdot C_{\max}+C_K^{1/2}\Bigg\} 
    \end{align*} 
    holds for a universal constant $c_1$ and an arbitrary constant $c_0 \geq 4$, where\\ $||\text{diag}(U)||_{\max}=\max_{1\leq j\leq K_1} U_{jj}$. 
    \item[(P2)]  \textit{(Stability)} Either (i) or (ii) holds:
        $$\text{(i)} \hspace{1cm} \begin{cases}
         \sqrt{\min(s,K_2)}\cdot|\lambda_{\max}(GG^T)|\cdot ||\text{diag}(U)||_{\max}\cdot C_{\max} ~\leq~ \sqrt{s}\cdot\sqrt{C_s}\\
        \frac{1}{\lambda_{\min}(GG^T)} ~\leq~ \frac{c_0}{8}\cdot\sqrt{\frac{C_K}{C_s}}\cdot\sqrt{\frac{K_2}{s}}      \end{cases}$$
        $$\text{(ii)} \hspace{1cm} 
        \begin{cases}  \sqrt{\min(s,K_2)}\cdot|\lambda_{\max}(GG^T)|\cdot ||\text{diag}(U)||_{\max}\cdot C_{\max} ~>~ \sqrt{s}\cdot\sqrt{C_s}\\
        \frac{\lambda_{\max}(GG^T)}{\lambda_{\min}(GG^T)}~\leq~ \frac{c_0}{8}\cdot\frac{C_{\min}}{C_{\max}}\cdot\frac{||\text{diag}(U)||_{\min}}{||\text{diag}(U)||_{\max}}\cdot \sqrt{\frac{K_2}{\min (s, K_2)}}
        \end{cases}$$
    where $c_0$ is the arbitrary constant from (P1), $C_K$ and $C_s$ are defined in (\ref{eq_CK}) and  (\ref{eq_CS}), and $||\text{diag}(U)||_{\min}=\min_{1\leq j\leq K_1} U_{jj}$. 
\end{enumerate} 
Then the conditions (A1) and (A2) in Theorem \ref{thm_upper} hold.
\end{proposition}

The assumption (P0) is standard and commonly used in the high-dimensional statistics literature \citep{basu2015regularized,cai2010optimal,bai2010spectral}. For (P1), it says that any two rows of the row covariance matrix $U$ cannot be nearly identical because otherwise the corresponding two clusters would not be identifiable. When the error variances are not too large (e.g., $\sum_{j \in [ t]}\sigma^2_{aj}\lesssim |[t]|^2$), then $C_K$ behaves like a constant. If we further assume $||\text{diag}(U)||_{\max}=O(1)$, then the cluster separation reduces to $\big|U_{jl}-U_{kl}\big|\gtrsim \sqrt{\frac{\log(p)}{nK_2}}$, which decreases with $n$ and $K_2$ as we have discussed before. Finally, (P2) requires the estimated cluster structure to be close enough to the true cluster structure. When the initial estimator $\hat B$ is reasonably accurate, we would expect $C_K$ and $C_s$ to be of the same order. When $K_2\asymp s$ and $||\text{diag}(U)||_{\max}\asymp ||\text{diag}(U)||_{\min}$, (P2) reduces to the condition that the smallest eigenvalue of $GG^T$ is bounded from below by a constant and its largest eigenvalue is bounded from above by a constant. Recall that when $\hat B$ is more accurate, $G$ would be closer to an identity matrix and therefore (P2) would be more likely to hold. Thus, (P2) essentially gives a sharp characterization of the ``contraction region" in which the algorithm with the imperfect initial estimator $\hat B$ still leads to clustering consistency. In fact, the following proposition gives a concrete example where $\hat{B}$ satisfies the condition (P2). 

\begin{proposition}\label{pro_stability}
Under the matrix normal model, assume that the noise variances are homogeneous and the cluster sizes are balanced, (i.e.,  $\forall i,j$, $\EE(\Gamma_{ij})=\sigma^2$, $\frac{M_q}{m_q}\leq C$ for some constant $C\geq 1$ where $M_q$ and $m_q$ are the largest and smallest column cluster sizes, respectively). If $c_0\geq 8C\big/\big(\frac{C_{\min}}{C_{\max}}\cdot \frac{||\text{diag}(U)||_{\min}}{||\text{diag}(U)||_{\max}}\big)$ and $\sqrt{M_q}\geq \frac{\sigma^2}{||\text{diag}(U)||_{\max}\cdot C_{\max}}$, then the stability condition (P2) in Proposition 4.1 is satisfied with $\hat{B}=I_q$. 
\end{proposition}

%This initial estimate $\hat{B}=I_q$, which in turn gives us $~\hat{W}_O~=~\frac{1}{q}\bar{B}(\bar{B}^T\bar{B})^{-2}\bar{B}^T~=~\frac{1}{q}I_q$ is actually equivalent to the $W_I$ that was considered in Section \ref{sec_optimal}. This ``naive" weight that assigns each column to its own column cluster is practical, since it does not require any prior knowledge on the column cluster structure. This is significant as it guarantees a feasible minimax optimal implementation of our Algorithm \ref{alg_1} for matrix normal models. 

The proof can be found in Section \ref{proof_pro_stability} of the Supplementary Material.

\section{Some Practical Considerations} \label{sec_practical}
In this section, we discuss the practical implementation of our algorithm using the optimal weight. First, we look at the iterative variant of our algorithm and afterwards we discuss miscellaneous aspects of the algorithm such as sample splitting, standardization and selecting the tuning parameter $\alpha$.
\subsection{The Iterative One-Step and Two-Step Methods}
 We first introduce the iterative one-step hierarchical algorithm with the optimal weight in Algorithm \ref{alg_2}. Skipping over the details on sample splitting which were already discussed in Section \ref{sec_consistency} and will be further discussed in Section \ref{sample_splitting}, we first apply our hierarchical Algorithm \ref{alg_1} to cluster the rows using $\bar{B}=I_q$ as the initial estimate for $B$ and $\hat{W}_{O,(r)} = \frac{1}{q}I_q$ as an estimate of the optimal weight matrix for the rows. This is the same as the initial estimate proposed in Proposition \ref{pro_stability}. Then we use the obtained $\hat{A}$ to construct an estimate of the optimal weight matrix for the columns $\hat{W}_{O,(c)} = \hat{A}(\hat{A}^T\hat{A})^{-2}\hat{A}^T/t$ (where $t$ is the number of clusters from $\hat{A}$), and use it in Algorithm \ref{alg_1} to cluster the rows to get $\hat{B}$. We provide the theoretical guarantees on consistency and minimax optimality for the iterative one-step Algorithm \ref{alg_2} in Section \ref{one_step_consistency} and Section \ref{one_step_minimax}, respectively, of the Supplementary Material.
 
 Intuitively, we can further repeat steps 2-3 in Algorithm \ref{alg_2} to iteratively cluster the rows and columns of $X$, leading to a multi-step algorithm. The two-step algorithm essentially repeats step 2 in Algorithm \ref{alg_2} one more time after step 3. The details can be found in Section \ref{two-step} of the Supplementary Material.\\ %In the simulations in Section \ref{sec_simu}, we compare our one-step algorithm with the two-step algorithm as well as the ``naive" algorithm, which only implements Algorithm \ref{alg_1} with the initial weight $W_I$.\\
 \\
\begin{algorithm}[H]
\caption{Iterative One-Step Hierarchical Algorithm with Optimal Weight}
\begin{enumerate}
    \item Split the data into two folds: \(D_1\), \(D_2\). Set an initial value $\bar B$ (e.g., $\bar B=I_q$). 
%    \item On data $D_1$, apply Algorithm \ref{alg_1} with the sample covariance matrix $\hat{\Sigma}_q = \frac{1}{p|D_1|}\sum_{i \in D_1}X^{(i)T}X^{(i)}$ to cluster the columns of $X$ and find an initial estimator $\bar B$.
    \item On data $D_1$, apply Algorithm \ref{alg_1} with $\hat\Sigma_{p,\hat W_{O,(r)}}=\frac{1}{|D_1|}\sum_{i \in D_1}X^{(i)}\hat W_{O, (r)} X^{(i)T}$ to \\
    cluster the rows of $X$, where $\hat W_{O,(r)}=\bar B(\bar B^T\bar B)^{-2}\bar B^T/s$ and $s$ denotes the \\
    estimated number of clusters from $\bar B$. Obtain the resulting row cluster $\hat\cG^{(r)}$ or equivalently the membership matrix $\hat A$.  
    \item On data $D_2$, apply Algorithm \ref{alg_1} with $\hat\Sigma_{q,\hat W_{O,(c)}}=\frac{1}{|D_2|}\sum_{i \in D_2}X^{(i)T}\hat W_{O,(c)} X^{(i)}$ to \\
    cluster the columns of $X$, where $\hat W_{O,(c)}=\hat A(\hat A^T\hat A)^{-2}\hat A^T/t$ and $t$ denotes the \\
    estimated number of clusters from $\hat A$. Obtain the resulting cluster $\hat\cG^{(c)}$ or \\
    equivalently the membership matrix $\hat B$.
\end{enumerate}\label{alg_2}
\end{algorithm}

\subsection{Sample Splitting, Standardization and Selecting $\alpha$} \label{sample_splitting}
We note that in Section \ref{sec_consistency}, to facilitate the theoretical analysis of our algorithm, we split the data into two folds. Such a sample splitting procedure is feasible in practice when $n$ is relatively large. However, in practice, when we apply our algorithm to the data with very small $n$ (e.g., $n=40$), sample splitting tends to yield unstable clustering results. Considering this, in practice, we chose to implement both steps 2 and 3 in Algorithm \ref{alg_2} on the entire dataset. A detailed comparison of the performance of our method with and without sample splitting is presented in a simulation study in Section \ref{data_split} of the Supplementary Material.

Another point to mention is that in practice, the feature matrix $X$ may have elements with differing variances. While the theoretical guarantees of our algorithm in Section \ref{sec_consistency} remain valid, the concentration bound for $(\hat\Sigma_{p, W}-\Sigma_{p, W})_{jk}$ via the Hanson-Wright inequality will be dominated by the variables with a larger variance. To tighten this upper bound, we recommend applying our algorithm to standardized data - $X$ whose elements all have mean 0 and variance 1. Empirically, we observe that clustering accuracy can be significantly improved when the algorithm is applied to the standardized data.

Finally, we note that, in steps 2 and 3 in Algorithm \ref{alg_2}, we need to choose the threshold value $\alpha$ in our hierarchical algorithm. In Section \ref{tuning} of the Supplementary Material, we present Algorithm \ref{alg_tuning}, a data-driven cross-validation scheme to choose the optimal threshold value $\alpha$. 

\section{Further Extensions} \label{sec_extension}
In this section we discuss further extensions of our model (\ref{eq_model}) and $\COD$ based methods of Algorithm \ref{alg_1}, \ref{alg_2} and \ref{alg_3} to account for a variety of more general settings.

\subsection{The Dependent Noise Model} \label{dependent_noise}
In model (\ref{eq_model}), we assume that the entries in the noise matrix $\Gamma$ are independent. In practice, the model can be more flexible by allowing the entries of $\Gamma$ to have a dependence structure. In this section, we consider the more general model with correlated noise variables and present the conditions needed for clustering consistency in Theorem \ref{correlated_consistency}. Define 
\begin{align}
    \gamma(\Sigma_{p,W}) ~~:=~~ \underset{1 \leq a,b \leq p}{\max}~~ \underset{c \neq a,b}{\max} ~~\Big| \big[\EE(\Gamma W\Gamma^T)\big]_{ac} - \big[\EE(\Gamma W\Gamma^T)\big]_{bc}\Big|. \label{gamma_def}
\end{align}
This quantity $\gamma(\Sigma_{p,W})$ plays an important role in determining how difficult the clustering problem is when using the $\COD$ method with correlated noise variables. More specifically, it contains information on the maximum difference of two different elements in the same row or column in the $p \times p$ weighted covariance matrix. Note that in the independent noise setting, $\gamma(\Sigma_{p,W})=0$  holds since all off-diagonal terms of $\EE(\Gamma W \Gamma^T)$ would be 0. %The smaller the $\gamma$ value, the easier it is for COD based clustering. 
The analog to the clustering consistency result in Theorem \ref{thm_upper} under this more general noise setting using $\gamma(\Sigma_{p,W})$ is outlined in the following theorem.
\begin{theorem} \label{correlated_consistency} 
(Consistency with Dependent Noise Variables)\\
Under the model $X=AZB^T+\Gamma$, assume that $\textrm{vec}(X)$ is multivariate Gaussian, $\log p=o(n)$ and the following two conditions hold: 
\begin{itemize}
    \item[(A1)] Cluster separation condition: 
    \begin{equation}\label{eq_corr_separation}
    \MCOD(\Sigma_{p,W})/\|X\|_W~~>~~c_0\eta ~+~ \frac{\gamma(\Sigma_{p,W})}{||X||_W},
    \end{equation}
    where $c_0 \geq 4$ is an arbitrary constant, $~\eta\geq c_1 \sqrt{\frac{\log p}{nK_2}}~$ for a universal constant $c_1 > 0$ and $\gamma(\Sigma_{p,W})$ is defined in (\ref{gamma_def}).
    \item[(A2)] Stability condition: 
    $$\bigg[\Big\{\MCOD(\Sigma_{p,\hat W})-\gamma(\Sigma_{p,\hat{W}})\Big\}\Big/\|X\|_{\hat W}\bigg] \Bigg/ \bigg[\Big\{\MCOD(\Sigma_{p,W})-\gamma(\Sigma_{p,W})\Big\}\Big/\|X\|_W\bigg]> \frac{4}{c_0}$$ 
    where $c_0$ is defined in (A1).
\end{itemize}
Then using our Algorithm \ref{alg_1} with $\widehat{\Sigma}_{p,\hat W}$ and the threshold  $\alpha=2\eta \cdot \|X\|_{\hat W} ~+~ \gamma(\Sigma_{p,\hat{W}})$, we obtain perfect row cluster recovery (i.e., $\hat\cG^{(r)}=\cG^{(r)}$) with probability greater than $1-\frac{c_2}{p}$ for some constant $c_2>0$.
\end{theorem}

Compared to Theorem \ref{thm_upper}, the cluster separation is  stricter as the $\frac{\gamma(\Sigma_{p,W})}{||X||_W}$ factor on the right hand side of (\ref{eq_corr_separation}) shows that $\MCOD$ has to be much larger to account for the correlated noise structure. Further discussions of this result can be found in Section \ref{supp_dependent_noise} of the Supplementary Material.

\subsection{Nested Clustering to Incorporate Mean Information} \label{mean_and_cov}
Our proposed model (\ref{eq_model}), Algorithm \ref{alg_1}, \ref{alg_2} and \ref{alg_3} on their own can have many uses in practice where the dependence structure of the data is of main importance. However, there may be instances where mean information is also informative. In order to take full advantage of both components, we generalize our model and propose a method to accommodate both the mean and covariance information in clustering. More specifically, we propose a non-centered latent variable model $X=M+AZB^T+\Gamma$, where $M=\EE(X)$ induces the first layer row and column cluster structures, and the membership matrices $A$ and $B$ encode the second layer row and column cluster structures (from the covariance of $X$), which is assumed to be nested inside the mean clusters. We further  propose a nested clustering algorithm in which a mean-based clustering method can be implemented first, and then on each cluster, our covariance-based method can be applied to capture the finer, more intricate partitions within each broad cluster. We also include a simulation study comparing the empirical performance of this nested method and our original Algorithm \ref{alg_3}. The full presentation can be found in Section \ref{supp_mean} of the Supplementary Material.

\subsection{Higher Order Tensor Models} \label{higher_tensor}
Recall that our original problem is to cluster $n$ i.i.d. $p \times q$ matrices over the $p$ and $q$ directions. We now discuss the extension our model to higher order tensor settings. Consider a three-way tensor $X\in \RR^{J\times P\times Q}$, which satisfies 
$X=Z\times_1 A\times_2 B\times_3 C~+~\Gamma$,  where $Z\in \RR^{K_1\times K_2\times K_3}$ is a three-way latent tensor with $\EE(Z)=0$, $A\in \RR^{J\times K_1}$, $B\in \RR^{P\times K_2}$ and $C\in \RR^{Q\times K_3}$ are the unknown membership matrices for the rows, columns and tubes of $X$, respectively, and $\Gamma\in\RR^{J\times P\times Q}$ represents the mean 0 random noise tensor. Here, $Z\times_1 A$ is the tensor $n$-mode product defined  in Section \ref{supp_tensor} of the Supplementary Material.  Given $n$ i.i.d copies of $X$, we generalize our algorithm to recover the membership matrices $A, B$ and $C$.  We also conduct a simulation study to assess the empirical performance of our algorithm. The full presentation can be found in Section \ref{supp_tensor} of the Supplementary Material.

\section{Simulation Results} \label{sec_simu}

We consider the following data generating process. We fix $p=q=100$, $K_1=K_2=10$ and a moderately unbalanced row and column cluster size structure (both having cluster sizes of $3$, $6$, $6$, $8$, $10$, $10$, $12$, $12$, $14$, $19$, respectively). This gives us the membership matrices $A$ and $B$. The latent variable $Z$ is generated from the matrix normal distribution $\mathrm{MN}(0_{K_1 \times K_2},U_{K_1 \times K_1},V_{K_2 \times K_2})$, where $U_{jk}=(-0.4)^{|j-k|}$ and $V_{jk}=0.3^{|j-k|}$. We further generate $\Gamma_{ij}\sim N(0,\sigma_{ij}^2)$, where we consider the following three settings for the noise variance: (1) homogeneous noise variances $\sigma_{ij}^2=15$; (2) heterogeneous noise variances proportional to the corresponding row and column cluster sizes: $\sigma_{ij}^2 = \frac{15pq\cdot v_{ij}}{\sum_{i,j}v_{ij}},  v_{ij}=\frac{m_p^{(i)}\cdot m_q^{(j)}}{\sqrt{\frac{pq}{K_1K_2}}}$; and (3) heterogeneous noise variances randomly generated from the Uniform distribution: $\sigma_{ij}^2 = \frac{15pq\cdot u_{ij}^{h}}{\sum_{i,j}u_{ij}^{h}}, u_{ij} \sim \mathrm{Unif}(0,1)$ where $h$ determines the level of heterogeneity. They will be referred to as the ``homogeneous", ``proportional" and ``random" cases, respectively. 

Recall that in Section \ref{sec_consistency}, the noise variances play an important role on the clustering consistency of our algorithm. We design these three settings to test the performance of our algorithm under different patterns of noise variances. We keep the mean noise variance to be 15 for all three cases. For the third setting, we set $h = 0.87$  because it gives us a similar level of heterogeneity as the second setting. In both the second and third settings, the standard deviation of the noise variances is around 7.95. Finally, we  generate $X$ from the model (\ref{eq_model}). We vary the sample size $n$ in the simulations and the simulations are repeated 30 times. To measure clustering accuracy, we consider the adjusted rand index (ARI) \citep{hubert1985comparing}. Note that an ARI value of 1 implies a perfect match between the true and estimated cluster partitions. The formal definition of the ARI is shown in Section \ref{metrics} of the Supplementary Material. 

We consider the following clustering methods: (1) \textsc{vanilla hierarchical}; (2) Algorithm \ref{alg_1} with the  initial weight $W_I$ (\textsc{naive cod}); (3) Algorithm \ref{alg_2} (\textsc{1-step cod}); and (4) Algorithm \ref{alg_3} (\textsc{2-step cod} in Section \ref{two-step} of the Supplementary Material). We emphasize that mean based clustering methods, such as K-means, cannot be used due to the centered nature of the data. This setting is indeed relevant as such data can be found in practice; see Section \ref{sec_real}. As such, a competing method, \textsc{vanilla hierarchical} was constructed by applying agglomerative hierarchical clustering on a certain similarity matrix that utilizes covariance information in the data. For the rows, the average of the second moment of the columns of $X$, $\EE(XX^T)$, was used, and for the columns, the average of the second moment of the rows of $X$, $\EE(X^TX)$, was used. The silhouette method was used to estimate the optimal number of clusters for this competing method. In comparison, we note that the data-driven cross-validation scheme in Section \ref{tuning} was used to choose the tuning parameter in our \textsc{cod} based algorithms.

The competing method works decently as we tailored it to utilize correlation information, but is outperformed by the \textsc{cod} based methods when $n\geq 18$. It is apparent that perfect clustering is attained at a much smaller $n$ value with \textsc{cod} based methods compared to the competing \textsc{vanilla hierarchical} method. 

In most cases, our iterative algorithms \textsc{1-step cod} and \textsc{2-step cod} improve the performance of \textsc{naive cod}, which is consistent with our theoretical analysis. In particular, \textsc{1-step cod} and \textsc{2-step cod} can achieve an ARI value close to 1 when $n=18$, whereas \textsc{naive cod} may require $n$ much larger than $30$ to attain the same level of accuracy. The phenomenon holds for both row and column clustering.  We also find that when $n$ is moderate (e.g., $n\geq 18$), \textsc{1-step cod} and \textsc{2-step cod} have very similar performances. Due to the extra computational cost of \textsc{2-step cod}, we generally recommend \textsc{1-step cod} for practical use if $n$ is moderate or large. However if $n$ is small, \textsc{2-step cod} may outperform \textsc{1-step cod}. 

We conducted additional simulation studies with other competing methods namely the model based tensor clustering methods \textsc{deem}, \textsc{tgmm} and \textsc{temm} in \cite{mai2022doubly} and \cite{deng2022tensor}. The results and ensuing discussion can be found along with the main simulation results with different noise variance settings in Section \ref{app_sim} of the Supplementary Material. 

\begin{figure}
\begin{center}
\includegraphics[width=\linewidth]{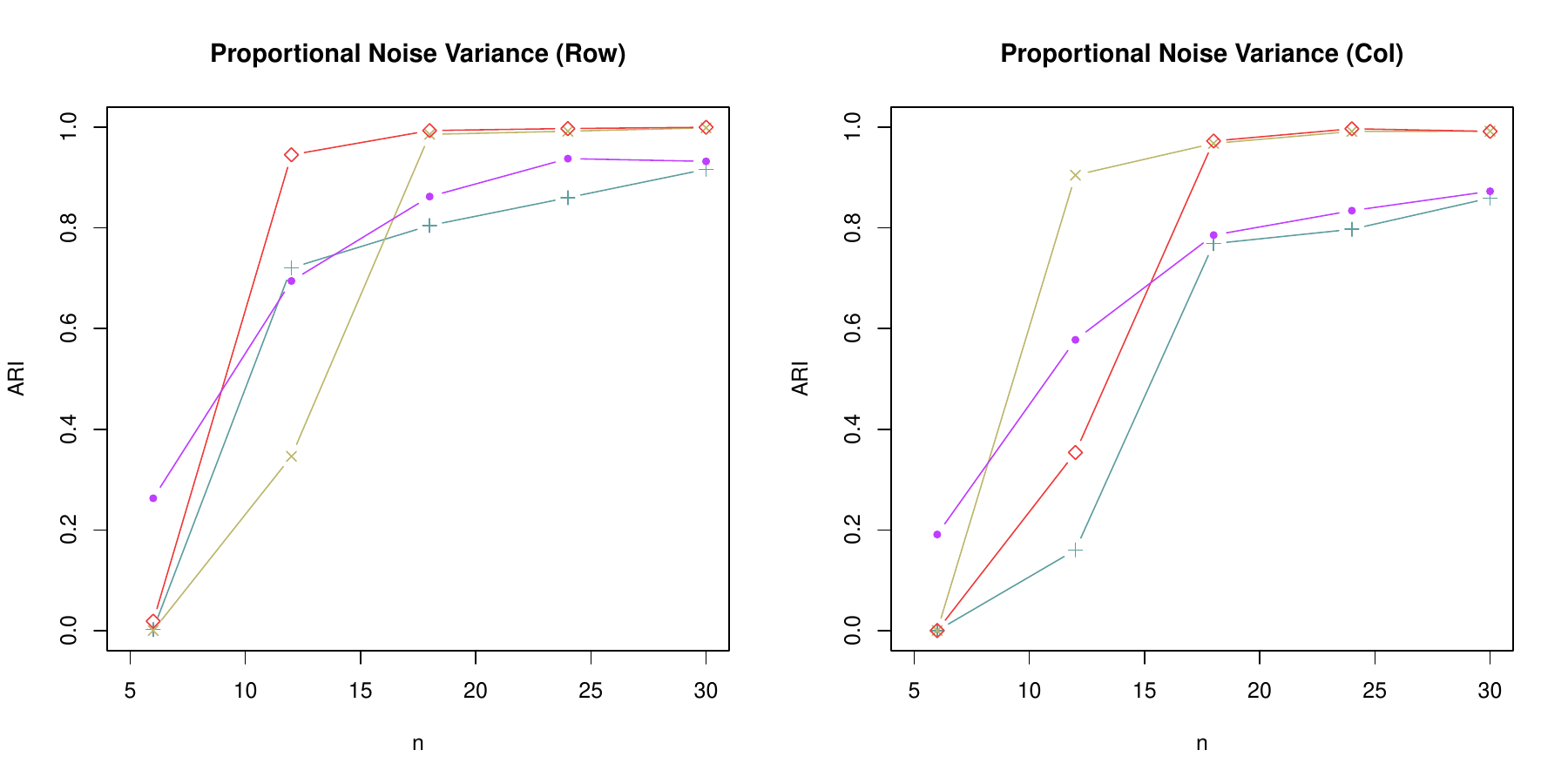}
\end{center}
\caption{A comparison of the adjusted Rand index (ARI) for row clustering (left) and column clustering (right). Our hierarchical clustering Algorithm \ref{alg_1} with $W_I$ (\textsc{naive cod}, $\color{teal}{-+-}$), \textsc{1-step cod} ($\color{olive}{-\times-}$), %with \ref{alg_2} 
    \textsc{2-step cod} ($\color{purple}{-\diamond-}$), %\ref{alg_3}
    and the competing method, \textsc{vanilla hierarchical} ($\color{violet}{-\bullet-}$).}
\label{fig_sim_main}
\end{figure}

\section{Genomic Data Analysis} \label{sec_real}
We apply our method to the atlas of gene expression in the mouse aging project dataset \citep{AGEMAP}, which contains gene expression values of 8932 genes in 16 tissues for 40 mice. Similar to \cite{Bigraphical} and \cite{BioRef2}, we only focus on a subset of the data belonging to the mouse vascular endothelial growth factor signaling pathway. In order to maximize our usage of the tissue data, we drop 4 samples that have missing data for the tissues. So, in the end, our data is a $37 \times 12 \times 36$ array that corresponds to $p=37$ genes, $q=12$ tissues, and $n=36$ mice.

\begin{figure}
\centering
\includegraphics[width=0.7\linewidth]{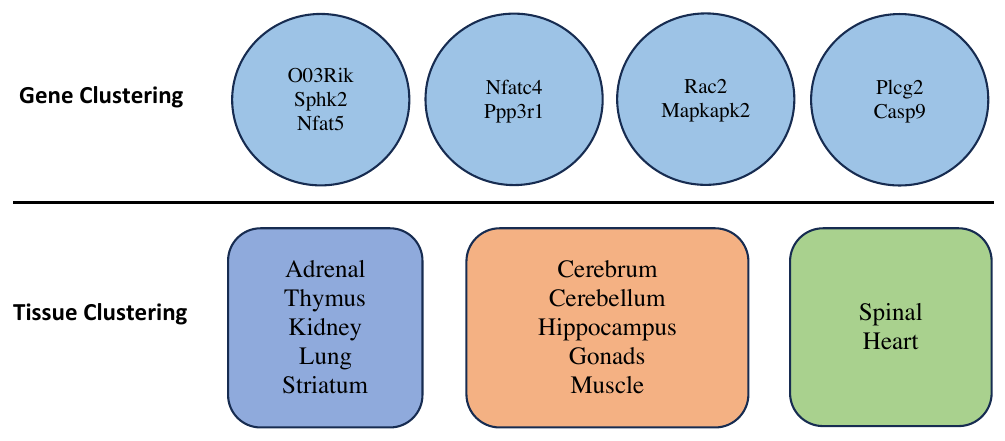}
\caption{Gene clusters and tissue clusters obtained from our Algorithm \ref{alg_3}.} \label{fig:cod}
\end{figure}

Figure \ref{fig:cod} shows the gene clusters and tissue clusters obtained from our two-step Algorithm \ref{alg_3} with a data-driven tuning parameter. Disregarding the singletons - clusters with only one element - the estimated number of gene clusters is 4 and the estimated number of tissue clusters is 3. In \cite{AGEMAP}, the authors group the tissues through hierarchical clustering, but they group genes that are similarly age-regulated through empirical meta analysis. Our method, on the other hand, can conveniently be applied to both the genes and the tissues at the same time. Interestingly, the tissue clustering result from our method agrees with the tissue clustering result in \cite{AGEMAP} which classifies tissues into 3 groups - vascular, neural, and steroid responsive. The ``Cerebrum", ``Cerebellum", and ``Hippocampus" are all neural tissues that are parts of the brain, and they are all clustered into one tissue cluster by our algorithm. Also, ``Adrenal" and ``Thymus" tissues are steroid responsive tissues, while the ``Lung" and ``Kidney" tissues are vascular tissues. The respective pairs appear in our clustering result as well. As for gene clustering, it is known in the biology community that the O03Rik gene inhibits the Nfat5 gene. More specifically, O03Rik acts as a negative regulator of the calcineurin/NFAT signaling pathway and it inhibits NFAT nuclear translocation and transcriptional activity by suppressing the calcium-dependent calcineurin phosphatase activity \citep{origene}. It is then reasonable that the two genes are clustered together, because they would be highly negatively correlated. Also, calcineurin (Ppp3) is a serine/threonine protein phosphatase that is dependent on calcium and calcium modulated proteins \citep{hogan2003transcriptional}. It activates nuclear factor of activated T cell cytoplasmic (Nfatc), a transcription factor, by dephosphorylating it. This Nfatc happens to be encoded in part by the Nfatc4 gene. On the other hand, there are three isozymes of the catalytic subunit of calcineurin and this is encoded in part by the Ppp3r1 gene \citep{liu2005characterization}. Thus, the Ppp3r1 and Nfatc4 genes are both related to calcineurin - one having to do with how it is generated, and one having to do with something it activates. This reasonably explains the highly correlated results obtained from our clustering algorithm. Another possible explanation is given in \cite{heit2006calcineurin} which postulates that calcineurin/NFAT signaling is critical in $\beta$-cell growth in the pancreas. All this external evidence suggests that our clustering results are biologically meaningful. We also have cluster results when applying a competing method \texttt{DEEM} \citep{deng2022tensor} on the same data. The cluster results along with a comparative discussion are presented in Section \ref{real_data_deem} of the Supplementary Material.

\section{Discussion}\label{sec_discussion}
In this paper, we study the variable clustering problem for matrix valued data $X$ under the latent variable model (\ref{eq_model}). A class of hierarchical clustering algorithms based on the weighted covariance difference is proposed. The theoretical and empirical performance of the algorithm heavily depends on the weight matrix $W$. Theoretically, we show that under mild conditions, our algorithm with a large class of weight matrices $W$ can attain clustering consistency with high probability. To further characterize the effect of the weight matrix $W$ on the theoretical performance of our algorithm, we establish the minimax lower bound under the latent variable model (\ref{eq_model}), from which we prove that our algorithm using the weight matrix $W_O$ is minimax optimal for variable clustering. In particular, we apply the theory to the more concrete matrix normal model and show that clustering consistency and minimax optimality can be achieved in practice with an implementable initial weight, $\hat{W}_{O,(r)}=\frac{1}{q}\bar{B}(\bar{B}^T\bar{B})^{-2}\bar{B}^T=\frac{1}{q}I_q$. Empirically, we develop iterative one-step and two-step algorithms based on the weight matrix $W_O$, which outperform competing methods.

While we introduced extensions of our latent variable model and algorithms in Section \ref{sec_extension}, it would be interesting to further investigate each topic on its own. In particular, the theoretical effect of using different weight matrices when clustering higher order tensors with $\COD$ based methods would be an interesting direction. Another interesting extension would be to generalize our method to the overlapping clustering setting \citep{bing2020adaptive}, where the rows and columns may simultaneously belong to multiple row and column clusters. For now, however, we leave them for future study.

\section*{Acknowledgment}
Ning is supported in part by National Science Foundation (NSF) CAREER award DMS-1941945, NSF award DMS-1854637 and NIH 1RF1AG077820-01A1.

\section*{Supplementary Material}
The supplement consists of additional technical details, derivations and discussions. Relevant proofs and additional numerical results are also included.

\bibliography{main}

\newpage
\appendix
\noindent \textbf{\huge Supplementary Material}
%We use the following notation: $[t]$ denotes the t-th column cluster (a set), and $m_q^{(t)}=|[t]|$ denotes the cardinality of the t-th column cluster. Likewise, $m_p^{(t)}$ denotes the cardinality of the t-th row cluster. In addition, $M_q = \max_{1 \leq t \leq K_2}|[t]|$ denotes the maximum column cluster size and $m_q = \min_{1 \leq t \leq K_2}|[t]|$ denotes the minimum column cluster size.

\section{Comparison with the Vectorized Approach in \cite{clustering}}
\label{vectorized}

\begin{figure}
\begin{center}
\includegraphics[width=1\linewidth]{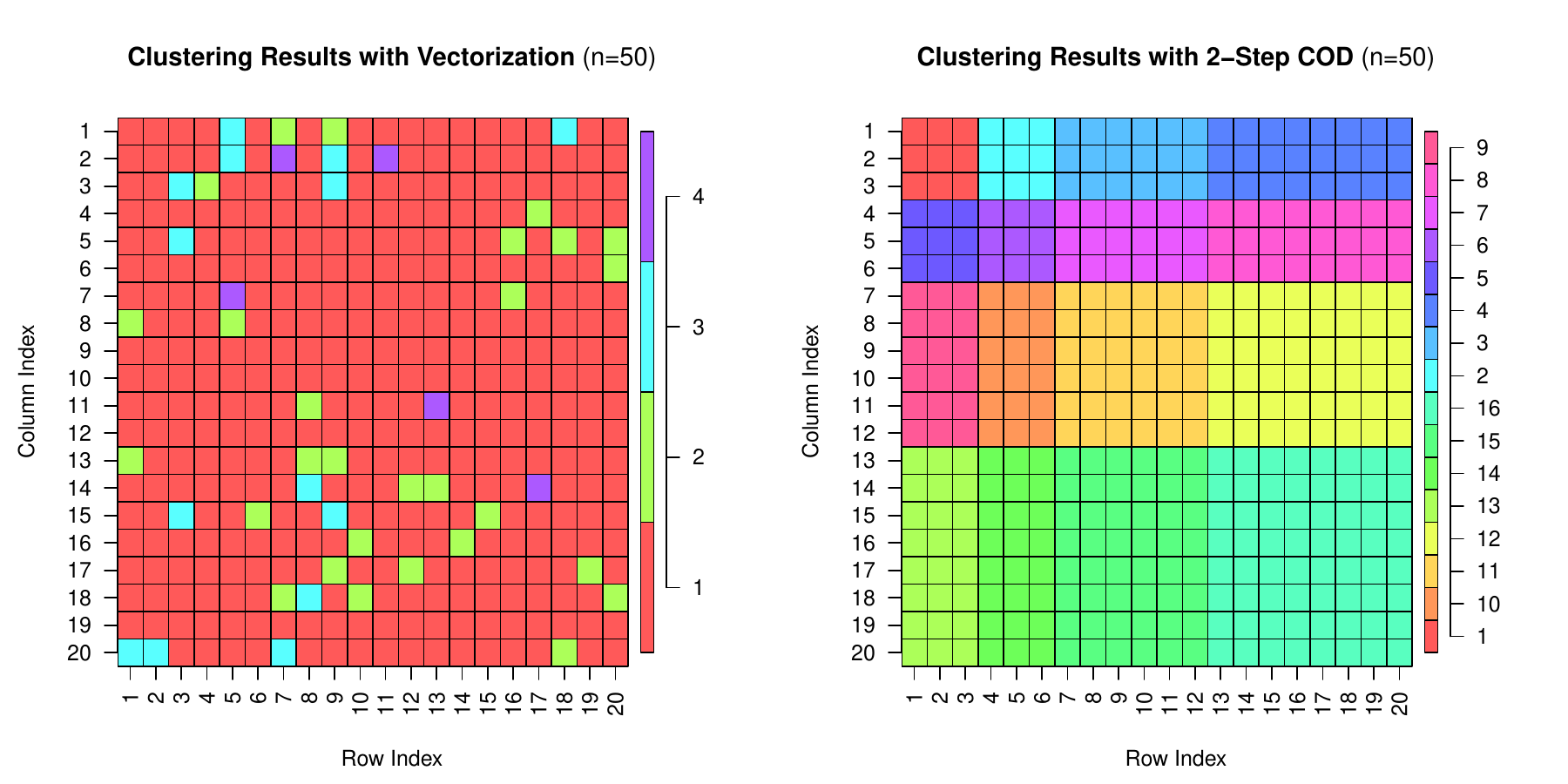}
\includegraphics[width=1\linewidth]{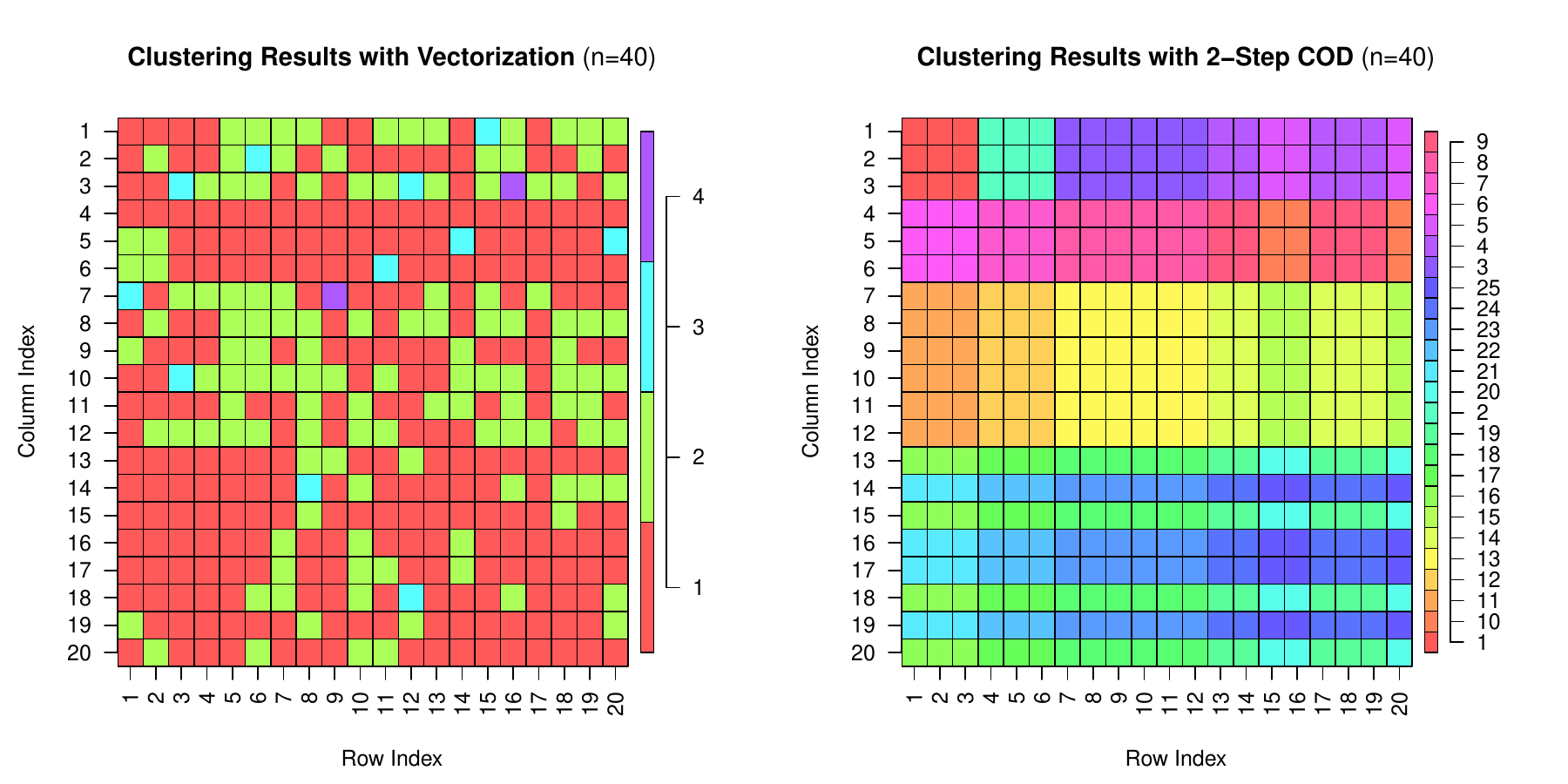}
\caption{$p=q=20$, $K_1=K_2=4$, and the row and column cluster sizes are $(3,3,6,8)$ each. The data was generated with the matrix normal distribution with a decaying Toeplitz matrix for the row and column covariance matrices and a proportional noise variance setting. The left matrices are the cluster results when using \texttt{cord} [\cite{clustering}] on the vectorized matrix, and the right matrices are the results when using our $\TWOSTEPCOD$ method. When $n=50$, our method perfectly recovers the block cluster structure, but even when the results are imperfect ($n=40$), the row and column clusters are still easily interpretable, unlike the left results, which has cluster elements scattered throughout the matrix without additional structure.}
\label{comparison_cord}
\end{center}
\end{figure}

Model (\ref{eq_model}) can be viewed as the extension of the G-block model for clustering a random vector in \cite{clustering} to matrix valued data. If we vectorize the matrix $X$, model (\ref{eq_model}) is equivalent to $\textrm{vec}(X)=M\textrm{vec}(Z)+\textrm{vec}(\Gamma)$ with $M=B\otimes A$, where $\textrm{vec}(X)$ denotes the vectorization of $X$, formed by stacking the columns of $X$ into a single column vector, and $\otimes$ denotes the Kronecker product. Thus, compared to \cite{clustering} which allows $M$ to be any unstructured $(pq)\times (K_1K_2)$ membership matrix, we impose the Kronecker product structure to the membership matrix $M$. While our model for $\textrm{vec}(X)$ is more restrictive than \cite{clustering}, it actually comes with two advantages for matrix clustering. First, as seen above, $A$ and $B$  are interpreted as the membership matrices for the rows and columns. Ignoring the Kronecker product structure and directly applying the model in \cite{clustering} would no longer produce interpretable results for matrix clustering, as shown in Figure \ref{comparison_cord}. Second, the Kronecker product of $A$ and $B$ provides a more parsimonious parametrization for the unknown membership matrix $M$, leading to stronger theoretical guarantees on clustering. %In particular, we show that under our model (\ref{eq_model}) the cluster separation condition required for clustering consistency is weaker than that required in \cite{clustering}. 

\section{Comparison of the Proposed Hierarchical Algorithm with the Algorithm in \cite{clustering}}\label{app_compare}

\begin{figure}[ht!]
\centering
\subfigure[]{\includegraphics[scale=0.69]{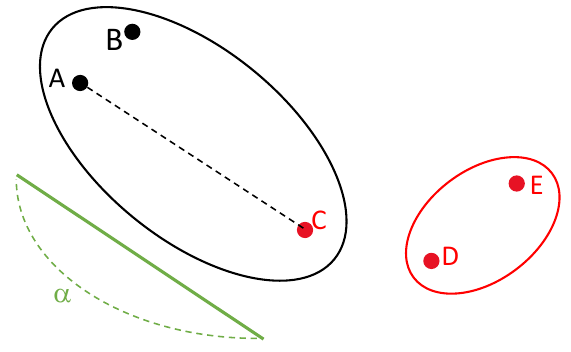}}~~~~
   \subfigure[]{\includegraphics[scale=0.69]{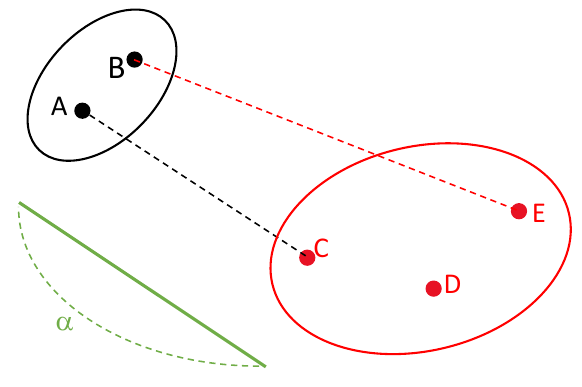}}
\caption{(a) Due to the nature of the algorithm from \cite{clustering}, A and B are closest together and thus will be clustered before C,D,E are considered. Then, since the distance between A and C is below the threshold value $\alpha$, C will be incorrectly clustered with A and B. (b) With the hierarchical algorithm, the distances between all the points are considered at the same time and thus C will be correctly clustered with D and E.}
\label{fig:prev}
\end{figure}

Like shown in Figure \ref{fig:prev}, if C,D,E have not been clustered yet and A and B are  considered first, due to the nature of the algorithm in \cite{clustering}, since the distance between C and A is less than $\alpha$, C will be clustered with A and B even though C,D,E are closer to each other. If we implement a hierarchical approach, the distances between all the pairs of points will be considered at the same time, and a tree structure will naturally be constructed. The two groups of \{A,B\} and \{C,D,E\} will be formed in the \say{lower} part of the tree first, and even if the same $\alpha$ threshold value is used,  the two groups will not be merged together.

\section{Conditions for $||X||_W =O(1)$}\label{app_norm}
In this section, we establish the conditions for $||X||_W=O(1)$ for a general class of weight matrices.
For brevity, we will discuss the row clustering case. Since our method is focused on using the columns to cluster the rows, we will focus on the class of weight matrices that are generated by arbitrary column membership matrices $\bar{B}$:
$$ \mathbf{W}_{(r)}(q) := \bigg\{ \bar W =\frac{1}{s}\bar{B}(\bar{B}^T\bar{B})^{-2}\bar{B}^T \bigg|~ \bar{B} \text{ is a membership matrix} \in \RR^{q \times s} \bigg\},$$
where $s$ is the number of clusters implied by $\bar B$. It can be shown that
\begin{align*}
    ||X||_{\bar{W}} ~&=~ c \cdot \frac{\sqrt{K_2}}{s} \cdot \underset{1 \leq a \leq p}{\max}\sqrt{\Big|\Big|G^TG\Big|\Big|_F^2 + \sum_{r=1}^s\Bigg(\frac{\sum_{j \in \bar{[r]}} \sigma^2_{aj}}{|\bar{[r]}|^2}\Bigg)^2}\\
    ~&\leq~ c \cdot \frac{\sqrt{K_2}}{s} \cdot \sqrt{\Big|\Big|G^TG\Big|\Big|_F^2 + s\cdot\Bigg(\frac{\sigma^2_{\max}}{\bar{m}_q}\Bigg)^2},
\end{align*}
where $G=B^T\bar{B}(\bar{B}^T\bar{B})^{-1}$ and $\bar{m}_q$ denotes the smallest column cluster size implied by $\bar B$. So, 
$$    ||X||_{\bar{W}}~\lesssim~ O(1) \hspace{0.5cm}\text{ if both }~~
\begin{cases}
    ~~||G^TG||_F^2 ~&\lesssim~~~ \frac{s^2}{K_2}\\
    ~~\sigma^2_{\max} ~&\lesssim~~~ \bar{m}_q \cdot \sqrt{\frac{s}{K_2}}
\end{cases} ~~~~\text{hold}.$$ 
Note that when $\bar{B}=B$, then $G=I_s=I_{K_2}$ and the condition for $||G^TG||_F^2$ is automatically satisfied. We expect that the condition is also satisfied for small perturbations $\bar{B} \approx B$, where $s\asymp K_2$. %The conditions on the maximum noise variance $\sigma^2_{\max}$ are also similar to the conditions mentioned for $W_I$ and $W_O$.

\section{Derivation for Remark 3.2}\label{app_remark}
We consider the setting where the column cluster has equal size (i.e., $K_2M_q=q$) and $\sigma_{ab}^2=\sigma^2$. First, we have shown in Section \ref{sec_optimal} that the off diagonal entries of $\Sigma_{p,W_I}$ and $\Sigma_{p,W_O}$ are the same, that is $(\Sigma_{p,W_I})_{jk}=(\Sigma_{p,W_O})_{jk},~ j\neq k$. Then we have $\MCOD(\Sigma_{p,W_I})=\MCOD(\Sigma_{p,W_O}):=\MCOD$. For $W_I$, we can show that 
\begin{align*}
  \frac{\MCOD(\Sigma_{p,W_I})}{ \|X\|_{W_I}}&\geq \frac{\MCOD}{\sqrt{\frac{1}{K_2}}\max_k||\Var(Z_{k\cdot})||_F+\sqrt{\frac{K_2}{q}}\sigma^2}\\
  &\geq C\min \bigg(\frac{\MCOD}{\sqrt{\frac{1}{K_2}}\max_k||\Var(Z_{k\cdot})||_F},~~ \frac{\MCOD}{\sqrt{\frac{K_2}{q}}\sigma^2}\bigg),
\end{align*}
where $C$ is a constant. Hence, the cluster separation condition (A1) for $W_I$ is implied by
$$
\frac{\MCOD}{\sqrt{\frac{1}{K_2}}\max_k||\Var(Z_{k\cdot})||_F}\geq c_0/C \eta, ~~\frac{\MCOD}{\sqrt{\frac{K_2}{q}}\sigma^2}\geq c_0/C \eta.
$$
Similarly, the cluster separation condition (A1) for $W_O$ is implied by
$$
\frac{\MCOD}{\sqrt{\frac{1}{K_2}}\max_k||\Var(Z_{k\cdot})||_F}\geq c_0/C \eta, ~~\frac{\MCOD}{{\frac{K_2}{q}}\sigma^2}\geq c_0/C \eta.
$$
In terms of the noise level $\sigma^2$, the condition for $W_I$ is $\sigma^2=O(\frac{\MCOD\sqrt{q}}{\eta\sqrt{K_2}})$, whereas the condition for $W_O$ is $\sigma^2=O(\frac{\MCOD \cdot q}{\eta K_2})$. The latter is clearly much weaker. Thus, when $q\gg K_2$, our algorithm with the optimal weight $W_O$ attains clustering consistency in the presence of a larger noise level $\sigma^2$. This is the benefit of using the optimal weight in our algorithm. 

\section{Minimax Lower Bound with a Perturbed $\bar{B}$}
\label{general_lower_bound}
Here we present a lower bound result when the cluster separation metric in the definition of the parameter space is defined based on an arbitrary column membership matrix $\bar B$. For example, the special cases $\bar B=I_q$ and $\bar B=B$ are considered.

Assume that we have an arbitrary column membership matrix $\bar B$, which is fixed and can be different from the true membership matrix $B$. Let $\bar W=\bar B(\bar B^T\bar B)^{-2}\bar B^T/s$ where $s$ denotes the number of clusters implied by $\bar B$. We define the following parameter space
\begin{equation*}
    M_{\bar W}=\big\{ \Sigma\in \RR^{pq\times pq}|~X~\textrm{satisfies model}~(1.1), ~~\MCOD(\Sigma_{p,\bar W})/\|X\|_{\bar W}\geq \eta\big\},
\end{equation*}
We now present a general lower bound for clustering over the parameter space $M_{\bar W}$.
\begin{theorem} (Minimax Lower Bound with a Perturbed $\bar{B}$) \label{gen_minimax}\\
For $K_1 \geq 3$, there exists a positive constant $c$ such that for any $\eta$ 
\begin{align}\label{eq_thm_lowerbound_1_eta_general}
    0 ~~\leq~~ \eta ~~<~~ c\sqrt{\frac{\log p}{nK_2}} \cdot \frac{||G||_F^2}{\sqrt{K_2} \cdot \sqrt{\big|\big| G^TG \big|\big|_F^2 + \Big(\frac{q}{K_2}\Big)^2 \cdot \frac{s}{\bar{m}_q^2}}}
\end{align}
we have
\begin{align*}
\underset{\widehat{\cG}}{\inf}~\underset{\Sigma \in M_{\bar W}}{\sup}~\PP_{\Sigma}(\widehat{\cG}\neq \cG) ~\geq~ \frac{1}{7},
\end{align*}
where $\bar{m}_q$ denotes the smallest column cluster size implied by $\bar B$ and $G=B^T \bar{B}(\bar{B}^T\bar{B})^{-1}$. The infimum is taken over all possible estimators of $\cG$.
\end{theorem}

This theorem shows that, if we define $\MCOD$ based on a perturbed $\bar B$, the rate of the cluster separation (\ref{eq_thm_lowerbound_1_eta_general}) depends on $G$ and $\bar m_q$. 
%Note that the first term $\sqrt{\frac{\log p}{nK_2}}$ denotes the minimax lower bound when we know the true $B$. It is apparent that the second term becomes of constant order when you plug in $G=I_{K_2}$, $s=K_2$, $\bar{m}_q=m_q=\frac{q}{K_2}$ since $||G^TG||_F^2 = K_2$ and $||G||_F^2=K_2$. Thus, this second term depends on how accurate our estimated column cluster structure was as it is a function of $G$, $s$ and $\bar{m}_q$. 
If $\bar B$ is near perfect, $G \approx I_{K_2}$, $G^TG \approx I_{K_2}$, $s \approx K_2$ and $\bar{m}_q \approx m_q = \frac{q}{K_2}$ in the construction of the lower bound. Then, $||G||_F^2$, $||G^TG||_F^2$ and $(\frac{q}{K_2})^2 \cdot \frac{s}{\bar{m}_q^2}$ will all be close to order $K_2$ and the bound in  (\ref{eq_thm_lowerbound_1_eta_general}) becomes $\sqrt{\frac{\log p}{nK_2}}$. In contrast, if we use $W_I=\frac{1}{q}I_q$, then $G$ is a $K_2 \times q$ matrix with row vectors $1^T_{m_q^{(i)}}$ as blocks on the diagonal. This implies that $G^T G$ is a $q \times q$ matrix with $1_{m_q^{(i)}}1^T_{m_q^{(i)}}$ as square blocks in the diagonal, $||G||_F^2=q$ and $||G^T G||_F^2 = K_2 \cdot (\frac{q}{K_2})^2 = \frac{q^2}{K_2}$ in our construction. The lower bound becomes $\sqrt{\frac{\log p}{nq}}$. These two cases show how the lower bound depends on the imperfect column cluster structure $\bar{B}$. The proof of this lower bound can be found in Section \ref{proof_gen_minimax} of the Supplementary Material.

\section{Supplementary Material for Section \ref{sec_practical}}\label{app_practical}

\subsection{Theoretical Guarantees for Algorithm \ref{alg_2}}
\subsubsection{Consistency} \label{one_step_consistency}
We present the consistency theorem for the iterative one-step Algorithm \ref{alg_2} that clusters both the rows and the columns. 
\begin{theorem} (Consistency with One-step Hierarchical COD with the Optimal Weight)\label{thm_one_step}\\
Under the model $X=AZB^T+\Gamma$, assume that $\textrm{vec}(X)$ is multivariate Gaussian, $\log p=o(n)$, $\log q=o(n)$ and the following conditions hold: 
\begin{itemize}
    \item[(R)] Row Separation Condition: $$\text{MCOD}(\Sigma_{p,W_{O,(r)}})/\|X\|_{W_{O,(r)}}~~>~~c_{0,(r)} \cdot \eta_{(r)} $$ where $c_{0,(r)} \geq 4$ is an arbitrary constant and $~\eta_{(r)} \geq c_{1,(r)} \sqrt{\frac{\log p}{nK_2}}~$ for a universal constant $c_{1,(r)} > 0$.
    \item[(C)] Column Separation Condition: $$\text{MCOD}(\Sigma_{q,W_{O,(c)}})/\|X\|_{W_{O,(c)}}~~>~~c_{0,(c)} \cdot \eta_{(c)} $$ where $c_{0,(c)} \geq 4$ is an arbitrary constant and $~\eta_{(c)} \geq c_{1,(c)} \sqrt{\frac{\log q}{nK_1}}~$ for a universal constant $c_{1,(c)} > 0$.    
    \item[(S)] Stability Condition: 
    $$\bigg[\Big\{\text{MCOD}(\Sigma_{p,\hat {W}_{O,(r)}})\Big\}\Big/\|X\|_{\hat {W}_{O,(r)}}\bigg] \Bigg/ \bigg[\Big\{\text{MCOD}(\Sigma_{p,W_{O,(r)}})\Big\}\Big/\|X\|_{W_{O,(r)}}\bigg]> \frac{4}{c_{0,(r)}}$$ 
    where $c_{0,(r)}$ is defined in condition (R).
  %  \item[(C2)] Column Stability Condition: 
%  $$\bigg[\Big\{\text{MCOD}(\Sigma_{q,\hat {W}_{(c)}})\Big\}\Big/\|X\|_{\hat {W}_{(c)}}\bigg] \Bigg/ \bigg[\Big\{\text{MCOD}(\Sigma_{q,W_{(c)}})\Big\}\Big/\|X\|_{W_{(c)}}\bigg]> \frac{4}{c_{0,(c)}}$$ 
%    where $c_{0,(c)}$ is defined in (C1).
\end{itemize}
Then using our Algorithm \ref{alg_2} with the threshold $\alpha_{(r)}=2\eta_{(r)} \cdot \|X\|_{\hat W_{O,(r)}}$ for the row clustering step and the threshold  $\alpha_{(c)}=2\eta_{(c)} \cdot \|X\|_{\hat W_{O,(c)}}$ for the column clustering step, we obtain perfect cluster recovery (i.e., $\hat\cG^{(r)}=\cG^{(r)}$ and $\hat\cG^{(c)}=\cG^{(c)}$) with probability greater than $1-\frac{c_3}{p}$ for some constant $c_3>0$.
\end{theorem}

The row and column separation conditions (R) and (C) are similar to the cluster separation condition (A1) in Theorem \ref{thm_upper}. The stability condition (S) is only imposed for row clustering, which can be viewed as the condition for the initial value $\bar B$ in Step 1 of Algorithm \ref{alg_2}. As shown in Proposition \ref{pro_stability}, the initial value $\bar B=I_q$ satisfies condition (S) under the matrix normal model with mild additional assumptions. Indeed, by Theorem \ref{thm_upper}, conditions (S) and (R) imply perfect row cluster recovery (i.e., $\hat A=A$) with high probability, which further implies that $\hat W_{O,(c)}=W_{O,(c)}$. Thus, the stability condition for column clustering automatically holds, which in turn guarantees perfect column cluster recovery with high probability. The proof is very similar to the proof for Theorem \ref{thm_upper} and will be omitted for brevity.

\subsubsection{Minimax Optimality} \label{one_step_minimax}
In the following, we establish a lower bound result for both row and column clustering which naturally incorporates the uncertainty in estimating the membership matrices $A$ and $B$ simultaneously. Following the notation used in Theorem \ref{thm_one_step}, we define parameter spaces
\begin{equation*}
    M_{O,(r)}=\big\{ \Sigma\in \RR^{pq\times pq}|~X~\textrm{satisfies model}~(1.1), ~~\text{MCOD}(\Sigma_{p,W_{O,(r)}})/\|X\|_{W_{O,(r)}}\geq \eta_{(r)}\big\},
\end{equation*}
and
\begin{equation*}
    M_{O,(c)}=\big\{ \Sigma\in \RR^{pq\times pq}|~X~\textrm{satisfies model}~(1.1), ~~\text{MCOD}(\Sigma_{q,W_{O,(c)}})/\|X\|_{W_{O,(c)}}\geq \eta_{(c)}\big\}
\end{equation*}
for the rows and columns, respectively. To study the minimax lower bound for both row and column clustering, we consider the parameter space $M_{O,(r)}\cap M_{O,(c)}$. The following theorem provides the lower bound for clustering over this parameter space $M_{O,(r)}\cap M_{O,(c)}$. 

\begin{theorem} \label{thm_lowerbound_row_column}
For $K_1, K_2 \geq 3$, there exists a positive constant $c$ such that, for any $\eta_{(r)}$ and $\eta_{(c)}$ 
\begin{align*}
    0 \leq \eta_{(r)} < c\sqrt{\frac{\log p}{nK_2}}~ ~~\textrm{or}~~    0 \leq \eta_{(c)} < c\sqrt{\frac{\log q}{nK_1}},
\end{align*}
we have
\begin{align*}
\underset{(\widehat{\cG}^{(r)},\widehat{\cG}^{(c)})}{\inf}~~\underset{\Sigma \in M_{O,(r)}\cap M_{O,(c)}}{\sup}~~\PP_{\Sigma}\Big((\widehat{\cG}^{(r)},\widehat{\cG}^{(c)})\neq (\cG^{(r)},\cG^{(c)})\Big) ~~\geq~~ \frac{1}{7},
\end{align*}
where the infimum is taken over all possible estimators of $(\cG^{(r)},\cG^{(c)})$.
\end{theorem}
This theorem shows that we need $\eta_{(r)} \geq c\sqrt{\frac{\log p}{nK_2}}$ {\bf and} $\eta_{(c)} \geq c\sqrt{\frac{\log q}{nK_1}}$ to attain both perfect row and column clustering. Together with Theorem \ref{thm_one_step}, we obtain that, as long as the initial estimate $\bar B$ satisfies the stability condition (i.e., it falls into a contraction region), the One-Step Hierarchical Algorithm with the Optimal Weight (Algorithm \ref{alg_2}) is minimax optimal for row and column clustering. Recall that this has been verified for the matrix normal model with $\bar{B}=I_q$ in Proposition \ref{pro_stability}. Finally, we note that since the above lower bound is concerned with both row and column clustering, the uncertainty in estimating both membership matrices $A$ and $B$ is taken into account. The proof of this theorem is in Section \ref{joint_lower} of the Supplementary Material.

\subsection{Two-Step Hierarchical Algorithm with the Optimal Weight}\label{two-step}
The two-step algorithm is shown in Algorithm \ref{alg_3}. In this algorithm, we ignore the sample splitting step for simplicity. 

\begin{algorithm}
\caption{Two-step Hierarchical Algorithm with the Optimal Weight}
\begin{enumerate}
    \item[(0a)] Set the initial value $\bar{B} = I_q$.
    \item[(0b)] Apply Algorithm \ref{alg_1} with $\hat\Sigma_{p,\hat W_{O_1,(r)}}=\frac{1}{n}\sum_{i =1}X^{(i)}\hat W_{O_1, (r)} X^{(i)T}$ to cluster the \\
    rows of $X$, where $\hat W_{O_1,(r)}=\bar B(\bar B^T\bar B)^{-2}\bar B^T/s$ and $s$ denotes the estimated number\\
    of clusters from $\bar B$. Obtain the resulting row cluster $\hat\cG^{(r)}$ or equivalently the membership matrix $\hat A_1$.
    \item[(1a)] Compute the estimate of the optimal column weight $\hat {W}_{O,(c)}=\hat A_1(\hat A_1^T\hat A_1)^{-2}\hat A_1^T/t$ where $t$ is the estimated number of row clusters in $\hat A_1$.
    \item[(1b)] Apply Algorithm \ref{alg_1} with $\hat{\Sigma}_{q,\hat{W}_{O,(c)}} = \frac{1}{n}\sum_{i=1}^nX^{(i)T}\hat{W}_{O,(c)} X^{(i)}$ to cluster the columns\\
    of $X$ and find the estimator $\hat B_1$.
    \item[(2a)] Compute the estimate of the optimal row weight $\hat W_{O_2,(r)}=\hat B_1(\hat B_1^T\hat B_1)^{-2}\hat B_1^T/s_1$,\\
    where $s_1$ denotes the estimated number of column clusters in $\hat{B}_1$.
    \item[(2b)] Apply Algorithm \ref{alg_1} with $\hat\Sigma_{p,\hat W_{O_2,(r)}}=\frac{1}{n}\sum_{i=1}^n X^{(i)}\hat W_{O_2,(r)} X^{(i)T}$ to cluster the rows\\
    of $X$. Obtain the final estimator $\hat A_2$.
\end{enumerate}\label{alg_3}
\end{algorithm}

\subsection{Simulation Results With and Without Sample Splitting} \label{data_split}
We present the results for an additional simulation study that highlights the difference of our method with and without sample splitting in Table \ref{table_data_split}. We have $n$ i.i.d. copies of $30 \times 30$ matrices with $K_1=K_2=4$ and moderately unbalanced cluster sizes of $4,6,9,11$ for both the rows and columns. The decay rate for the Toeplitz covariance matrices is -0.2 and 0.2 for the rows and columns, respectively. We consider the proportional noise variance setting from the main paper. We vary $n$ from $20$ up to $100$ in increments of $20$ and the ARI values for $\TWOSTEPCOD$ are recorded. We see that, for $n$ relatively small (say $n\leq 60$), our $\TWOSTEPCOD$ without data splitting performs significantly better than the method using data splitting. In addition, both methods yield very high clustering accuracy when $n$ is large enough (say $n\geq 100$). Thus, in practice, we recommend using our $\TWOSTEPCOD$ without data splitting, especially when the sample size is small or moderate. 
\begin{table}
\centering
\resizebox{\columnwidth}{!}{%
\begin{tabular}{|c|c||c|c||c|c|}
\hline
\hspace{0.5cm}& \hspace{0.5cm} &\multicolumn{1}{c|}{Data Split (Row)} &\multicolumn{1}{c||}{No Data Split (Row)} & \multicolumn{1}{c|}{Data Split (Col)}& \multicolumn{1}{c|}{No Data Split (Col)}\\
\hline
\hline
\multirow{2}{*}{$n$} & 20 & 0 & 0.4984 & 0 & 0.2723\\
     \cline{2-6}
     & 40 & 0.0639 & 0.9939 & 0.0712 & 0.9562 \\
     \cline{2-6}
     & 60 & 0.5642 & 1 & 0.1647 & 0.9979 \\
     \cline{2-6}
     & 80 & 0.9685 & 1 & 0.6834 & 0.9934 \\
     \cline{2-6}
     & 100 & 0.9849 & 1 & 0.9528 & 0.9962 \\
\hline
\end{tabular}
}
\caption{The ARI values obtained from the $\TWOSTEPCOD$ Algorithm with $p=q=30$, $K_1=K_2=4$, moderately unbalanced cluster sizes with proportional noise variance under the $n=20,40,60,80,100$ setting with and without data splitting.}
\label{table_data_split}
\end{table}

\subsection{Data-Driven Tuning Parameter Selection Process for $\alpha$}\label{tuning}

In the following, we describe a data-driven selection method for $\alpha$ that was also used in \cite{clustering}. To fix the notation, we consider our Algorithm \ref{alg_1} with some sample covariance matrix $\hat{\Sigma}_p$. We assume the data are standardized. The method can be similarly applied to $\ONESTEPCOD$, $\TWOSTEPCOD$ with the optimal weight and $\NAIVE \COD$. The steps are outlined in Algorithm \ref{alg_tuning}.

We first split the data into two, $D_1$, $D_2$, and calculate $\hat{\Sigma}_{p}^{(1)}$, $\hat{\Sigma}_{p}^{(2)}$, respectively. For each tuning parameter $\alpha_l$ in the grid, we perform our algorithm on $\hat{\Sigma}_{p}^{(1)}$ to get a cluster structure $\cG_l$. We then take the average of all the non-diagonal elements in the cluster blocks of $\hat{\Sigma}_{p}^{(1)}$ via the smoothing operator $\Upsilon(\hat{\Sigma}_{p}^{(1)},\cG_l)$. Finally we calculate the Frobenius loss of $\Upsilon(\Sigma_{p}^{(1)},\cG_l)$ and $\Sigma_{p}^{(2)}$, and choose $\alpha_l$ that yields the smallest value. 

\begin{algorithm}
\caption{A Data-Driven Tuning Parameter Selection Process}
\begin{enumerate}
    \item Split the data into two: \(D_1\) and \(D_2\)
    \item Using \(D_1\), calculate \(\hat{\Sigma}_{p}^{(1)}\).
    \item Using \(D_2\), calculate \(\hat{\Sigma}_p^{(2)}\).
    \item For \(r > 1\) and each value \(\alpha_l\) on a grid   $(l=1,...,r)$, perform Algorithm \ref{alg_1} with \(\hat{\Sigma}_{p}^{(1)}\) \\to get a row cluster structure $\cG_l^{(r)}$.
    \item Perform the smoothing operator $\Upsilon(\hat{\Sigma}_{p}^{(1)},\cG_l^{(r)})$ where $\Upsilon$ is defined as the following:
    \[ [\Upsilon(\hat{\Sigma}_{p},\cG^{(r)})]_{ab} = \begin{cases} 
      \Big|G_k^{(r)}\Big|^{-1}\bigg(\Big|G_k^{(r)}\Big|-1\bigg)^{-1}\underset{i,j \in G_k, ~ i \neq j}{\sum}\big[\hat{\Sigma}_{p}\big]_{ij} & \text{if}~ a \neq b, ~ \text{and} ~ k = k' \\
     \Big|G_k^{(r)}\Big|^{-1}\Big|G_{k'}^{(r)}\Big|^{-1}\underset{i \in G_k, ~ j \in G_{k'}}{\sum} \big[\hat{\Sigma}_{p}\big]_{ij} & \text{if}~ a \neq b, ~ \text{and} ~ k \neq k'\\
      1 & \text{if}~a=b. 
   \end{cases}
    \]
    \item Our data dependent tuning parameter for the threshold is:
    \[ \hat{\alpha} = \underset{\alpha_l}{\mathrm{argmin}}~L\bigg(\Upsilon(\hat{\Sigma}_p^{(1)},\cG_l^{(r)}),~\hat{\Sigma}_p^{(2)}\bigg),\] 
    where $L(A,B):=||A-B||_F$ is the Frobenius loss.
\end{enumerate}\label{alg_tuning}
\end{algorithm}

\section{Supplementary Material for Section \ref{sec_extension}} \label{supp_extension}
\subsection{The Dependent Noise Model} \label{supp_dependent_noise}
\subsubsection{Discussion of Theorem \ref{correlated_consistency}}
Under model (\ref{eq_model}) with dependent noise elements, since 
$\Sigma_{p,W} = \EE(AZB^TWBZ^TA^T) + \EE(\Gamma W\Gamma^T)$, if $~a \underset{\cG}{\sim} b$,
\begin{align*}
     \COD_{\Sigma_{p,W}}(a,b) ~~ &= ~~ \underset{c \neq a,b}{\max}~ \Big| \big[\Sigma_{p,W}\big]_{ac} - \big[\Sigma_{p,W}\big]_{bc} \Big| \\
     ~~&= ~~ \underset{c \neq a,b}{\max}~ \Big|\big[\EE(AZB^TWBZ^TA^T)\big]_{ac} - \big[\EE(AZB^TWBZ^TA^T)\big]_{bc} \\
     ~~ & ~~\hspace{5.5cm} + 
     \big[\EE(\Gamma W \Gamma^T)\big]_{ac} - 
     \big[\EE(\Gamma W \Gamma^T)\big]_{bc}\Big|\\
     ~~&= ~~ \underset{c \neq a,b}{\max}~ \Big|\big[\EE(\Gamma W \Gamma^T)\big]_{ac} - 
     \big[\EE(\Gamma W \Gamma^T)\big]_{bc}\Big|\\
     ~~ &\leq ~~ \gamma(\Sigma_{p,W})
\end{align*}
and if $~a \underset{\cG}{\not\sim} b$,
\begin{align*}
     \COD_{\Sigma_{p,W}}(a,b) ~~ &= ~~ \underset{c \neq a,b}{\max}~ \Big|\big[\EE(AZB^TWBZ^TA^T)\big]_{ac} - \big[\EE(AZB^TWBZ^TA^T)\big]_{bc} \\
     ~~ & ~~\hspace{5.5cm} + 
     \big[\EE(\Gamma W \Gamma^T)\big]_{ac} - 
     \big[\EE(\Gamma W \Gamma^T)\big]_{bc}\Big|\\
     ~~ &\geq ~~ \underset{c \neq a,b}{\max}~ \Big|\big[\EE(AZB^TWBZ^TA^T)\big]_{ac} - \big[\EE(AZB^TWBZ^TA^T)\big]_{bc}\Big| \\
      ~~ &~~\hspace{5.5cm} - \underset{c \neq a,b}{\max}~ \Big|\big[\EE(\Gamma W \Gamma^T)\big]_{ac} - \big[\EE(\Gamma W \Gamma^T)\big]_{bc}\Big |\\
      ~~&\geq ~~ \text{MCOD*}(\Sigma_{p,W}) ~~ - ~~ \underset{c \neq a,b}{\max}~ \Big|\big[\EE(\Gamma W \Gamma^T)\big]_{ac} - \big[\EE(\Gamma W \Gamma^T)\big]_{bc}\Big |\\
      ~~& \geq~~ \text{MCOD*}(\Sigma_{p,W}) ~~ - ~~ \gamma(\Sigma_{p,W}),
\end{align*}
where $\MCOD^{*}(\Sigma_{p,W}) = \underset{a \underset {\cG}{\not\sim} b}{\min}~ \underset{c \neq a,b}{\max}~ \Big|\big[\EE(AZB^TWBZ^TA^T)\big]_{ac} - \big[\EE(AZB^TWBZ^TA^T)\big]_{bc}\Big|$. In other words, $\MCOD^{*}$ is the counterpart to $\MCOD$ that only considers the signal (the first component) from $\Sigma_{p,W} = \EE(AZB^TWBZ^TA^T) + \EE(\Gamma W\Gamma^T)$. 

Comparing the population quantity $\COD_{\Sigma_{p,W}}(a,b)$ in the two cases $~a \underset{\cG}{\sim} b$ and $~a \underset{\cG}{\not\sim} b$, it is apparent that if the signal term is strong enough in the following sense
$$\MCOD^{*}(\Sigma_{p,W}) ~~ > ~~ 2 \cdot \gamma(\Sigma_{p,W}),$$ 
then even when the noise variables are dependent, the $\COD$ measures are well separated and on the population level, the $\COD$ method is still viable for clustering.

The proof of this extended model can be constructed using the above logic and the proof from Theorem \ref{thm_upper} which is in Section \ref{proof_upper} of the Supplementary Material. The remaining parts will be omitted for brevity. 

\subsubsection{A Slightly Different Measure, $\gamma_s(\Sigma_{p,W})$}
We can swap out the definition of $\gamma(\Sigma_{p,W})$ in (\ref{gamma_def}) for a slightly more restrictive but more intuitive defintion:
\begin{align}
    \gamma_s(\Sigma_{p,W}) ~:=~ 2 \underset{1 \leq a \neq b \leq p}{\max}~~\Big| \big[\EE(\Gamma W\Gamma^T)\big]_{ab} \Big|
\end{align}
This $\gamma_s(\Sigma_{p,W})$ can be used in place of the $ \gamma(\Sigma_{p,W})$ in the above theorem since the former is an upper bound for the latter. $\gamma_s(\Sigma_{p,W})$ gives information on the largest off-diagonal entry in $\EE(\Gamma W \Gamma^T)$, i.e. the largest weighted covariance between two \textit{different} noise variables. The relationship between this weighted covariance and the actual covariance between noise variables depends on the weight. For example, with the optimal weight $W_O$, $ \gamma_s(\Sigma_{p,W_O})$ can be upper bounded with $2\cdot \sigma_{\text{offmax}}^2$, where $\sigma_{\text{offmax}}^2 := \underset{(i,j)\neq(i',j')}{\max}\text{Cov}(\Gamma_{ij},\Gamma_{i'j'})$, the maximum of the unweighted covariance between two different noise variables.

\subsection{Nested Clustering to Incorporate Mean Information} \label{supp_mean}
\subsubsection{The Generalized Model}
To generalize the proposed method to account for both mean and covariance information in clustering, we extend our latent variable model to 
\begin{align}
X=M+AZB^T+\Gamma, \label{model_mean_cov}
\end{align}
where $M=\EE(X)$, $Z$ and $\Gamma$ are mean 0 random matrices.  Like before, $AZB^T$ induces the row and column clustering structures based on the covariance of $X$. Specifically, $A\in \RR^{p\times K_1}$ and $B\in \RR^{q\times K_2}$ are the unknown membership matrices for the rows and columns, respectively. We define the row clusters as 
\begin{equation}\label{eq_cluster_cov}
\cG^{(r),[2]}=\{G_1^{(r),[2]},...,G_{ K_1}^{(r),[2]}\}, ~~\textrm{where}~~ G_k^{(r),[2]}=\{a: A_{ak}=1\}
\end{equation} 
for any $1\leq k\leq K_1$. The column clusters $\cG^{(c),[2]}$ can be defined similarly. To incorporate the mean information, we further assume that the matrix $M$ induces the row and column clustering structures based on the mean of $X$. In particular, we assume $M=\tilde A T\tilde B^T$, where $\tilde A\in \RR^{p\times \tilde K_1}$ and $B\in \RR^{q\times \tilde K_2}$ are the unknown membership matrices for the rows and columns based on the mean information, and $T\in \RR^{\tilde K_1\times \tilde K_2}$ is an unknown deterministic matrix. The corresponding row clusters are defined as 
\begin{equation}\label{eq_cluster_mean}
\cG^{(r),[1]}=\{G_1^{(r),[1]},...,G_{\tilde K_1}^{(r),[1]}\}, ~~\textrm{where}~~ G_k^{(r),[1]}=\{a: \tilde A_{ak}=1\}
\end{equation} 
for any $1\leq k\leq \tilde K_1$. The column clusters $\cG^{(c),[1]}$ can be defined similarly. To link these two row clusters $\cG^{(r),[2]}$ and $\cG^{(r),[1]}$, we assume $\cG^{(r),[2]}$ is nested inside $\cG^{(r),[1]}$. \\
\\
\textbf{Definition 1.} \citep{li2010cluster} A clustering $\cG^{[2]}$ with $K'$ clusters is said to be \textbf{nested} inside another clustering $\cG^{[1]}$ with $K$ clusters if:
\begin{enumerate}
    \item (Hierarchical Structure) For any cluster $G^{[2]}_j \in \cG^{[2]}, (1 \leq j \leq K')$, there is a cluster $G^{[1]}_i \in \cG^{[1]}, (1 \leq i \leq K)$ such that $G^{[2]}_j \subseteq G^{[1]}_i$.
    \item (Proper Subset Structure) There exists at least one cluster in $\cG^{[2]}$, (i.e. $G^{[2]}_{j*}$), which satisfies $G^{[2]}_{j*} \subset G^{[1]}_{i*}$ and $G^{[2]}_{j*} \neq G^{[1]}_{i*}$, for some cluster $G^{[1]}_{i*} \in \cG^{[1]}$.
\end{enumerate} 
In other words, the clustering $\cG^{(r),[2]}$ based on the covariance information provides a more refined partition on top of the initial clustering $\cG^{(r),[1]}$ from the mean information. 
For the ease of interpretation, we only consider the case that $\cG^{(r),[2]}$ is nested in $\cG^{(r),[1]}$. This assumption can be relaxed or even removed by defining clusters at different levels (e.g., clustering from the mean and covariance) and new ways of combining the clusters. 

%Under the nested clustering assumption, while one may still possibly recover the clustering structure by applying our COD algorithm to $\Cov(X)$, the mean information may be also useful as a dimension reduction step to stabilize the algorithm especially when the dimension of the feature matrix is unbalanced. In particular, 

\subsubsection{Nested Clustering (Algorithm \ref{alg_5})}

To recover the nested cluster structure, we propose a two-step nested clustering algorithm, in which a mean-based clustering method is implemented first, and then on each cluster, our covariance-based method is applied to capture the finer, more intricate relationships within each broad cluster. 

There are many existing mean-based clustering methods in the literature. Here we use \texttt{SparseBC} from \cite{sparseBC} as the mean-based clustering method. The resulting clusterings are denoted by $\hat{\cG}^{(r),[1]}$ and $\hat{\cG}^{(c),[1]}$. Then, our proposed two-step Algorithm \ref{alg_3} is implemented to get the second layer cluster structures within each first layer cluster. More specifically, for row clustering, we apply the two-step Algorithm \ref{alg_3} to each $ \big|\hat G^{(r),[1]}_j\big|\times q$ submatrix corresponding to the variables in $\hat G^{(r),[1]}_j$ for $1 \leq j \leq \big|\hat \cG^{(r),[1]}\big|$. Similarly, for column clustering, we apply our two-step Algorithm \ref{alg_3} to each $ p \times \big|\hat G^{(c),[1]}_j\big|$ submatrix corresponding to the variables in $\hat G^{(c),[1]}_j$ for $1 \leq j \leq \big|\hat \cG^{(c),[1]}\big|$. In this way, we can use mean information to reduce the dimension of the matrices, to which our two-step Algorithm \ref{alg_3} is applied. Thus, both the mean and the covariance information is utilized in deriving the final clustering result. We summarize this method in Algorithm \ref{alg_5}. 

\begin{algorithm}
    \caption{Mean and Covariance Based Nested Clustering with COD}
    \begin{enumerate}
        \item[(1)] Perform mean-based clustering (\texttt{SparseBC}) to get the first layer cluster structures $\hat{\cG}^{(r),[1]}$ and $\hat{\cG}^{(c),[1]}$
        \item[(2a)] Perform Algorithm \ref{alg_3} on each $\Big|\hat{G}_j^{(r),[1]}\Big| \times q$ submatrix $(1 \leq j \leq \Big|\hat{\cG}^{(r),[1]}\Big|)$
        \item[(2b)] Perform (the col. version of) Algorithm \ref{alg_3} on each $p \times \Big|\hat{G}_j^{(c),[1]}\Big|$ submatrix $(1 \leq j \leq \Big|\hat{\cG}^{(c),[1]}\Big|)$
        \item[(3a)] Combine the row cluster results from each submatrix in (2a) to construct $\hat{A}$
        \item[(3b)] Combine the col. cluster results from each submatrix in (2b) to construct $\hat{B}$
    \end{enumerate}\label{alg_5}
\end{algorithm}

\begin{figure}
\begin{center}
\includegraphics[width=0.5\linewidth]{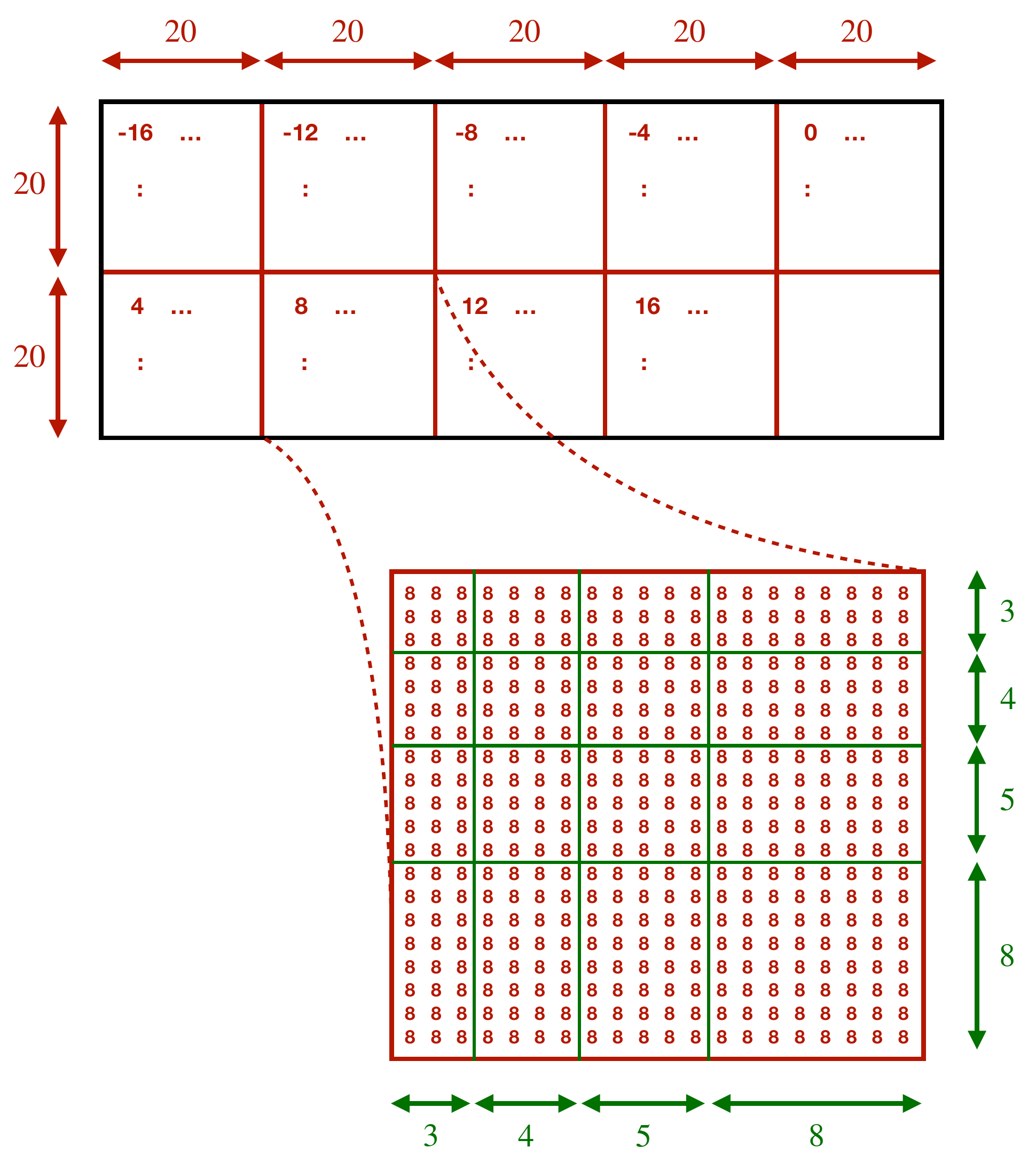}
\end{center}
\caption{The $40 \times 100$ matrix $M$. The first clustering layer ($\cG^{(r),[1]}$ and $\cG^{(c),[1]}$) is delineated in red while the second layer ($\cG^{(r),[2]}$ and $\cG^{(c),[2]}$) is delineated in green. The dimensions of the ensuing grid like structure are denoted as numbers outside the matrix. The numbers in red inside the matrix denote the mean elements $M_{ij}$ which generate the data (and are the same throughout the same ``red" first layer nested cluster). For our simulation setting, the first layer (red) is divided according to the mean value, while the second layer (green) is divided according to the covariance structure of $Z$.}
\label{nested_matrix_AE}
\end{figure}

\subsubsection{Simulation Study Comparing Algorithm \ref{alg_3} and Algorithm \ref{alg_5}}
In order to illustrate the feasibility of this nested algorithm, we include simulation results with the aforementioned  nested structure. For $n=5,\hspace{0.1cm}10,\hspace{0.1cm}15,\hspace{0.1cm}20,\hspace{0.1cm}25,\hspace{0.1cm}30,\hspace{0.1cm}35,\hspace{0.1cm}40,\hspace{0.1cm}45$, we generate $n$ i.i.d. copies of a $40 \times 100$ matrix ($p=40,~ q=100$) with two nested row clusterings $\{\cG^{(r),[1]}, \cG^{(r),[2]}\}$ and two nested column clusterings $\{\cG^{(c),[1]}, \cG^{(c),[2]}\}$ where the row clusterings are given by
\begin{align*}
    \cG^{(r),[1]} &~=~ \{G^{(r),[1]}_1~~~~~~~~\hspace{3.2cm},~G^{(r),[1]}_2~~~~~~~~\hspace{3.2cm} \}\\
    \cG^{(r),[2]} &~=~ \{G^{(r),[2]}_1,~G^{(r),[2]}_2,~G^{(r),[2]}_3, ~G^{(r),[2]}_4,~G^{(r),[2]}_5,~G^{(r),[2]}_6,~G^{(r),[2]}_7, ~G^{(r),[2]}_8 \},~~\textrm{where}~\\
    G&^{(r),[1]}_1 = \{1,2,~...~,20\}, ~~~~~ \hspace{2.73cm} G^{(r),[1]}_2 = \{21,22,~...~,40\} \\
    G&^{(r),[2]}_1 = \{1,2,3\}, ~~~\hspace{0.1cm}G^{(r),[2]}_2 = \{4,5,6,7\}, ~~G^{(r),[2]}_5 = \{21,22,23\},~G^{(r),[2]}_6 = \{24,25,26,27\}\\
    G&^{(r),[2]}_3 = \{8,...,12\}, ~G^{(r),[2]}_4 = \{13,...,20\}, ~ \hspace{0.03cm}G^{(r),[2]}_7 = \{28,...,32\},~G^{(r),[2]}_8 = \{33,...,40\}.
\end{align*} 
The column clustering can be defined in a similar fashion, but for brevity it is omitted here. The true clustering can be visualized with Figure \ref{nested_matrix_AE}.\\

\begin{figure}
\begin{center}
\includegraphics[width=0.8\linewidth]{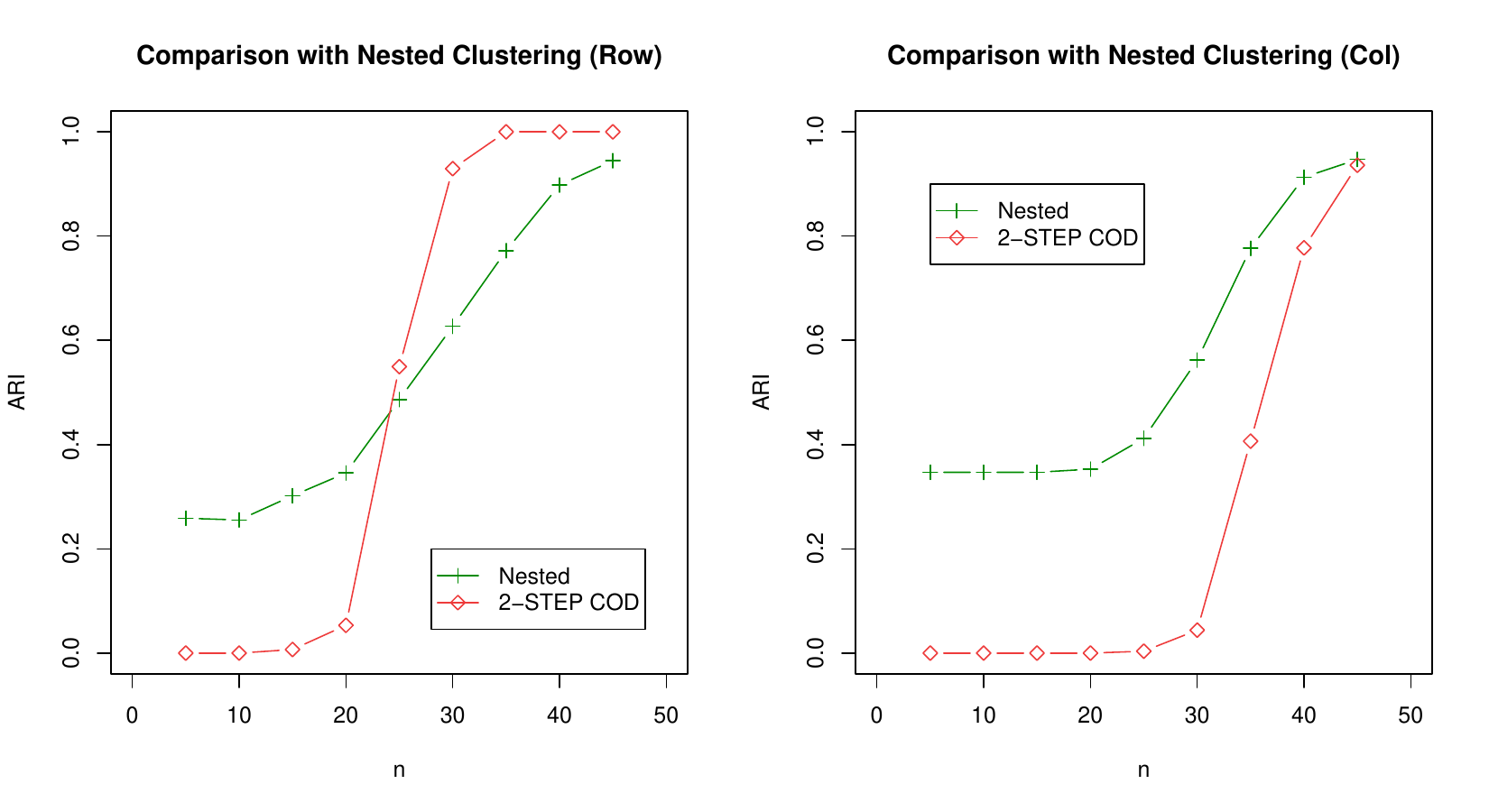}
\end{center}
\caption{The performance of the nested clustering algorithm (using both mean and covariance information) and the $\TWOSTEPCOD$ method (using covariance information only). The simulation setting is  $p=40$, $q=100$, $K_1=8,~K_2=20$, $n\in\{5,10,15,20,25,30,35,40,45\}$. The ARI values for both methods are averaged over 30 repetitions.}
\label{fig_nested_AE}
\end{figure}
We use the ARI (Adjusted Rand Index, \cite{vinh2009information}) to compare the performance of the nested clustering method with our original $\TWOSTEPCOD$ method. The results are presented in Figure \ref{fig_nested_AE}. Overall, considering the feature matrices are very large ($40 \times 100$), both methods work very well, as we get near perfect recovery once $n \geq 45$. Taking a closer look, for column clustering, the nested method is uniformly better than the original $\TWOSTEPCOD$ method over all $n$ values. For row clustering, for very small $n$ values of $5,10,15$ and $20$, the nested method performs better, but for slightly larger $n$ values of $25,30,35,40$ and $45$, the original $\TWOSTEPCOD$ method performs better. We observe that the closer the matrix is to a square, the easier it is for $\TWOSTEPCOD$ to cluster the rows and columns. If the matrix is highly unbalanced ($p>>q$ or $p<<q$), then due to the iterative nature of $\TWOSTEPCOD$ (using rows to cluster columns, then using columns to cluster the rows, etc.), clustering one of the two dimensions becomes extremely difficult. This may lead to the final result being inaccurate. In our simulation settings, for row clustering, $\TWOSTEPCOD$ clusters over the entire $40 \times 100$ matrix while the nested clustering method has to cluster over the two $20 \times 100$ submatrices separately (if the mean-based clustering can perfectly recover $\cG^{(r),[1]}$). This explains why the nested algorithm may perform worse than the original $\TWOSTEPCOD$ for row clustering. 

%Even though the submatrices are technically smaller, they are more unbalanced than the original matrix. The performance of clustering ends up depending on $n$. When $n$ is extremely small, the effect of dimension reduction is dominant, which leads to the nested method outperforming the original 2-STEP COD, but when $n$ is larger, the effect of having a more balanced matrix is dominant, which leads to the original 2-STEP COD being better. 

We also implement the competing methods \texttt{DEEM}, \texttt{TGMM} and \texttt{TEMM} (\cite{mai2022doubly}, \cite{deng2022tensor}), which are also included in the main simulation results, but none are able to be implemented successfully in this nested clustering simulation study. As explained in the main paper, their algorithms depend on different modeling assumptions and may fail to converge in our simulation setting.

All in all, we conclude that if the user wishes to incorporate mean information and also look at the finer clustering for the rows and columns that cannot be discerned by the mean alone, our $\COD$ based method can be used as the second step in a nested clustering type process (Algorithm \ref{alg_5}) to obtain reliable results.

\subsection{Higher Order Tensor Models} \label{supp_tensor}

\subsubsection{The Tensor Cluster Model and Algorithm \ref{alg_tensor}}
We present in the following the analog of our matrix clustering setting in a setting that is one order higher. Consider a three-way tensor $X\in \RR^{J\times P\times Q}$. Assume that $X$ can be decomposed as 
\begin{equation}\label{eq_tensor}
X ~=~ Z\times_1 A\times_2 B\times_3 C~+~\Gamma,
\end{equation}
where $Z\in \RR^{K_1\times K_2\times K_3}$ is a three-way latent tensor (i.e., each entry of $Z$ is a random variable) with $\EE(Z)=0$, $A\in \RR^{J\times K_1}$, $B\in \RR^{P\times K_2}$ and $C\in \RR^{Q\times K_3}$ are the unknown membership matrices for the rows, columns and tubes of $X$, respectively, and $\Gamma\in\RR^{J\times P\times Q}$ represents the mean 0 random noise tensor. Here, we use the $n$-mode product notation from the tensor literature, e.g., $Z\times_1 A$ is a $J\times K_2\times K_3$ tensor with $(Z\times_1 A)_{jst}=\sum_{k=1}^{K_1} Z_{kst}A_{jk}$. Similar to the matrix clustering setting, we assume the entries of $\Gamma$ are mutually independent and are also independent of $Z$. Each entry of the membership matrix $A$ takes values in $\{0,1\}$, such that $A_{ak}=1$ if row $a$ belongs to row cluster $k$ and $A_{ak}=0$ otherwise. The membership matrices $B$ and $C$ are defined similarly. Assuming that $n$ i.i.d. samples $X^{(1)},...,X^{(n)}$ of a random tensor $X\in \RR^{J\times P\times Q}$ are observed, our goal is to recover the membership matrices $A, B$ and $C$. 

Next, we define the so-called matricization (or equivalently, ``unfolding") of the tensor $X$. Let $X_{(1)}\in\RR^{J\times PQ}$ denote the mode-1 matricization of $X$, which arranges the mode-1 fibers of $X$ as columns into a matrix. Similarly, we can define $X_{(2)}\in\RR^{P\times JQ}$ and $X_{(3)}\in\RR^{Q\times JP}$ as the mode-2 and mode-3 matricizations of $X$. Similar to the matrix clustering setting, we consider the weighted covariance matrix $\Sigma_{1,W_1}=\EE(X_{(1)}W_1X_{(1)}^T)$, where $W_1\in\RR^{PQ\times PQ}$ is some positive semi-definite weight matrix to be chosen. It is easily seen that $\Sigma_{1,W_1}$ exhibits the clustering structure for the rows, which can be identified via our $\COD$ algorithm. In particular, let $\hat{\Sigma}_{j,W_j} = \frac{1}{n}\sum_{i=1}^n X_{(j)}^{(i)}W_jX_{(j)}^{(i)T}$ for $j=1,2,3$. The tensor clustering algorithm with the identity weight is shown in Algorithm \ref{alg_tensor}. For simplicity, we only consider the identity weight. The algorithm may further be improved by using the optimal weight as in our one-step Algorithm \ref{alg_2}  or our two-step Algorithm \ref{alg_3}, but we leave the detailed analysis for future study.\\

\begin{algorithm}[H]
\caption{Tensor Clustering Algorithm with the Identity Weight}
\begin{itemize}
\item[] \texttt{for} $j=1,2,3$
\begin{itemize} 
\item[-] Compute $\hat{\Sigma}_{j,W_j}$, where $W_1=\frac{1}{PQ}\cdot I_{PQ}$, $W_2=\frac{1}{JQ}\cdot I_{JQ}$ and $W_3=\frac{1}{PJ}\cdot I_{PJ}$. 
\item[-] Apply Algorithm \ref{alg_1} to $\hat{\Sigma}_{j,W_j}$ to construct $\hat A, \hat B$, and $\hat C$, respectively.    
\end{itemize}
\end{itemize}\label{alg_tensor}
\end{algorithm}

\subsubsection{Simulation Study}
To illustrate the feasibility of our proposed method, we also conduct a simulation study with Algorithm \ref{alg_tensor} for which the results are presented in Table \ref{table_tensor}. We have $n$ i.i.d. copies of a $15 \times 10 \times 10$ tensor ($n=10,20,30,40,50$) that have Toeplitz covariance matrices in each direction with decay rates of -0.4, 0.3 and -0.2, respectively. The noise is generated in the homogeneous setting with the noise variance being $15$ to match our main simulation settings. It is seen that clustering is more accurate as the sample size grows, which provides empirical evidence for applying Algorithm \ref{alg_tensor} to higher-order tensors. 
\begin{table}
\centering
\begin{tabular}{|c|c||c|c|c|}
\hline
\hspace{0.5cm}& \hspace{0.5cm} &\multicolumn{1}{c|}{~~~~Row (ARI)~~~~} &\multicolumn{1}{c|}{~~Column (ARI)~~} & \multicolumn{1}{c|}{~~~~Tube (ARI)~~~~}\\
\hline
\hline
\multirow{2}{*}{$n$} & 10 & 0.0863 & 0 & 0.4449 \\
     \cline{2-5}
     & 20 & 0.6637 & 0.3095 & 0.8904\\
     \cline{2-5}
     & 30 & 0.7594 & 0.7936 & 0.9334\\
     \cline{2-5}
     & 40 & 0.7864 & 0.8680 & 0.9033\\
     \cline{2-5}
     & 50 & 0.8107 & 0.8753 & 0.9889\\
\hline
\end{tabular}
\caption{The ARI values obtained from the Tensor Clustering Algorithm with the Identity Weight (Algorithm \ref{alg_tensor}) under the $n=10,20,30,40,50$ setting.}
\label{table_tensor}
\end{table}

While our preliminary simulation results are indeed promising, proving minimax optimality would be challenging in the tensor clustering. Also, as mentioned above, since using more intricate weight matrices instead of the identity matrix may improve the clustering performance, a rigorous analysis of the algorithm would also be interesting. All of these directions are promising fields of future study.

\section{Proofs}\label{app_proof}

\subsection{Clustering consistency}
First, we state the Hanson-Wright inequality, which is instrumental in the proof.
\begin{lemma}
\label{Hanson Wright}
\textbf{(Hanson-Wright)} There exist positive constants $c,c'$ such  that for all $n \times n$ matrices $H$, if $\xi = (\xi_1,...,\xi_n)^T$ is a vector of independent mean 0 sub-Gaussian random variables with $||\xi_i||_{\psi_2} \leq L$ for some $L > 0$, then for all t, the following holds:
\begin{align*}
    \PP\bigg[ \big|\xi^TH\xi - \EE(\xi^TH\xi) \big|>cL^2(||H||_F\sqrt{t} + c'||H||_{\text{op}}t) \bigg] \leq e^{-t}.
\end{align*}
\end{lemma}

\subsubsection{Proof of Theorem \ref{thm_upper}} \label{proof_upper}

\begin{proof} 
Denote $\hat L\hat L^T=\hat W$ and $\bar X^{(i)}=X^{(i)}\hat L$. So, $\hat\Sigma_{p,\hat W}=\frac{1}{|D_2|}\sum_{i\in D_2}\bar X^{(i)}\bar X^{(i)T}$. For simplicity, we assume $|D_2|=n/2$, and the summation is from $1$ to $n/2$. Furthermore, denote $H_{k,m}^{+}=\Var(\bar X_{k\cdot} + \bar X_{m\cdot}|\hat W)$ and $H_{k,m}^{-}=\Var(\bar X_{k\cdot} - \bar X_{m\cdot}|\hat W)$. Denote $\tilde{X}_{k,m}^{+}=(H_{k,m}^{+})^{-1/2}(\bar X_{k\cdot}^T + \bar X_{m\cdot}^T)$. Since $\tilde{X}_{k,m}^{+}$ has mean 0 variance 1 given $\hat W$ and $X$ is Gaussian, $\tilde{X}_{k,m}^{+}$ is a vector of independent Gaussian random variables given $\hat W$. 

Denote
\begin{align*}
    \tilde{\XX}_{k,m}^{+}=\begin{bmatrix}
    (H_{k,m}^{+})^{-1/2}&&\\
    &...&\\
    &&(H_{k,m}^{+})^{-1/2}\\
    \end{bmatrix}\begin{bmatrix}
    (\bar X_{k\cdot}^{(1)}+\bar X_{m\cdot}^{(1)})^T\\
    :\\
    (\bar X_{k\cdot}^{(n/2)}+\bar X_{m\cdot}^{(n/2)})^T\\
    \end{bmatrix}=\begin{bmatrix}
    \tilde{X}_{k,m}^{+(1)}\\
    :\\
    \tilde{X}_{k,m}^{+(n/2)}\\
    \end{bmatrix}\\
     \tilde{\XX}_{k,m}^{-}=\begin{bmatrix}
    (H_{k,m}^{-})^{-1/2}&&\\
    &...&\\
    &&(H_{k,m}^{-})^{-1/2}\\
    \end{bmatrix}\begin{bmatrix}
    (\bar X_{k\cdot}^{(1)}-\bar X_{m\cdot}^{(1)})^T\\
    :\\
    (\bar X_{k\cdot}^{(n/2)}-\bar X_{m\cdot}^{(n/2)})^T\\
    \end{bmatrix}=\begin{bmatrix}
    \tilde{X}_{k,m}^{-(1)}\\
    :\\
    \tilde{X}_{k,m}^{-(n/2)}\\
    \end{bmatrix}\\
    \text{and} \hspace{5cm}\\
    \HH_{k,m}^{+}=\begin{bmatrix}
    (H_{k,m}^{+})^{-1/2}&&\\
    &...&\\
    &&(H_{k,m}^{+})^{-1/2}\\
    \end{bmatrix}, \hspace{0.1cm} \HH_{k,m}^{-}=\begin{bmatrix}
    (H_{k,m}^{-})^{-1/2}&&\\
    &...&\\
    &&(H_{k,m}^{-})^{-1/2}\\
    \end{bmatrix},
\end{align*}
where $\HH_{k,m}^{+}$ and $\HH_{k,m}^{-}$ each have $n/2$ blocks on the diagonal. Note that the quadratic forms can be used to get the sum of the products. The following calculation holds for one observation:
\begin{align*}
(\tilde{X}_{k,m}^{+})^{T}H_{k,m}^{+}\tilde{X}_{k,m}^{+} - (\tilde{X}_{k,m}^{-})^{T}H_{k,m}^{-}\tilde{X}_{k,m} &= \sum_{j=1}^q (\bar X_{kj}^2 + 2\bar X_{kj}\bar X_{mj} + \bar X_{mj}^2) \\ &\hspace{3cm} - \sum_{j=1}^q (\bar X_{kj}^2 - 2\bar X_{kj}\bar X_{mj} + \bar X_{mj}^2)\\ &= 4\sum_{j=1}^q \bar X_{kj}\bar X_{mj}.
\end{align*}
We can extend this to $n/2$ observations by using block matrix notation from above.
\begin{align*}
&(\tilde{\XX}_{k,m}^{+})^T\HH_{k,m}^{+}\tilde{\XX}_{k,m}^{+} - (\tilde{\XX}_{k,m}^{-})^T\HH_{k,m}^{-}\tilde{\XX}_{k,m}^{-} \\
&= \sum_{i=1}^{n/2}\sum_{j=1}^q (\bar X_{kj}^{(i)2} + 2\bar X_{kj}^{(i)}\bar X_{mj}^{(i)} + \bar X_{mj}^{(i)2})- \sum_{i=1}^{n/2}\sum_{j=1}^q (\bar X_{kj}^{(i)2} - 2\bar X_{kj}^{(i)}\bar X_{mj}^{(i)} + \bar X_{mj}^{(i)2})\\
&=4\sum_{i=1}^{n/2}\sum_{j=1}^q \bar X_{kj}^{(i)}\bar X_{mj}^{(i)}.
\end{align*}
Thus, we have
\begin{align*}
||\hat{\Sigma}_{p,\hat W} - \Sigma_{p,\hat W}||_{\max} &= \underset{k,m}{\max}\bigg|\frac{2}{n}\sum_{i=1}^{n/2}\sum_{j=1}^{q}\bar X_{kj}^{(i)}\bar X_{mj}^{(i)} - \sum_{j=1}^{q}\EE(\bar X_{kj} \bar X_{mj}|\hat W)\bigg|\\
&=\underset{k,m}{\text{max}}\bigg|\frac{1}{2n}\bigg(  (\tilde{\XX}_{k,m}^{+})^T\HH_{k,m}^{+}\tilde{\XX}_{k,m}^{+}-\EE((\tilde{\XX}_{k,m}^{+})^T\HH_{k,m}^{+}\tilde{\XX}_{k,m}^{+}|\hat W) \bigg) \\
&~~~ - \frac{1}{2n}\bigg( (\tilde{\XX}_{k,m}^{-})^T\HH_{k,m}^{-}\tilde{\XX}_{k,m}^{-}-\EE((\tilde{\XX}_{k,m}^{-})^T\HH_{k,m}^{-}\tilde{\XX}_{k,m}^{-}|\hat W)  \bigg)\bigg|.
\end{align*}
By Lemma \ref{Hanson Wright}, we know that there exist positive constants $c,c',c'',c'''$ such that: conditioned on $\hat W$ with probability greater than $1-e^{-t}$,
\begin{align*}
    \big|(\tilde{\XX}_{k,m}^{+})^T\HH_{k,m}^{+}\tilde{\XX}_{k,m}^{+}-\EE((\tilde{\XX}_{k,m}^{+})^T\HH_{k,m}^{+}\tilde{\XX}_{k,m}^{+}|\hat W)\big| &\leq c\bigg(||\HH_{k,m}^{+}||_F\sqrt{t} + c'||\HH_{k,m}^{+}||_{\text{op}}\cdot t\bigg).
\end{align*}
Similarly, conditioned on $\hat W$ with probability greater than $1-e^{-t}$,
$$
    \big|(\tilde{\XX}_{k,m}^{-})^T\HH_{k,m}^{-}\tilde{\XX}_{k,m}^{-}-\EE((\tilde{\XX}_{k,m}^{-})^T\HH_{k,m}^{-}\tilde{\XX}_{k,m}^{-}|\hat W)\big| \leq c''\bigg(||\HH_{k,m}^{-}||_F\sqrt{t} + c'''||\HH_{k,m}^{-}||_{\text{op}}\cdot t\bigg). 
$$    
By the definition of $\HH_{k,m}^{+}$, we know $||\HH_{k,m}^{+}||_F^2=n||H_{k,m}^{+}||_F^2/2$ and $||\HH_{k,m}^{+}||_{\text{op}}=||H_{k,m}^{+}||_{\text{op}}$. Thus, with a union bound over $k,m$, we get the following concentration inequality:
\begin{align}
||\hat{\Sigma}_{p,\hat W} - \Sigma_{p,\hat W}||_{\max}&\lesssim \frac{1}{n}\cdot\max_{k,m}\bigg( \sqrt{||H_{k,m}^{+}||_F^2 \cdot n t} + ||H_{k,m}^{+}||_{\text{op}}\cdot t \bigg) \nonumber \\
   & \hspace{2cm} +\frac{1}{n}\cdot\max_{k,m}\bigg(\sqrt{||H_{k,m}^{-}||_F^2 \cdot n t} + ||H_{k,m}^{-}||_{\text{op}} \cdot t \bigg) \label{hw}
\end{align}
conditioned on $\hat W$ with probability greater than $1-\frac{p(p-1)}{2}e^{-t}$, and setting $t=3\log p$, we know the inequality (\ref{hw}) holds with conditional probability greater than $1-\frac{1}{2p}$. By taking another expectation with $\hat W$, the inequality (\ref{hw}) holds with probability greater than $1-\frac{1}{2p}$ unconditionally. 

Because $||\cdot||_F \geq ||\cdot||_{op}$ holds in general, the Frobenius term dominates in the regime of $\frac{\log p}{n} = o(1)$, so all that is left to do is to bound $||H_{k,m}^{+}||_{F}^2$. An identical argument can be made for $||H_{k,m}^{-}||_{F}^2$, and we know the following upper bound holds for both terms,
\begin{align}\label{ff}
    ||H_{k,m}^{+}||_{F} &\leq ||\Var(\bar X_{k\cdot}|\hat W)||_F + ||\Var(\bar X_{m\cdot}|\hat W)||_F + 2\cdot||\cov(\bar X_{k\cdot},\bar X_{m\cdot}|\hat W)||_F \nonumber \\
    &\leq C_1\cdot \underset{k}{\text{max}}||\Var(\bar X_{k\cdot}|\hat W)||_F. \nonumber
    \end{align}
This holds for all pairs $(k,m)$, and so it also holds for the maximum over the pairs:
\begin{align}
    \max_{k,m}||H_{k,m}^{+}||_{F} &\leq C_1\cdot \underset{k}{\max}||\Var(\bar X_{k\cdot}|\hat W)||_F= C_1 \max_{k}\|\hat L^T\Var(X_{k\cdot})\hat L\|_F= C_1 K_2^{-1/2} \|X\|_{\hat W}. 
\end{align}
Thus, with probability greater than $1-\frac{1}{2p}$,
\begin{equation}\label{eq_sigmabound}
 ||\hat{\Sigma}_{p,\hat W} - \Sigma_{p,\hat W}||_{\max}\leq c_1\sqrt{\frac{\log p}{nK_2}} \|X\|_{\hat W}\leq \eta \|X\|_{\hat W}
\end{equation}
for some constant $c_1$, where the last step is from condition (A1). 

Note that conditions (A1) and (A2) together imply that 
\begin{equation}\label{eq_mcodbound}
\MCOD(\Sigma_{p,\hat W})/\|X\|_{\hat W}\geq 4\eta. 
\end{equation}
In the following, we show how perfect clustering is achieved under the event that the inequality (\ref{eq_sigmabound}) holds. For simplicity, we drop the subscript $\hat W$ in $\hat{\Sigma}_{p,\hat W}, \Sigma_{p,\hat W}$ and $\|X\|_{\hat W}$ and just write  $\hat{\Sigma}_{p}, \Sigma_{p}$ and $\|X\|$. Also, for simplicity, in the following derivation, we will denote the $(i,j)$-th element of $\Sigma_p$ as $\Sigma_{p,ij}$. Define the following quantity
\[\tau := \underset{\underset{c \neq a,b}{a,b,c}}{\max}\big|\hat{\Sigma}_{p,ac}-\hat{\Sigma}_{p,bc}-(\Sigma_{p,ac}-\Sigma_{p,bc})\big|.\]
Then we have:
\begin{align*}
    \tau &= \underset{\underset{c \neq a,b}{a,b,c}}{\max}\big|\hat{\Sigma}_{p,ac}-\Sigma_{p,ac}-(\hat{\Sigma}_{p,bc}-\Sigma_{p,bc})\big| \\
    &\leq \underset{\underset{c \neq a,b}{a,b,c}}{\max} \big|\hat{\Sigma}_{p,ac}-\Sigma_{p,ac}\big| + \underset{\underset{c \neq a,b}{a,b,c}}{\max} \big|\hat{\Sigma}_{p,bc}-\Sigma_{p,bc}\big|\\
    &\leq 2 \cdot ||\hat{\Sigma}_p-\Sigma_p||_{\max}\\
    &\leq 2 \eta \cdot ||X||,
\end{align*}
where the last step is from (\ref{eq_sigmabound}). We now want to show the following inequality:
\begin{equation}\label{eq_mcodbound2}
\COD_{\widehat{\Sigma}_p}(a,b) - \tau \leq \COD_{\Sigma_p}(a,b) \leq \COD_{\widehat{\Sigma}_p}(a,b) + \tau.
\end{equation} 
The inequality on the left holds because of the following: 
\begin{align}
    \COD_{\widehat{\Sigma}_p}(a,b) &= \underset{c\neq a,b}{\max} \big|\hat{\Sigma}_{p,ac} - \hat{\Sigma}_{p,bc}\big|\nonumber\\ 
    &= \underset{c\neq a,b}{\max} \big|\hat{\Sigma}_{p,ac} - \hat{\Sigma}_{p,bc} - (\Sigma_{p,ac}-\Sigma_{p,bc}) + (\Sigma_{p,ac}-\Sigma_{p,bc})\big|\nonumber\\
    &\leq \underset{c\neq a,b}{\max} \big| \Sigma_{p,ac}-\Sigma_{p,bc} \big| +  \underset{c\neq a,b}{\max} \big|\hat{\Sigma}_{p,ac} - \hat{\Sigma}_{p,bc} - (\Sigma_{p,ac}-\Sigma_{p,bc}) \big|\nonumber\\
    &\leq \COD_{\Sigma_p}(a,b) + \tau.\label{eq_mcodbound3}
\end{align}
With a similar technique, we can get the inequality on the right as well. Now, for any two sets $A$ and $B$, if $A$ and $B$ indeed belong to the same cluster (i.e., $a,b\in G_k$ for some $k$ and for any $a\in A$ and $b\in B$), we have $\COD_{{\Sigma}_p}(a,b)=0$ and therefore by (\ref{eq_mcodbound2})
$$
\COD_{\widehat{\Sigma}_p}(A,B)=\max_{a\in A, b\in B} \COD_{\widehat{\Sigma}_p}(a,b)\leq \tau\leq 2 \eta \cdot ||X||.
$$
If $A$ and $B$ are not the same clusters, there must exist $a\in A$ and $b\in B$ with $a\in G_k$ and $b\in G_j$ for some $k\neq j$. Then
$$
\COD_{\widehat{\Sigma}_p}(A,B)\geq \COD_{\widehat{\Sigma}_p}(a,b)\geq  \COD_{\Sigma_p}(a,b) - \tau \geq \MCOD(\Sigma_p) - \tau \geq \MCOD(\Sigma_p) - 2 \eta \cdot ||X||,
$$
where the first inequality is from the definition of $\COD_{\widehat{\Sigma}_p}(A,B)$, the second one is from (\ref{eq_mcodbound2}), the third one is from the definition of $\MCOD$, and finally the last one is from (\ref{eq_mcodbound3}). 

Together with (\ref{eq_mcodbound}), finally we show that $\COD_{\widehat{\Sigma}_p}(A,B)\geq 2 \eta \cdot ||X||$, if $A$ and $B$ are not the same clusters. By taking the threshold value $\alpha = 2 \eta \cdot ||X||$, we attain perfect 
clustering using our hierarchical algorithm. This completes the proof. 
\end{proof}

\subsection{Minimax Lower Bound}
The following lemma is Birge's Lemma applied to our specific setting, similar to Lemma C.1 in \cite{clustering}. Define $M_O(p,q,K_1,K_2,\eta)$ as in Section \ref{sec_minimax}. For ease of notation, we will simply use $M$.
\begin{lemma}
\label{Birge}
 For any partition estimator \(\widehat{\cG}\), and for any collection of distinct covariance matrices \(\Sigma^{(j)} \in M(p,q,K_1,K_2,\eta)\),
\begin{align*}
    \underset{\Sigma \in M(p,q,K_1,K_2,\eta)}{\sup}\PP_{\Sigma}(\widehat{\cG} \neq \cG^{*}) &\geq \underset{j=1,...,N}{\max}\PP_{\Sigma^{(j)}}(\widehat{\cG} \neq \cG^{(j)})\\ &\geq \frac{1}{2e+1} \wedge  (1-\underset{j\geq2}{\max}\frac{\mathrm{KL}(\Sigma^{(j)},\Sigma^{(1)})}{\log(N)}).
\end{align*}
\end{lemma}

\subsubsection{Proof of Theorem \ref{thm_lowerbound_2}}
Similar to the construction in \cite{clustering}, set our collection of covariance matrices as follows: $K_1=3, m_p^{(1)}=m_p^{(2)}=m_p^{(3)}=\frac{p}{3}$ (equal row cluster size),  $X=AZB^T+\Gamma$, where $\Gamma_{p\times q}$ has entries $\Gamma_{ij} \underset{iid}{\sim} N(0,\sigma_{(t)}^2)$ where $\sigma_{(t)}^2$ denotes the noise variance of elements in column cluster $t$. Note that for this construction, we assume the same noise variance within the same column cluster. In addition, we set 
\begin{align*}
Z&=\begin{bmatrix} 
        Z_{11}&...&Z_{1K_2}\\
        Z_{21}&...&Z_{2K_2}\\
        Z_{31}&...&Z_{3K_2}\\
\end{bmatrix} = \begin{bmatrix} 
        Z_{\cdot 1}&...&Z_{\cdot K_2}\\
\end{bmatrix}, \hspace{0.1cm} Z_{\cdot 1},...,Z_{\cdot K_2} \underset{iid}{\sim} N(0,C(\epsilon)),\\
&\text{ where } C(\epsilon) = \begin{bmatrix}
        \epsilon&\epsilon-\epsilon^2&-\epsilon\\
        \epsilon-\epsilon^2&\epsilon&\epsilon\\
        -\epsilon&\epsilon&2\\
\end{bmatrix},~\textrm{and}~
A = \begin{bmatrix}
        1&&\\
        :&&\\
        1&&\\
        &1&\\
        &:&\\
        &1&\\
        &&1\\
        &&:\\
        &&1\\
\end{bmatrix}_{p\times 3},\hspace{0.2cm} B = \begin{bmatrix}
        1&&&&\\
        :&&&\\
        1&&&\\
        &1&&\\
        &:&&\\
        &1&&\\
        &&:&\\
        &&&1\\
        &&&:\\
        &&&1\\
\end{bmatrix}_{q \times K_2},
\end{align*}
where $0<\epsilon<1$ is a quantity to be specified later.  We will consider \(N=(\frac{p}{3})^2+1\) many covariance matrices that are obtained by switching one of the rows in the first third of the rows in $A$ and one of the rows in the second third of the rows in $A$ (and also counting the original matrix $A$ as well).

It follows that \(\Var(\text{vec}(Z))=I_{K_2} \otimes C(\epsilon)\) and that
\begin{align*}
\Sigma ~ (\in \RR^{pq \times pq}) &=  \Var(\text{vec}(X))\\ &= (B\otimes A)(\Var(\text{vec}(Z)))(B\otimes A)^T + \Var(\text{vec}(\Gamma))\\
&=(B\otimes A)(I_{K_2} \otimes C(\epsilon))(B\otimes A)^T + \Var(\text{vec}(\Gamma))\\
&=\big\{BB^T \otimes AC(\epsilon)A^T\big\} + \Var(\text{vec}(\Gamma))\\
&=\begin{bmatrix}
        1_{m_q^{(1)}}1_{m_q^{(1)}}^T\otimes AC(\epsilon)A^T&&\\
        &...&\\
        &&1_{m_q^{(K_2)}}1_{m_q^{(K_2)}}^T\otimes AC(\epsilon)A^T\\
\end{bmatrix}  \\
 &\hspace{8cm}+ \begin{bmatrix}
        \sigma_{(1)}^2 I_{pm_{q}^{(1)}}&&\\
        &...&\\
        &&\sigma_{(K_2)}^2 I_{pm_{q}^{(K_2)}}\\
\end{bmatrix}.
\end{align*}
So we can set \(\tilde{\Sigma}^{(t)}=\big\{1_{m_q^{(t)}}1_{m_q^{(t)}}^T\otimes AC(\epsilon)A^T\big\} + \sigma^2_{{(t)}}I_{pm_q^{(t)}}\) as one of the (\(pm_q^{(t)} \times pm_q^{(t)}\)) blocks in the block diagonal matrix \(\Sigma \in \RR^{pq \times pq}\).

Note that the Kullback-Leibler Divergence for \(n\) iid multivariate normal observations in \(\RR^{d}\) with mean \(0\) is \(KL(\Sigma^{'},\Sigma)=\frac{n}{2}\big[ \text{tr}(\Sigma^{-1}\Sigma^{'}-I_{d})-\text{log}\text{ det}(\Sigma^{-1}\Sigma^{'})  \big]\), where $\Sigma$ and $\Sigma'$ are two covariance matrices constructed as above. Thus, since \(\Sigma\) is a block diagonal matrix with blocks \(\tilde{\Sigma}^{(t)}\), we have the relationship \(KL(\Sigma^{'},\Sigma) =  \sum_{t=1}^{K_2} KL(\tilde{\Sigma}^{(t)'},\tilde{\Sigma}^{(t)})\), and it suffices to just calculate \(KL(\tilde{\Sigma}^{(t)'},\tilde{\Sigma}^{(t)})\) in order to calculate \(KL(\Sigma^{'},\Sigma)\). 

Denote the eigenvalues and eigenvectors of \(C(\epsilon)\) as $\lambda_1,\lambda_2,\lambda_3$ and $u_1,u_2,u_3$, respectively. Also, define the $pm_q^{(t)} \times 1$ vector $v_k^{(t)}$ as $v_k^{(t)}=\frac{1}{\sqrt{m_pm_q^{(t)}}}(1_{m_q^{(t)}}\otimes Au_k)$. 
\begin{align*}
    \tilde{\Sigma}^{(t)} &= \big\{1_{m_q^{(t)}}1_{m_q^{(t)}}^T\otimes AC(\epsilon)A^T\big\} + \sigma^2_{(t)}I_{pm_q^{(t)}}
    = \sigma^2_{(t)} \big[ \sum_{k=1}^2 \frac{\lambda_k m_pm_q^{(t)}}{\sigma^2_{(t)}}\cdot v_k^{(t)}v_k^{(t)T} + I_{pm_q^{(t)}} \big],
\end{align*}
where we note that $\lambda_3=0$. Therefore, 
\begin{align*}
    (\tilde{\Sigma}^{(t)})^{-1} &= \frac{1}{\sigma^2_{(t)}} \big[ (-\sum_{k=1}^2 \frac{\lambda_k m_pm_q^{(t)}}{\lambda_k m_pm_q^{(t)} + \sigma^2_{(t)}} \cdot v_k^{(t)}v_k^{(t)T} ) + I_{pm_q^{(t)}} \big]. 
\end{align*}
Let $Q_{p \times p}$ denote a $p\times p$ perturbation matrix switching one row in the first third of the rows in $A$ and one row in the second third of the rows in $A$. 
Define $\tilde{Q}^{(t)}=I_{m_q^{(t)}} \otimes Q_{p\times p}$. Then $\tilde{Q}^{(t)}(\tilde{Q}^{(t)})^T=I_{pm_q^{(t)}}$ holds. Now, set $\tilde{Q}^{(t)}v_k^{(t)} = v_k^{(t)} + \Delta_k^{(t)}$. It follows that 
\begin{align*}
    \Delta_k^{(t)} = \frac{1}{\sqrt{m_pm_q^{(t)}}}\cdot(1_{m_q^{(t)}}\otimes \begin{bmatrix}
    (u_k)_2 - (u_k)_1\\
    0\\
    :\\
    0\\
    (u_k)_1 - (u_k)_2\\
    0\\
    :\\
    0\\
    0\\
    :\\
    0\\
    \end{bmatrix}_{p \times 1}).
\end{align*}
Then we can calculate $\tilde{\Sigma}^{(t)'}-\tilde{\Sigma}^{(t)}$:
\begin{align*}
    \tilde{\Sigma}^{(t)'}-\tilde{\Sigma}^{(t)}&=\tilde{Q}^{(t)}\tilde{\Sigma}^{(t)}\tilde{Q}^{(t)T} - \tilde{\Sigma}^{(t)}\\
    &=\tilde{Q}^{(t)}\bigg(  \sigma^2_{(t)} \big[ \sum_{k=1}^2 \frac{\lambda_k m_pm_q^{(t)}}{\sigma^2_{(t)}}\cdot v_k^{(t)} v_k^{(t)T} + I_{pm_q^{(t)}} \big] \bigg) \tilde{Q}^{(t)T} \\ &\hspace{7cm} - \sigma^2_{(t)} \big[ \sum_{k=1}^2 \frac{\lambda_k m_pm_q^{(t)}}{\sigma^2_{(t)}}\cdot v_k^{(t)}v_k^{(t)T} + I_{pm_q^{(t)}} \big]\\
    &=\sigma^2_{(t)} \bigg[ \big\{\sum_{k=1}^2\frac{\lambda_k m_pm_q^{(t)}}{\sigma^2_{(t)}}(\tilde{Q}^{(t)}v_k^{(t)}v_k^{(t)T}\tilde{Q}^{(t)T} - v_k^{(t)}v_k^{(t)T})\big\} + \tilde{Q}^{(t)}\tilde{Q}^{(t)T} - I_{pm_q^{(t)}}\bigg]\\
    &=\sum_{k=1}^2\lambda_k m_pm_q^{(t)}(\tilde{Q}^{(t)}v_k^{(t)}v_k^{(t)T}\tilde{Q}^{(t)T} - v_k^{(t)}v_k^{(t)T})\\
    &= \lambda_2m_pm_q^{(t)}(v_2^{(t)}\Delta_2^{(t)T} + \Delta_2^{(t)}v_2^{(t)T} + \Delta_2^{(t)}\Delta_2^{(t)T}),
\end{align*}
where the last equality holds because $\Delta_1=0$. Now we compute 
$(\tilde{\Sigma}^{(t)})^{-1}\tilde{\Sigma}^{(t)'}$ in the following way:
\begin{align*}
   &(\tilde{\Sigma}^{(t)})^{-1}\tilde{\Sigma}^{(t)'} \\
   &= I +  (\tilde{\Sigma}^{(t)})^{-1}(\tilde{\Sigma}^{(t)'}-\tilde{\Sigma}^{(t)})\\ &= I + \frac{1}{\sigma^2_{(t)}} \bigg[ (-\sum_{k=1}^2 \frac{\lambda_k m_pm_q^{(t)}}{\lambda_k m_pm_q^{(t)} + \sigma^2_{(t)}} \cdot v_k^{(t)}v_k^{(t)T} ) + I_{pm_q^{(t)}} \bigg] \bigg[ \lambda_2m_pm_q^{(t)}(v_2^{(t)}\Delta_2^{(t)T} + \Delta_2^{(t)}v_2^{(t)T} + \Delta_2^{(t)}\Delta_2^{(t)T}) \bigg]\\
   &=I + \frac{\lambda_2m_pm_q^{(t)}}{\sigma^2_{(t)}}\bigg[ \big( I - \frac{\lambda_2m_pm_q^{(t)}}{\lambda_2m_pm_q^{(t)}+ \sigma^2_{(t)}}v_2^{(t)}v_2^{(t)T} \big) \big( v_2^{(t)}\Delta_2^{(t)T} + \Delta_2^{(t)}v_2^{(t)T} + \Delta_2^{(t)}\Delta_2^{(t)T}\big)  \bigg]\\
   &= I + \frac{\lambda_2m_pm_q^{(t)}}{\sigma^2_{(t)}}F^{(t)},
\end{align*}
where the second to third line holds because $v_1^{(t)}v_1^{(t)T}\Delta_2^{(t)} = v_1^{(t)}v_1^{(t)T}v_2^{(t)} = 0$ and the last line holds from
\begin{align*}
    F^{(t)}&=\big( I - \frac{\lambda_2m_pm_q^{(t)}}{\lambda_2m_pm_q^{(t)}+ \sigma^2_{(t)}}v_2^{(t)}v_2^{(t)T} \big) \big( v_2^{(t)}\Delta_2^{(t)T} + \Delta_2^{(t)}v_2^{(t)T} + \Delta_2^{(t)}\Delta_2^{(t)T}\big)\\
    &=\big( I - \tilde{\rho}^{(t)} v_2^{(t)}v_2^{(t)T} \big) \big( v_2^{(t)}\Delta_2^{(t)T} + \Delta_2^{(t)}v_2^{(t)T} + \Delta_2^{(t)}\Delta_2^{(t)T}\big)\\
    &=v_2^{(t)}\Delta_2^{(t)T}(1-\tilde{\rho}^{(t)}(1+s^{(t)})) + \Delta_2^{(t)}v_2^{(t)T} + \Delta_2^{(t)}\Delta_2^{(t)T} - \tilde{\rho}^{(t)}s^{(t)}v_2^{(t)}v_2^{(t)T},
\end{align*}
where $\tilde{\rho}^{(t)}=\frac{\lambda_2m_pm_q^{(t)}}{\lambda_2m_pm_q^{(t)}+ \sigma^2_{(t)}}$ and  $s^{(t)}=\Delta_2^{(t)T}v_2^{(t)}$. Note that the two non-zero eigenvalues of $F^{(t)}$, which are $\mu_1^{(t)}$, and $\mu_2^{(t)}$, satisfy $\mu_1^{(t)} + \mu_2^{(t)} = -\tilde{\rho}^{(t)}s^{(t)}(2+s^{(t)})$ and $\mu_1^{(t)} \mu_2^{(t)} = (1-\tilde{\rho}^{(t)})s^{(t)}(2+s^{(t)})$. Using these facts, we can finally calculate the KL divergence
\begin{align*}
    KL(\tilde{\Sigma}^{(t)'},\tilde{\Sigma}^{(t)}) &= \frac{n}{2}\bigg[ \text{tr}\bigg((\tilde{\Sigma}^{(t)})^{-1}\tilde{\Sigma}^{(t)'}-I_{pm_q^{(t)}}\bigg)-\text{log}\text{ det}\bigg((\tilde{\Sigma}^{(t)})^{-1}\tilde{\Sigma}^{(t)'}\bigg)  \bigg]\\
    &= \frac{n}{2}\big[ \text{tr}(   \frac{\lambda_2m_pm_q^{(t)}}{\sigma^2_{(t)}}F^{(t)})-\text{log}\text{ det}(I_{pm_q^{(t)}}+\frac{\lambda_2m_pm_q^{(t)}}{\sigma^2_{(t)}}F^{(t)})  \big]\\
    &=\frac{n}{2}\bigg[ \frac{\lambda_2m_pm_q^{(t)}}{\sigma^2_{(t)}}(-\tilde{\rho}^{(t)}s^{(t)}(2+s^{(t)})) \\
    &\hspace{0.4cm}- \text{log}\bigg\{ 1+ \frac{\lambda_2m_pm_q^{(t)}}{\sigma^2_{(t)}}(-\tilde{\rho}^{(t)}s^{(t)}(2+s^{(t)})) + \big(\frac{\lambda_2m_pm_q^{(t)}}{\sigma^2_{(t)}}\big)^2(1-\tilde{\rho}^{(t)})s^{(t)}(2+s^{(t)}) \bigg\}  \bigg]\\
    &=\frac{n}{2}\bigg[ \frac{\lambda_2m_pm_q^{(t)}}{\sigma^2_{(t)}}(-\tilde{\rho}^{(t)}s^{(t)}(2+s^{(t)}))\bigg],
\end{align*}
where the last equality holds because $(1-\tilde{\rho}^{(t)})\cdot\frac{\lambda_2m_pm_q^{(t)}}{\sigma^2_{(t)}} = \tilde{\rho}^{(t)}$. {Now, plugging in $s^{(t)}=-\frac{2\epsilon^2}{m_p(2+\epsilon^2)}$}, we get:
\begin{align*}
    KL(\tilde{\Sigma}^{(t)'},\tilde{\Sigma}^{(t)}) &= 2n\tilde{\rho}^{(t)}\cdot\frac{m_q^{(t)}\lambda_2}{\sigma^2_{(t)}}\frac{\epsilon^2}{2+\epsilon^2}(1-\frac{\epsilon^2}{m_p(2+\epsilon^2)})\leq 2n\epsilon^2\frac{m_q^{(t)}}{\sigma^2_{(t)}},
    \end{align*}
where we use the fact that $\lambda_2=2+\epsilon^2$. Thus,
\begin{equation}\label{eq_KLD}
KL(\Sigma^{'},\Sigma) \leq 2n\epsilon^2\sum_{t=1}^{K_2}\frac{m_q^{(t)}}{\sigma^2_{(t)}}.
\end{equation}
In the following, we will show an lower bound for the rate of $\frac{\MCOD(\Sigma_{p,W_O})}{||X||_{W_O}}$.
We will first compute $\MCOD(\Sigma_{p,W_O})$,  
%\begin{align*}
%    \Sigma_{p,W_I} = \frac{1}{q}\EE(XX^T)&=\frac{1}{q}\EE\big[ (AZB^T+\Gamma)(BZ^TA^T + \Gamma^T)\big]\\
%    &=\frac{1}{q}\big[ A\EE(ZB^TBZ^T)A^T+\EE(\Gamma\Gamma^T)\big]\\
%    &=\frac{1}{q}A\EE 
%    \bigg[\sum_{t=1}^{K_2}|[t]|\begin{bmatrix} Z_{1t}\\Z_{2t}\\Z_{3t}
%    \end{bmatrix}Z_{1t}, \hspace{0.3cm} \sum_{t=1}^{K_2}|[t]| \begin{bmatrix} Z_{1t}\\Z_{2t}\\Z_{3t}
%    \end{bmatrix} Z_{2t}, \hspace{0.3cm} \sum_{t=1}^{K_2}|[t]| \begin{bmatrix} Z_{1t}\\Z_{2t}\\Z_{3t}
%    \end{bmatrix} Z_{3t}\bigg]A^T \\ &\hspace{1cm} + \frac{1}{q}\EE(\Gamma\Gamma^T)\\
%    &=\frac{1}{q}A 
%    \bigg[\sum_{t=1}^{K_2}|[t]|C(\epsilon)_{\cdot1}, \hspace{0.3cm} \sum_{t=1}^{K_2}|[t]| C(\epsilon)_{\cdot2}, \hspace{0.3cm} \sum_{t=1}^{K_2}|[t]| C(\epsilon)_{\cdot3}\bigg]A^T + \frac{1}{q}\EE(\Gamma\Gamma^T)\\
%    &=\frac{1}{q}\bigg(\sum_{t=1}^{K_2}|[t]|\bigg)AC(\epsilon)A^T + \frac{1}{q}\EE(\Gamma\Gamma^T)\\
%    &=AC(\epsilon)A^T + \frac{1}{q}\EE(\Gamma\Gamma^T)
%\end{align*}
\begin{align*}    
     \COD_{\Sigma_{p,W_O}}(u,v) &= \underset{w\neq u,v}{{\text{max}}}\frac{1}{K_2}\bigg| \Big[A\EE(ZZ^T)A^T + \EE(\Gamma B(B^TB)^{-2}B^T\Gamma^T)\Big]_{uw}\\
     &\hspace{3cm} - \Big[A\EE(ZZ^T)A^T + \EE(\Gamma B(B^TB)^{-2}B^T\Gamma^T)\Big]_{vw}\bigg|\\
     &= \underset{w\neq u,v}{{\text{max}}}\frac{1}{K_2}\bigg| \big[A\EE(ZZ^T)A^T \big]_{uw} - \big[A\EE(ZZ^T)A^T \big]_{vw} \bigg|\\ &\hspace{5cm}\text{ (}\EE(\Gamma B(B^TB)^{-2}B^T\Gamma^T) ~\text{is diagonal)}\\
     &= \underset{w\neq u,v}{{\text{max}}}\frac{1}{K_2}\bigg| \sum_{t=1}^{K_2}\EE\big\{(Z_{r(u)t}-Z_{r(v)t})Z_{r(w)t}\big\} \bigg|\\
     &=\underset{w\neq u,v}{{\text{max}}}\bigg| \EE\big\{(Z_{r(u)1}-Z_{r(v)1})Z_{r(w)1}\big\} \bigg|.
\end{align*}
and from the definition of $C(\epsilon)$, it is apparent that the minimum of the maximum difference between rows is $2\epsilon$, that is
\begin{align}
    \MCOD(\Sigma_{p,W_O})=2\epsilon. \label{MCOD}
\end{align}
%Now, we need to look at the quantity $||X||_{W_I}$. Recall that $||X||_{W_I}=\frac{\sqrt{K_2}}{q}\cdot \underset{1\leq k\leq p}{\max}||\Var(X_{k\cdot})||_F$. After some tedious calculation, we have
%\begin{align}
%||X||_{W_I}&=\frac{\sqrt{K_2}}{q}\sqrt{\sum_{t=1}^{K_2}m_q^{(t)}(2 +  \sigma^2_{(t)})^2  +\sum_{t=1}^{K_2}m_q^{(t)}(m_q^{(t)}-1)\cdot 4} \nonumber \\
%&=\frac{\sqrt{K_2}}{q}\sqrt{\bigg[4\sum_{t=1}^{K_2}(m_q^{(t)})^2+\sum_{t=1}^{K_2}m_q^{(t)}\sigma^4_{(t)}+4\sum_{t=1}^{K_2}m_q^{(t)}\sigma^2_{(t)}\bigg]}. \label{lowerexpression}
%\end{align}
Let us choose 
\begin{align}\label{eq_epsilon}
 \epsilon = \sqrt{\bigg(\frac{\log(\frac{p}{3})}{n}\bigg)\cdot\bigg(\frac{1}{\sum_{t=1}^{K_2}\frac{m_q^{(t)}}{\sigma^2_{(t)}}}\bigg)\cdot\bigg(\frac{2e}{2e+1}\bigg)}.
\end{align}
The choice of $\epsilon$ will be explained later.

Furthermore, we can show that
\begin{align*}
||X||_{p,W_O}&=\frac{1}{\sqrt{K_2}}\cdot\sqrt{4K_2 + \sum_{t=1}^{K_2}\bigg[\frac{4\sigma^2_{(t)}}{|[t]|}+\frac{\sigma^4_{(t)}}{|[t]|^2}\bigg]}.
\end{align*}
So, getting rid of the term $\frac{4\sigma^2_{(t)}}{|[t]|}$, and using $\epsilon$ in (\ref{eq_epsilon}) and $m_q^{(t)}=\frac{q}{K_2}$, $\sigma_{(t)}=\sigma=O(\sqrt{\frac{q}{K_2}})$, we get the following inequalities:
\begin{align*}
    \frac{\MCOD(\Sigma_{p,W_O})}{||X||_{p,W_O}} ~&=~ \frac{2\epsilon}{\frac{1}{\sqrt{K_2}}\cdot\sqrt{4K_2 + \sum_{t=1}^{K_2}\bigg[\frac{4\sigma^2_{(t)}}{|[t]|}+\frac{\sigma^4_{(t)}}{|[t]|^2}\bigg]}}\\
    ~&=~\frac{2\sqrt{\bigg(\frac{\log(\frac{p}{3})}{n}\bigg)\cdot\bigg(\frac{1}{\sum_{t=1}^{K_2}\frac{m_q^{(t)}}{\sigma^2_{(t)}}}\bigg)\cdot\bigg(\frac{2e}{2e+1}\bigg)}}{\frac{1}{\sqrt{K_2}}\cdot\sqrt{4K_2 + \sum_{t=1}^{K_2}\bigg[\frac{4\sigma^2_{(t)}}{|[t]|}+\frac{\sigma^4_{(t)}}{|[t]|^2}\bigg]}}\\
    ~&=~ c\cdot\sqrt{\frac{\log p}{n}}\cdot \frac{1}{\sqrt{4\sum_{t=1}^{K_2}\frac{q}{K_2\sigma^2} ~+~ (\sum_{t=1}^{K_2}\frac{q}{K_2\sigma^2})(\sum_{t=1}^{K_2}[\frac{4\sigma^2}{q}+\frac{K_2\sigma^4}{q^2}])}}\\
    ~&=~ c\cdot\sqrt{\frac{\log p}{n}}\cdot \frac{1}{\sqrt{\frac{4q}{\sigma^2}+(\frac{q}{\sigma^2})(\frac{4K_2\sigma^2}{q} + \frac{K_2^2\sigma^4}{q^2})}}\\
    ~&\geq~ c\cdot \sqrt{\frac{\log p}{n}} \cdot \frac{1}{\sqrt{3}\sqrt{\frac{4q}{\sigma^2} \bigvee 4K_2 \bigvee \frac{\sigma^2 K_2^2}{q}}}\\
    ~&=~c'' \cdot\sqrt{\frac{\log p}{nK_2}} 
\end{align*}
%Changed the proof because of an error in lower bound logic.
%\begin{align*}
% \frac{\mathrm{MCOD}(\Sigma_{p,W_O})}{||X||_{p,W_O}}&\leq\sqrt{K_2}\frac{2\epsilon}{\sqrt{4K_2+\sum_{t=1}^{K_2}\frac{\sigma^4_{(t)}}{(m_q^{(t)})^2}}} \\
%  &\leq \sqrt{K_2} \sqrt{\bigg(\frac{\text{log}(p/3)}{n}\bigg)\bigg(\frac{2e}{2e+1}\bigg)\bigg(\frac{1}{(\sum_{t=1}^{K_2}\frac{m_q^{(t)}}{\sigma^2_{(t)}})(4K_2+\sum_{t=1}^{K_2}\frac{\sigma^4_{(t)}}{(m_q^{(t)})^2})}\bigg) }\\
% &\leq c\sqrt{\frac{\log(\frac{p}{3})}{nK_2}},
%\end{align*}
where for our construction we use the noise variance setting of $\sigma=O\big(\sqrt{m_q}\big)=O\Big(\sqrt{\frac{q}{K_2}}\Big)$ in the last equality. This ensures that the lower bound is as tight as possible. This implies that our constructed $\Sigma$ belongs to the parameter space $M_O(p,q,K_1,K_2,\eta)$ for some $\eta$ s.t. $\eta \leq c'' \sqrt{\frac{\log p}{nK_2}}$.

%Using (\ref{MCOD}) and (\ref{lowerexpression}), for $\epsilon$ in (\ref{eq_epsilon}) and $m_q^{(t)}=\frac{q}{K_2}$, $\sigma_{(t)}=\sigma$ we have
%\begin{align*}
% \frac{\mathrm{MCOD}(\Sigma_{p,W_O})}{||X||_{W_O}}&=\frac{2\epsilon}{\sqrt{\frac{K_2(4\sum_{t=1}^{K_2}(m_q^{(t)})^2+\sum_{t=1}^{K_2}m_q^{(t)}\sigma^4_{(t)}+4\sum_{t=1}^{K_2}m_q^{(t)}\sigma^2_{(t)})}{q^2}}}\\
% &=c\cdot\sqrt{\frac{\log(\frac{p}{3})}{n}}\frac{1}{\sqrt{\frac{K_2(4\sum_{t=1}^{K_2}(m_q^{(t)})^2+\sum_{t=1}^{K_2}m_q^{(t)}\sigma^4_{(t)}+4\sum_{t=1}^{K_2}m_q^{(t)}\sigma^2_{(t)})}{q^2}(\sum_{t=1}^{K_2}\frac{m_q^{(t)}}{\sigma^2_{(t)}})}}\\
% &=c\cdot\sqrt{\frac{\log(\frac{p}{3})}{n}}\sqrt{\frac{1}{\frac{1}{\sigma^2}(4q+K_2(\sigma^4+4\sigma^2))}}\\
% &\geq c\cdot\sqrt{\frac{\log(\frac{p}{3})}{n}}\sqrt{\frac{1}{(\sigma^2 \bigvee \frac{4}{\sigma^2} \bigvee 4)\cdot q}}\\
% &=c'\cdot\sqrt{\frac{\log p}{nq}}
%\end{align*} 
%for some universal constants $c,c'>0$. In the last equality, in order to make the lower bound as tight as possible, we use the $\sigma = O(1)$ homogeneous variance setting for our construction. This implies that our constructed $\Sigma$ belongs to the parameter space $M_I(p,q,K_1,K_2,\eta)$ for some $\eta$ s.t. $\eta \leq c' \sqrt{\frac{\log p}{nq}}$. 

Finally, we are ready to invoke Lemma \ref{Birge}. Recall that $KL(\Sigma^{'},\Sigma)$ is upper bounded in (\ref{eq_KLD}). Since we choose $\epsilon$ in (\ref{eq_epsilon}), we obtain 
$$
1-\frac{KL(\Sigma^{'},\Sigma)}{\log[(p/3)^2]} 
\geq 1- \frac{2n\epsilon^2\sum_{t=1}^{K_2}\frac{m_q^{(t)}}{\sigma^2_{(t)}}}{\log[(p/3)^2]}=\frac{1}{2e+1}.
$$ 
As a result, Lemma \ref{Birge} implies
$$
\underset{\Sigma \in M_I(p,q,K_1,K_2,\eta)}{\sup}\PP_{\Sigma}(\widehat{\cG} \neq \cG^{*}) \geq \frac{1}{2e+1},
$$
for any $\hat \cG$, which completes the proof of Theorem \ref{thm_lowerbound_2}.

%\subsubsection{Proof of Theorem \ref{thm_lowerbound_2}}
%We use the same construction from the proof of Theorem \ref{thm_lowerbound_1}. The only difference here is that our model class is defined with respect to $\Sigma_{p,W_O}$ instead of $\Sigma_{p,W_I}$. However, by the below reasoning, this also doesn't make a difference. Recall that for any index $u$ of rows of $X$, we use $r(u)$ to denote the row cluster label. It can be shown that 
%\begin{align*}
%    \mathrm{COD}_{\Sigma_{p,W_I}}(u,v) &= \underset{w\neq u,v}{{\text{max}}}\frac{1}{q}\bigg| \big[A\EE(ZB^TBZ^T)A^T + \EE(\Gamma\Gamma^T)\big]_{uw} - \big[A\EE(ZB^TBZ^T)A^T + \EE(\Gamma\Gamma^T)\big]_{vw} \bigg|\\
%    &= \underset{w\neq u,v}{{\text{max}}}\frac{1}{q}\bigg| \big[A\EE(ZB^TBZ^T)A^T \big]_{uw} - \big[A\EE(ZB^TBZ^T)A^T \big]_{vw} \bigg| \hspace{0.3cm}\text{ (}\EE(\Gamma\Gamma^T) ~\text{is diagonal)}\\
%    &=\underset{w\neq u,v}{{\text{max}}}\frac{1}{q}\bigg| \sum_{t=1}^{K_2}\EE\big\{(Z_{r(u)t}-Z_{r(v)t})Z_{r(w)t}\cdot |[t]|\big\} \bigg|\\
%    &=\underset{w\neq u,v}{{\text{max}}}\frac{1}{q}\bigg| \bigg(\sum_{t=1}^{K_2}|[t]|\bigg)\EE\big\{(Z_{r(u)1}-Z_{r(v)1})Z_{r(w)1} \big\} \bigg|\\
%    &=\underset{w\neq u,v}{{\text{max}}}\bigg| \EE\big\{(Z_{r(u)1}-Z_{r(v)1})Z_{r(w)1}\big\} \bigg|, 
%\end{align*}
%where the fourth equality holds because the columns of $Z$ are iid. Similar, we get

 %The rest of the proof is identical to the proof of Theorem \ref{thm_lowerbound_1} and therefore is omitted. 

\subsubsection{Proof for Theorem \ref{gen_minimax} (Minimax Lower Bound with a Perturbed $\bar{B}$)} \label{proof_gen_minimax}
\begin{proof}
To study the lower bound over $M_{\bar W}$, we follow the same construction as in the proof for Theorem \ref{thm_lowerbound_2}. We emphasize that all the parts leading up to calculating the KL - divergence is the same regardless of the value of $\bar{B}$ as the KL-divergence is related to the model on the population level. Since the true column clustering structure $B$ does not change, the derivations up to this point remain unchanged. The part that does indeed change is the calculation of $\frac{\text{MCOD}(\Sigma_{p,\bar{W}})}{||X||_{\bar{W}}}$ as both the MCOD value and $||X||_{\bar{W}}$ depend on $\bar{B}$. We can show that
\begin{align*}
    \Sigma_{p,\bar{W}} ~=~ \EE(X\bar{W}X^T) ~&=~ \EE\Big[(AZB^T+\Gamma)\bar{W}(BZ^TA^T + \Gamma^T)\Big] \\
    ~&=~ \EE\big(AZB^T\bar{W}BZ^TA^T\big) + \EE\big(\Gamma \bar{W} \Gamma^T\big)\\
    ~&=~ \frac{1}{s} \EE \Big[AZB^T\bar{B}(\bar{B}^T\bar{B})^{-2}\bar{B}^TZ^TA^T\Big] + \EE(\Gamma \bar{W}\Gamma^T)\\
    ~&=~ \frac{1}{s}\EE\big( AZGG^TZ^TA^T\big) + \EE\big(\Gamma \bar{W}\Gamma^T\big),
\end{align*}
where $G=B^T\bar{B}(\bar{B}^T\bar{B})^{-1}$.   Thus  
\begin{align*}
    \COD_{\Sigma_{p,\bar{W}}}(u,v) ~&=~ \underset{w \neq u,v}{\max}\Big|\big(\Sigma_{p,\bar{W}}\big)_{uw} - \big(\Sigma_{p,\bar{W}}\big)_{vw} \Big|\\
    ~&=~ \frac{1}{s}\underset{w \neq u,v}{\max} \Bigg| \EE \bigg\{\sum_{t_1,t_2}^{K_2} Z_{r(u)t_1}Z_{r(w)t_2}\big(GG^T\big)_{t_1 t_2} - Z_{r(v)t_1}Z_{r(w)t_2}\big(GG^T\big)_{t_1 t_2}\bigg\} \Bigg|\\
    ~&~ \hspace{3cm} \text{since  } \EE(\Gamma \hat{W}\Gamma^T) \text{ is diagonal}\\
    ~&=~ \frac{1}{s}\underset{w \neq u,v}{\max} \Bigg| \sum_{t_1, t_2}^{K_2} \big(GG^T\big)_{t_1 t_2} \cdot \EE\Big\{ Z_{r(u)t_1}Z_{r(w)t_2} - Z_{r(v)t_1}Z_{r(w)t_2}\Big\}\Bigg|\\
    ~&=~ \frac{1}{s}\underset{w \neq u,v}{\max} \Bigg| \sum_{t=1}^{K_2} \big(GG^T\big)_{tt} \cdot \EE\Big\{ Z_{r(u)t}Z_{r(w)t} - Z_{r(v)t}Z_{r(w)t}\Big\}\Bigg|\\
    ~&~ \hspace{3cm} \text{since in our construction, } Z_{\cdot 1},~Z_{\cdot 2},~Z_{\cdot 3} \text{ are i.i.d.}\\
    ~&=~ \frac{1}{s}\underset{w \neq u,v}{\max} \Bigg| \sum_{t=1}^{K_2} \big(GG^T\big)_{tt} \cdot \EE\Big\{ Z_{r(u)1}Z_{r(w)1} - Z_{r(v)1}Z_{r(w)1}\Big\}\Bigg|\\
    ~&=~ \frac{1}{s}\underset{w \neq u,v}{\max} \Bigg| \big|\big|G\big|\big|_F^2 \cdot \EE\Big\{ Z_{r(u)1}Z_{r(w)1} - Z_{r(v)1}Z_{r(w)1}\Big\}\Bigg|.
\end{align*}
Thus, we have 
$$
\MCOD(\Sigma_{p,\bar{W}}) ~=~ \frac{1}{s}\cdot \big|\big|G\big|\big|_F^2 \cdot 2\epsilon
$$ 
where the $\epsilon$ is from our construction. 
Note that all of the entries of $G$ are non-negative and all the column sums are 1. Thus, every entry in $G$ is $\leq 1$, and the entries in $G$ add up to $s$. We can conclude that $||G||_F^2 \leq s$. Thus, $\frac{1}{s}||G||_F^2 \leq 1$. It is apparent that the $\MCOD$ value potentially decreases with a weight $\bar{W}$ with an incorrect estimated column cluster structure. When $\bar{B}=B$, then $G=I_{K_2}$ and $s=K_2$, and the $\MCOD$ value reduces to $2\epsilon$. Also, when $\bar{B}=I_q$, then $G=B^T$, $s=q$, and the $\MCOD$ value reduces to $2\epsilon$ as well. It appears that these two are the exception, as in other cases, the $\MCOD$ value becomes strictly smaller than $2\epsilon$. For example, when $\bar{B}=\mathbf{1}_{q}$ (when we put all the columns in the same column cluster) we have $s=1$, $G=\frac{1}{K_2}\cdot\mathbf{1}_{K_2}$ and $\frac{1}{s}||G||_F^2 = \frac{1}{K_2}$ which is strictly smaller than 1. The $\MCOD$ value becomes $\frac{2\epsilon}{K_2}$. 

Now let's look at $||X||_{\bar{W}}$:
\begin{align*}
    ||X||_{\bar{W}} ~&=~ \sqrt{K_2} \cdot \underset{a}{\max} \Big|\Big|\text{Var}(X_{a\cdot}\hat{L}) \Big|\Big|_F \\
    ~&=~ \sqrt{K_2} \cdot \underset{a}{\max} \Big|\Big|\text{Var}(Z_{r(a)\cdot}B^T\hat{L})+ \text{Var}(\Gamma_{a\cdot}\hat{L})\Big|\Big|_F\\
    ~&=~ \sqrt{K_2} \cdot \underset{a}{\max} \Big|\Big|\hat{L}^T B\EE(Z_{r(a)\cdot}^T Z_{r(a)\cdot})B^T\hat{L} + \text{Var}(\Gamma_{a\cdot}\hat{L})\Big|\Big|_F\\
    ~&=~ \sqrt{K_2} \cdot \underset{a}{\max} \Big|\Big|\hat{L}^T B\EE(Z_{r(a)1}^2)\cdot I_{K_2}B^T\hat{L} + \text{Var}(\Gamma_{a\cdot}\hat{L})\Big|\Big|_F\\
    ~&~ \hspace{3cm} \text{since in our construction, } Z_{\cdot 1},~Z_{\cdot 2},~Z_{\cdot 3} \text{ are i.i.d.}\\
    ~&=~ \sqrt{K_2} \cdot \underset{a}{\max}\Big|\Big|\EE(Z_{r(a)1}^2)\cdot \frac{1}{s}G^TG + \text{Var}(\Gamma_{a\cdot}\hat{L})\Big|\Big|_F\\
     ~&~ \hspace{3cm} \text{since } \hat{L}=\frac{1}{\sqrt{s}}\bar{B}(\bar{B}^T\bar{B})^{-1} \text{ and } G = \sqrt{s} B^T \hat{L} \\
     ~&=~ \sqrt{K_2} \cdot \underset{a}{\max}\Big|\Big|\frac{2}{s}G^TG + \text{Var}(\Gamma_{a\cdot}\hat{L})\Big|\Big|_F\\
     ~&~ \hspace{3cm} \text{since all the entries in } \frac{1}{s}G^TG \text{ and } \text{Var}(\Gamma_{a\cdot}\hat{L}) \text{ are nonnegative }\\
     ~&~ \hspace{3cm} \text{(we can take the maximum diagonal value in $C(\epsilon)$.)}\\
     ~& \leq ~ \frac{\sqrt{K_2}}{s} \cdot \underset{a}{\max} \sqrt{8\big|\big| G^TG \big|\big|_F^2 + 2 \sum_{r=1}^s \Big( \frac{\sum_{j \in \hat{[r]}}\sigma^2_{aj}}{|\hat{[r]}|^2}\Big)^2}\\
     ~&~ \hspace{1.5cm} \text{where $\hat{[r]}$ denotes the r-th estimated column cluster and $\sigma_{aj}^2 = \text{Var}(\Gamma_{aj})$}\\ 
     ~&=~ \frac{\sqrt{K_2}}{s} \cdot \sqrt{8\big|\big| G^TG \big|\big|_F^2 + \Big(\frac{q}{K_2}\Big)^2 \cdot 2\sum_{r=1}^s \frac{1}{|\hat{[r]}|^2}}\\
     ~&~ \hspace{3cm} \text{ since we assume $\sigma_{aj}^2 = \sigma^2 = m_q = \frac{q}{K_2}$ in our construction}\\
     ~&\leq~ c \cdot \frac{\sqrt{K_2}}{s} \cdot \sqrt{\big|\big| G^TG \big|\big|_F^2 + \Big(\frac{q}{K_2}\Big)^2 \cdot \frac{s}{\bar{m}_q^2}}. \\
      ~&~ \hspace{3cm} \text{ where $\bar{m}_q^2$ denotes the smallest estimated column cluster size}
\end{align*}
Finally, we can derive the lower bound rate for $\frac{\MCOD(\Sigma_{p,\bar{W}})}{||X||_{\bar{W}}}$:
\begin{align*}
    \frac{\MCOD(\Sigma_{p,\bar{W}})}{||X||_{\bar{W}}} ~&\geq~ c' \cdot 2\epsilon \cdot \frac{\frac{1}{s}||G||_F^2}{\frac{\sqrt{K_2}}{s} \cdot \sqrt{\big|\big| G^TG \big|\big|_F^2 + \Big(\frac{q}{K_2}\Big)^2 \cdot \frac{s}{\bar{m}_q^2}}} \\
    ~&=~ c'' \cdot \sqrt{\frac{\log p}{nK_2}} \cdot \frac{||G||_F^2}{\sqrt{\big|\big| G^TG \big|\big|_F^2 + \Big(\frac{q}{K_2}\Big)^2 \cdot \frac{s}{\bar{m}_q^2}}} \cdot \frac{1}{\sqrt{\sum_{t=1}^{K_2} \frac{m_q^{(t)}}{\sigma^2_{(t)}}}} \\
    ~&= c'' \cdot \sqrt{\frac{\log p}{nK_2}} \cdot \frac{||G||_F^2}{\sqrt{K_2} \cdot \sqrt{\big|\big| G^TG \big|\big|_F^2 + \Big(\frac{q}{K_2}\Big)^2 \cdot \frac{s}{\bar{m}_q^2}}}.\\
     ~&~ \hspace{2.5cm} \text{ since we assume $m_q^{(t)} = m_q = \sigma^2 =  \sigma^2_{(t)}$ in our construction.}
\end{align*}
This completes the proof of the lower bound. 
\end{proof}

\subsubsection{Proof of Theorem \ref{thm_lowerbound_row_column} (Joint Lower Bound) }
\label{joint_lower}
\begin{proof}
The extension from considering just one direction (either the rows or columns) is quite trivial. 
\begin{align*}
    \PP_{\Sigma}\Big((\widehat{\cG}^{(r)},\widehat{\cG}^{(c)})\neq (\cG^{(r)},\cG^{(c)})\Big) ~&=~ \PP_{\Sigma}\Big( \{\widehat{\cG}^{(r)} \neq \cG^{(r)}\} \cup \{\widehat{\cG}^{(c)} \neq \cG^{(c)}\} \Big) \\
    ~&=~ \PP_{\Sigma}\Big( \widehat{\cG}^{(r)} \neq \cG^{(r)}\Big) ~+~ \PP_{\Sigma}\Big( \widehat{\cG}^{(c)} \neq \cG^{(c)}\Big) \\
    & \hspace{2cm}~-~ \PP_{\Sigma}\Big( \{\widehat{\cG}^{(r)} \neq \cG^{(r)}\} \cap \{\widehat{\cG}^{(c)} \neq \cG^{(c)}\} \Big)\\
    ~&\geq~ \PP_{\Sigma}\Big( \widehat{\cG}^{(r)} \neq \cG^{(r)}\Big) \\
    ~&\geq~ \frac{1}{7},
\end{align*}
where the last line holds from the minimax lower bound construction for Theorem \ref{thm_lowerbound_2}. The only thing left to check for the proof is that the constructed $\Sigma$ in the previous minimax lower bound proof is actually an element of \textbf{both} parameter spaces - $M_{O,(r)}$ and $M_{O,(c)}$ - so that the supremum logic is valid. We will show that the $\Sigma$ constructed in the minimax lower bound proof for the row clusters also satisfies $\MCOD(\Sigma_{q,W_{O,(c)}})/\|X\|_{W_{O,(c)}}\geq \eta_{(c)}$ for $\eta_{(c)} \asymp\sqrt{\frac{\log q}{nK_1}}$. We have
\begin{align*}
    \Sigma_{q,W_{O,(c)}} ~=~ \EE(X^TW_{O,(c)}X) ~&=~ \EE\Big[\big(BZ^TA^TW_{O,(c)}AZB^T\big) + \big(\Gamma^T W_{O,(c)} \Gamma\big)  \Big] \\
    ~&=~ \EE\Big[ \big(BZ^TA^TA(A^TA)^{-2}A^TAZB^T \big)/K_1 + \Gamma^T W_{O,(c)} \Gamma\Big] \\
    ~&=~ \frac{1}{K_1}\EE(BZ^TZB^T) ~+~ \EE(\Gamma^T W_{O,(c)}\Gamma)\\
    \COD_{\Sigma_{q,W_{O,(c)}}} (u,v) ~&=~ \underset{w \neq u,v}{\max} \bigg| \Big[ \Sigma_{q,W_{O,(c)}}\Big]_{uw} ~-~ \Big[ \Sigma_{q,W_{O,(c)}}\Big]_{vw} \bigg|\\
    ~&=~ \underset{w \neq u,v}{\max} \bigg| \frac{1}{K_1}\Big[B\EE(Z^TZ)B^T\Big]_{uw} ~-~ \frac{1}{K_1}\Big[B\EE(Z^TZ)B^T\Big]_{vw} \bigg| \\
    ~&=~ \frac{1}{K_1} \underset{w \neq u,v}{\max} \bigg| \EE\Big( Z_{\cdot c(u)}^T Z_{\cdot c(w)} ~-~ Z_{\cdot c(v)}^T Z_{\cdot c(w)} \Big) \bigg| \\
    ~&=~ \frac{1}{K_1} \underset{w \neq u,v}{\max} \Bigg| \sum_{t=1}^{K_1} \EE\Big(Z_{t,c(u)}Z_{t,c(w)} ~-~ Z_{t,c(v)}Z_{t,c(w)} \Big) \Bigg|\\
    ~&=~ \begin{cases}
    \frac{1}{K_1} \underset{1 \leq u \leq q}{\max} \Big| \sum_{t=1}^{K_1} \EE\big( Z_{t,c(u)}^2\big) \Big| & \text{(If $u \underset{\cG}{\nsim} v$)} \\
    0 & \text{(If $u \underset{\cG}{\sim} v$)}
    \end{cases}\\
    & \hspace{0.2cm} \text{(Note the columns of $Z$ are independent in our construction.)}\\
    ~&=~ \begin{cases} 
    \frac{1}{K_1} \Big| \sum_{t=1}^{K_1} \EE\big( Z_{t,1}^2\big) \Big|  & \text{(If $u \underset{\cG}{\nsim} v$)} \\
    0 & \text{(If $u \underset{\cG}{\sim} v$)}
    \end{cases}\\
     & \hspace{0.2cm} \text{(Again, from column independence of $Z$ in our construction.)}\\
    \MCOD(\Sigma_{q,W_{O,(c)}})~&=~ \underset{u \underset{\cG}{\nsim} v}{\min}~~\COD_{\Sigma_{q,W_{O,(c)}}} (u,v)\\
    ~&=~  \frac{1}{K_1} \Bigg| \sum_{t=1}^{K_1} \EE\Big( Z_{t,1}^2\Big) \Bigg| \\
    ~&=~ \frac{2\epsilon + 2}{3} ~~~~~~~~\text{(From our construction with $C(\epsilon)$.)}\\
    ~&\geq~ \frac{2}{3}\\
    ||X||_{W_{O,(c)}} ~&=~ \sqrt{K_1} \cdot \underset{1 \leq b \leq q}{\max}~ \Big|\Big|L^T_{(c)} \text{Var}(X_{\cdot b}) L_{(c)}\Big|\Big|_F \\
    ~&=~ \sqrt{K_1} \cdot \underset{1 \leq b \leq q}{\max}~ \Big|\Big|\text{Var}\big(L^T_{(c)}X_{\cdot b}\big)\Big|\Big|_F 
    \end{align*}
    \newpage
    \begin{align*}
    ~&=~ \sqrt{K_1} \cdot \underset{1 \leq b \leq q}{\max}~ \Big|\Big|  \text{Var}\big(\frac{1}{\sqrt{K_1}}Z_{\cdot c(b)} \big) ~+~  \text{Var}\big( L_{(c)}^T \Gamma_{\cdot b}\big) \Big|\Big|_F \\
    ~&=~ \sqrt{K_1} \cdot \underset{1 \leq b \leq q}{\max}~ \Big|\Big| \frac{1}{K_1}\EE\big(Z_{\cdot c(b)}Z_{\cdot c(b)}^T\big)  ~+~ \text{Var}\big( L_{(c)}^T \Gamma_{\cdot b}\big)\Big|\Big|_F \\
    ~&=~ \sqrt{K_1} \Big|\Big| \frac{1}{K_1} C(\epsilon) ~+~ \frac{\sigma^2}{p} I_{K_1}\Big|\Big|_F \\
    ~&=~ \sqrt{K_1} \cdot \Bigg|\Bigg| \begin{bmatrix}
    \frac{\epsilon}{K_1} + \frac{\sigma^2}{p} & \frac{\epsilon - \epsilon^2}{K_1} & -\frac{\epsilon}{K_1} \\
    \frac{\epsilon - \epsilon^2}{K_1} & \frac{\epsilon}{K_1} + \frac{\sigma^2}{p} & \frac{\epsilon}{K_1} \\
    -\frac{\epsilon}{K_1} & \frac{\epsilon}{K_1} & \frac{2}{K_1} + \frac{\sigma^2}{p}
    \end{bmatrix} \Bigg|\Bigg|_F \\
    ~&\leq~ \sqrt{K_1} \cdot \sqrt{ 4\cdot\Big(\frac{2}{K_1}\Big)^2 \cdot K_1^2} \\
    ~&=~ 4 \sqrt{K_1} \\
    \frac{\MCOD(\Sigma_{q,W_{O,(c)}})}{||X||_{W_{O,(c)}}} ~&\geq~ c\cdot \frac{(2/3)}{\sqrt{K_1}} = c'\cdot\sqrt{\frac{1}{K_1}} \\
    ~& >>~ \sqrt{\frac{\log q}{nK_1}}~~~~~~ \text{since $\log q = o(n)$}.
\end{align*}
We have shown that the $\Sigma$ constructed for the minimax lower bound proof for the \textbf{rows} is also an element of $M_{O,(c)}$. The same argument can be made symmetrically to show that the $\Sigma$ constructed for the minimax lower bound proof for the \textbf{columns} is also an element of $M_{O,(r)}$. Thus, the proof for joint minimax optimality is complete. 
\end{proof}

\subsection{Proofs for the Matrix Normal Model}
\subsubsection{Proof of Proposition \ref{matrixnormal}} \label{matnor}

\begin{proof}
For any rows $w,x$ of $\Sigma_{p,W_O}$, without loss of generality, we can assume they belong to clusters $i,h$. Then
\begin{align}
    \COD_{\Sigma_{p,W_O}}(w,x)
     &= \underset{1 \leq c \leq K_1}{ {\text{max}}}\frac{1}{K_2}\bigg| \sum_{l=1}^{K_2}\EE\big\{(Z_{il}-Z_{hl})Z_{cl}\big\} \bigg| \nonumber\\
     &= \underset{1 \leq c \leq K_1}{ {\text{max}}}\frac{1}{K_2}\bigg| \sum_{l=1}^{K_2}(U_{ic}-U_{hc})V_{ll} \bigg| \nonumber\\
     &= \underset{1 \leq c \leq K_1}{ {\text{max}}}\frac{1}{K_2}\big| U_{ic}-U_{hc} \big| \bigg|\sum_{l=1}^{K_2}V_{ll} \bigg| \nonumber \\
     &= \underset{1 \leq c \leq K_1}{ {\text{max}}}\frac{1}{K_2}\big| U_{ic}-U_{hc} \big| \mathrm{tr}(V). \nonumber
\end{align}
Looking at the cluster separation condition first, we need this $\COD$ value to be greater than $c_o \eta \cdot||X||_{W_O} = c_0\cdot c_1\sqrt{\frac{\log p}{nK_2}}\cdot||X||_{W_O}$ for all $w,x$ (where $c_1$ is a universal constant and $c_0$ is an arbitrary constant $\geq 4$.). We know that $||X||_{W_O} \leq \frac{1}{\sqrt{K_2}}\underset{a}{\max}||\Var(Z_{r(a)\cdot})||_F + \sqrt{K_2} \underset{a}{\max}||\Var(\Gamma_{a\cdot}L)||_F$ from the definition of $||X||_{W_O}$. In the matrix normal setting, we have
\begin{align*}
    \Var(Z_{r(a)\cdot}) &= \begin{bmatrix}
    U_{r(a)r(a)}\cdot V_{11} & U_{r(a)r(a)}\cdot V_{12} & ... & ... & \\
    U_{r(a)r(a)}\cdot V_{21} & U_{r(a)r(a)}\cdot V_{22} & ... & ... & \\
    :& : & & & \\
    \end{bmatrix}= U_{r(a)r(a)} \cdot V
\end{align*}
and so, $||\Var(Z_{r(a)\cdot})||_F = U_{r(a)r(a)}\cdot ||V||_F$ and we have 
$$
\underset{a}{\max}||\Var(Z_{r(a)\cdot})||_F = ||\text{diag}(U)||_{\max} \cdot ||V||_{F}\leq ||\text{diag}(U)||_{\max} K_2^{1/2}C_{\max}. 
$$
It is also easily shown that 
$$
\max_a||\Var(\Gamma_{a\cdot}L)||_F=\max_a\frac{1}{K_2}\sqrt{\sum_{t=1}^{K_2}\frac{(\sum_{j \in [t]}\sigma^2_{aj})^2}{|[t]|^4}}=C^{1/2}_K/K_2^{1/2}. 
$$
Rearranging with algebra gives us the cluster separation condition.

Now, let's consider the stability condition. Again, we need only consider rows in different row clusters when looking at the $\MCOD$ value. That is why we assume $i \neq h$. Now for $\MCOD(\Sigma_{p,\hat{W}_O})$, we would need to look at the following equivalent expressions:
\begin{align}
\COD_{\Sigma_{p,\hat{W}_O}}(w,x)
     &= \underset{1\leq c \leq K_1}{ {\text{max}}} \frac{1}{s}\bigg| \EE\bigg( \sum_{l=1}^{K_2} (Z_{il}-Z_{hl})Z_{cl}(\sum_{j=1}^{s}\frac{|G_l \cap \hat{G}_{j}|^2}{|\hat{G}_{j}|^2}) \bigg| \hat{B} \bigg) \nonumber\\
&\hspace{2cm}+ \EE\bigg( \sum_{l' \neq l''}^{K_2} (Z_{il'}-Z_{hl'})Z_{cl''}(\sum_{j=1}^{s}\frac{|G_{l'} \cap \hat{G}_{j}|}{|\hat{G}_{j}|}\cdot\frac{|G_{l''} \cap \hat{G}_{j}|}{|\hat{G}_{j}|})\bigg| \hat{B} \bigg)  \bigg| \nonumber\\
&= \underset{1\leq c \leq K_1}{ {\text{max}}} \frac{1}{s}\bigg|  \sum_{l=1}^{K_2} (U_{ic}-U_{hc})V_{ll}(\sum_{j=1}^{s}\frac{|G_l \cap \hat{G}_{j}|^2}{|\hat{G}_{j}|^2}) \nonumber\\
&\hspace{2cm}+  \sum_{l' \neq l''}^{K_2} (U_{ic}-U_{hc})V_{l'l''}(\sum_{j=1}^{s}\frac{|G_{l'} \cap \hat{G}_{j}|}{|\hat{G}_{j}|}\cdot\frac{|G_{l''} \cap \hat{G}_{j}|}{|\hat{G}_{j}|})  \bigg|\nonumber\\
&= \underset{1\leq c \leq K_1}{ {\text{max}}} \frac{1}{s}\big|U_{ic}-U_{hc}\big|\bigg|  \sum_{l=1}^{K_2} V_{ll}(\sum_{j=1}^{s}\frac{|G_l \cap \hat{G}_{j}|^2}{|\hat{G}_{j}|^2}) \nonumber \\
&\hspace{2cm} +  \sum_{l' \neq l''}^{K_2} V_{l'l''}(\sum_{j=1}^{s}\frac{|G_{l'} \cap \hat{G}_{j}|}{|\hat{G}_{j}|}\cdot\frac{|G_{l''} \cap \hat{G}_{j}|}{|\hat{G}_{j}|})  \bigg|\nonumber\\
&= \underset{1\leq c \leq K_1}{ {\text{max}}} \frac{1}{s}\big|U_{ic}-U_{hc}\big|\big| \textrm{tr}(V^T\GG\GG^T)\big|, \nonumber
\end{align}
where for simplicity we drop the upperscript $(c)$ in $G_l$ and $\hat G_j$.  

Note that both $\COD_{\Sigma_{p,\hat{W}_O}}(w,x)$ and $\COD_{\Sigma_{p,W_O}}(w,x)$ are influenced by $i$ and $h$ ($i \neq h$) only through the common $|U_{ic}-U_{hc}|$ term, and so the $(i,h)$ pair that gives the minimum of the two $\COD$ expressions is the same.  
Thus, the following holds:
\[
\frac{\MCOD(\Sigma_{p,\hat{W}_O})}{\MCOD(\Sigma_{p,W_O})} = \frac{|\textrm{tr}(V^T\GG\GG^T)|}{|\mathrm{tr}(V)|} \cdot \frac{K_2}{s}.
\]
Thus,
\[
\Bigg\{\frac{\MCOD(\Sigma_{p,\hat{W}_O})}{\|X\|_{\hat {W}_O}}\Bigg\}\Bigg/\Bigg\{\frac{\MCOD(\Sigma_{p,W_O})}{\|X\|_{W_O}}\Bigg\}=\frac{|\textrm{tr}(V^T\GG\GG^T)|}{|\mathrm{tr}(V)|} \cdot \frac{K_2}{s}\cdot \frac{||X||_{W_{O}}}{||X||_{\hat{W}_O}} > \frac{4}{c_0}
\]
needs to hold. By Von Neumann's trace inequality, we have $|\textrm{tr}(V^T\GG\GG^T)| \geq |\textrm{tr}(V)\cdot\lambda_{\text{min}}(GG^T)|$, and so it suffices to have: 
\[
|\lambda_{\text{min}}(GG^T)| \cdot \frac{K_2}{s}\cdot \frac{||X||_{W_{O}}}{||X||_{\hat{W}_O}} > \frac{4}{c_0}.
\]
To show the above inequality, we need to further lower bound $\frac{||X||_{W_{O}}}{||X||_{\hat{W}_O}}$,
\begin{align*}
    \frac{||X||_{W_{O}}}{||X||_{\hat{W}_O}} &=\frac{\underset{a}{\max}~||\frac{1}{K_2}\Var(Z_{r(a)\cdot}) + \Var(\Gamma_{a\cdot}L)||_F}{\underset{a}{\max}~||\frac{1}{s}\Var(Z_{r(a)\cdot}G) + \Var(\Gamma_{a\cdot}\hat{L})||_F}\\
    &\geq \frac{\underset{a}{\max}~\sqrt{\frac{1}{K_2^2}||\Var(Z_{r(a)\cdot})||_F^2 + ||\Var(\Gamma_{a\cdot}L)||_F^2}}{\underset{a}{\max}~\frac{1}{s}||G^T\Var(Z_{r(a)\cdot})G||_F + \underset{a}{\max}~||\Var(\Gamma_{a\cdot}\hat{L})||_F}.
\end{align*}
For the term $||G^T\Var(Z_{r(a)\cdot})G||_F^2$ in the denominator, we have 
\begin{align}
||G^T\Var(Z_{r(a)\cdot})G||_F^2 &= \sum_{i=1}^{K_2} \lambda_i\{ (G^T\Var(Z_{r(a)\cdot}) G)^T(G^T\Var(Z_{r(a)\cdot})G) \}\label{f:op1}\\
&\leq \text{min}(s,K_2)\cdot||G^T\Var(Z_{r(a)\cdot}) G||_{\text{op}}^2 \label{f:op4}\\
& \leq \min(s,K_2)\cdot||G||^4_{\text{op}}\cdot ||\Var(Z_{r(a)\cdot})||^2_{\text{op}}, \nonumber
\end{align}
where (\ref{f:op1}) holds from the fact that \(||A||_F^2 = \text{tr}(A^TA) = \text{ the sum of the eigenvalues of } A^TA\), and (\ref{f:op4}) holds from the fact there can be at most min\((s,K_2)\) non-zero eigenvalues of \(G^T\Var(Z_{r(a)\cdot}) G\) because rank(\(G^T\Var(Z_{r(a)\cdot}) G\))\(\leq\) min\{rank(\(G\)),~rank(\(\Var(Z_{r(a)\cdot})\))\} and rank(\(G\)) \(\leq\) min(\(s,K_2\)). Using the similar argument, we also show that
$$
||\Var(Z_{r(a)\cdot})||_F^2\geq ||\text{diag}(U)||^2_{\min} \cdot ||V||^2_{F}\geq C^2_{\min}K_2||\text{diag}(U)||^2_{\min}. 
$$
Finally, noting that $||G||_{\text{op}}^2 = |\lambda_{\max}(GG^T)|$, and
$$
\max_a||\Var(\Gamma_{a\cdot}L)||_F=C^{1/2}_K/K_2^{1/2}, ~~ \max_a||\Var(\Gamma_{a\cdot}\hat L)||_F=C^{1/2}_s/s^{1/2},
$$
we have  
\begin{align*}
    \frac{||X||_{W_{O}}}{||X||_{\hat{W}_O}} 
    & \geq \frac{\sqrt{\frac{1}{K_2} C^2_{\min}\cdot||\text{diag}(U)||^2_{\min}+ C_K/K_2}}{\underset{a}{\max}~\frac{1}{s}\sqrt{\min(s,K_2)}\cdot|\lambda_{\max}(GG^T)|\cdot||\Var(Z_{r(a)\cdot})||_{\text{op}} + C^{1/2}_s/s^{1/2}}\nonumber\\
    & \geq \frac{\sqrt{\frac{1}{K_2} C^2_{\min}\cdot||\text{diag}(U)||^2_{\min}+ C_K/K_2}}{\frac{1}{s}\sqrt{\min(s,K_2)}\cdot|\lambda_{\max}(GG^T)| \cdot ||\text{diag}(U)||_{\max} \cdot C_{\max}+ C^{1/2}_s/s^{1/2}}.
\end{align*}
So, combining with the above inequality, we need to show the following condition:
\begin{align}\label{eq_proof_MN1}
    \frac{ K^{1/2}_2 |\lambda_{\min}(GG^T)|\sqrt{ C^2_{\min}\cdot||\text{diag}(U)||^2_{\min}+ C_K}}{\sqrt{\min(s,K_2)}\cdot |\lambda_{\max}(GG^T)| \cdot ||\text{diag}(U)||_{\max} \cdot  C_{\max}+ s^{1/2}C^{1/2}_s} ~\geq~  \frac{4}{c_0}.
\end{align}
To show (\ref{eq_proof_MN1}), let's consider the following two cases.\\ If $\sqrt{\min(s,K_2)}\cdot|\lambda_{\max}(GG^T)| ||\text{diag}(U)||_{\max} C_{\max}~\leq ~ C^{1/2}_ss^{1/2}$ holds, then 
\begin{align*}
    &\frac{ K^{1/2}_2 |\lambda_{\min}(GG^T)|\sqrt{ C^2_{\min}\cdot||\text{diag}(U)||^2_{\min}+ C_K}}{\sqrt{\min(s,K_2)}\cdot |\lambda_{\max}(GG^T)| \cdot ||\text{diag}(U)||_{\max} \cdot C_{\max}+ s^{1/2}C^{1/2}_s}\\
    &\geq  \frac{ K^{1/2}_2 |\lambda_{\min}(GG^T)|C_K^{1/2}}{2 s^{1/2}C^{1/2}_s}\geq~  \frac{4}{c_0},
\end{align*}
when $\frac{1}{\lambda_{\min}(GG^T)}\leq \frac{c_0C_K^{1/2}K_2^{1/2}}{8C^{1/2}_ss^{1/2}}$. In the second case $\sqrt{\min(s,K_2)}\cdot|\lambda_{\max}(GG^T)| ||\text{diag}(U)||_{\max} C_{\max}>  C^{1/2}_ss^{1/2}$, we can similarly show that
\begin{align*}
    &\frac{ K^{1/2}_2 |\lambda_{\min}(GG^T)|\sqrt{ C^2_{\min}||\text{diag}(U)||^2_{\min}+ C_K}}{\sqrt{\min(s,K_2)}\cdot |\lambda_{\max}(GG^T)| \cdot ||\text{diag}(U)||_{\max} \cdot C_{\max}+ s^{1/2}C^{1/2}_s}\\
    &\geq  \frac{ K^{1/2}_2 |\lambda_{\min}(GG^T)|C_{\min}||\text{diag}(U)||_{\min}}{2\sqrt{\min(s,K_2)}\cdot |\lambda_{\max}(GG^T)| \cdot ||\text{diag}(U)||_{\max} \cdot C_{\max}}\geq~  \frac{4}{c_0},
\end{align*}
when $\frac{\lambda_{\max}(GG^T)}{\lambda_{\min}(GG^T)}\leq \frac{c_0\cdot C_{\min}\cdot ||\text{diag}(U)||_{\min}}{8\cdot C_{\max}\cdot ||\text{diag}(U)||_{\max}}\sqrt{\frac{K_2}{\min (s, K_2)}}$. This completes the proof. 
\end{proof}

\subsubsection{Proof of Proposition \ref{pro_stability}} \label{proof_pro_stability}
\begin{proof}
Recall that our definition of $G$ from Section \ref{sec_example} is as follows:
\begin{align*}
    G ~=~ B^T \hat{B}(\hat{B}^T\hat{B})^{-1} ~&=~ \begin{bmatrix}
\frac{|G_1^{(c)} \cap \widehat{G}_{1}^{(c)}|}{|\widehat{G}_{1}^{(c)}|}&\frac{|G_1^{(c)} \cap \widehat{G}_{2}^{(c)}|}{|\widehat{G}_{2}^{(c)}|}&...&\frac{|G_1^{(c)} \cap \widehat{G}_{s}^{(c)}|}{|\widehat{G}_{s}^{(c)}|}\\
:&:&&:\\
\frac{|G_{K_2}^{(c)} \cap \widehat{G}_{1}^{(c)}|}{|\widehat{G}_{1}^{(c)}|}&\frac{|G_{K_2}^{(c)} \cap \widehat{G}_{2}^{(c)}|}{|\widehat{G}_{2}^{(c)}|}&...&\frac{|G_{K_2}^{(c)} \cap \widehat{G}_{s}^{(c)}|}{|\widehat{G}_{s}^{(c)}|}
\end{bmatrix} 
\end{align*}
Note that $\hat{G}_{1}^{(c)}=\hat{G}_{2}^{(c)}=...=\hat{G}_{s}^{(c)}=1$ when $\hat{B}=I_q$. It is easily seen that $GG^T = \text{diag}(m_q^{(1)},m_q^{(2)},~...~,m_q^{(K_2)})$ (up to order) where $m_q^{(i)}$ denotes the $i$-th smallest column cluster size. Thus, $\lambda_{\min}(GG^T) = m_q$ and $\lambda_{\max}(GG^T) = M_q$, where $m_q$ and $M_q$ denote the smallest and largest column cluster sizes, respectively.

Recall we have defined:
\begin{align*}
C_K ~&=~ \max_{1\leq a\leq p}\frac{1}{K_2}\sum_{t=1}^{K_2}\frac{(\sum_{j \in [t]}\sigma^2_{aj})^2}{|[t]|^4}\\
C_s ~&=~ \max_{1\leq a\leq p}\frac{1}{s}\sum_{t=1}^{s}\frac{(\sum_{j \in [\hat t]}\sigma^2_{aj})^2}{|[\hat t]|^4}.
\end{align*}
Under the assumption that the noise is homogeneous and the cluster size is balanced (i.e., for all $i,j$, $\EE(\Gamma_{ij})=\sigma^2=O(1)$, $\frac{M_q}{m_q}\leq C$ for some constant $C\geq 1$), we have that $C_K = \frac{1}{K_2}\sum_{t=1}^{K_2} \frac{\sigma^4}{|[t]|^2} \geq \frac{\sigma^4}{M_q^2}$. Also, by the formulation of our weight $\hat{W}=\frac{1}{q}I_q$, we have that $C_s=\sigma^4$. 

In order to satisfy the stability condition in Proposition \ref{matrixnormal} (which is the condition related to the effect of the initial weight), when using $\hat{W} = \frac{1}{q}I_q$, the second set of conditions, (ii), is more relaxed. These conditions are implied by the following:
$$\begin{cases}
    M_q ~&\geq~~~ \sqrt{\frac{q}{K_2}} \cdot \frac{\sigma^2}{||\text{diag}(U)||_{\max}\cdot C_{\max}}\\
    C ~&\leq~~~ \frac{c_0}{8} \cdot \frac{C_{\min}}{C_{\max}}\cdot \frac{||\text{diag}(U)||_{\min}}{||\text{diag}(U)||_{\max}}.
\end{cases}$$
Assuming that $\frac{C_{\min}}{C_{\max}}\cdot \frac{||\text{diag}(U)||_{\min}}{||\text{diag}(U)||_{\max}}$ is a constant, the latter condition will hold as long as we take a large enough $c_0$ value. Recall this is an arbitrary constant $\geq 4$ that determines the tradeoff between separation and stability. As for the former condition, since $m_q^{(1)} + m_q^{(2)} + ... + m_q^{(K_2)}=q$, we have $\frac{q}{K_2} \leq M_q$. Thus, as long as $\sqrt{M_q}\geq \frac{\sigma^2}{||\text{diag}(U)||_{\max}\cdot C_{\max}}$, the first condition is satisfied. 
\end{proof}

\section{Additional Simulations}\label{app_sim}
\subsection{Main Simulation Results with Alternative Noise Settings} \label{app_sim_main}

\begin{figure}
\begin{center}
\includegraphics[width=0.8\linewidth]{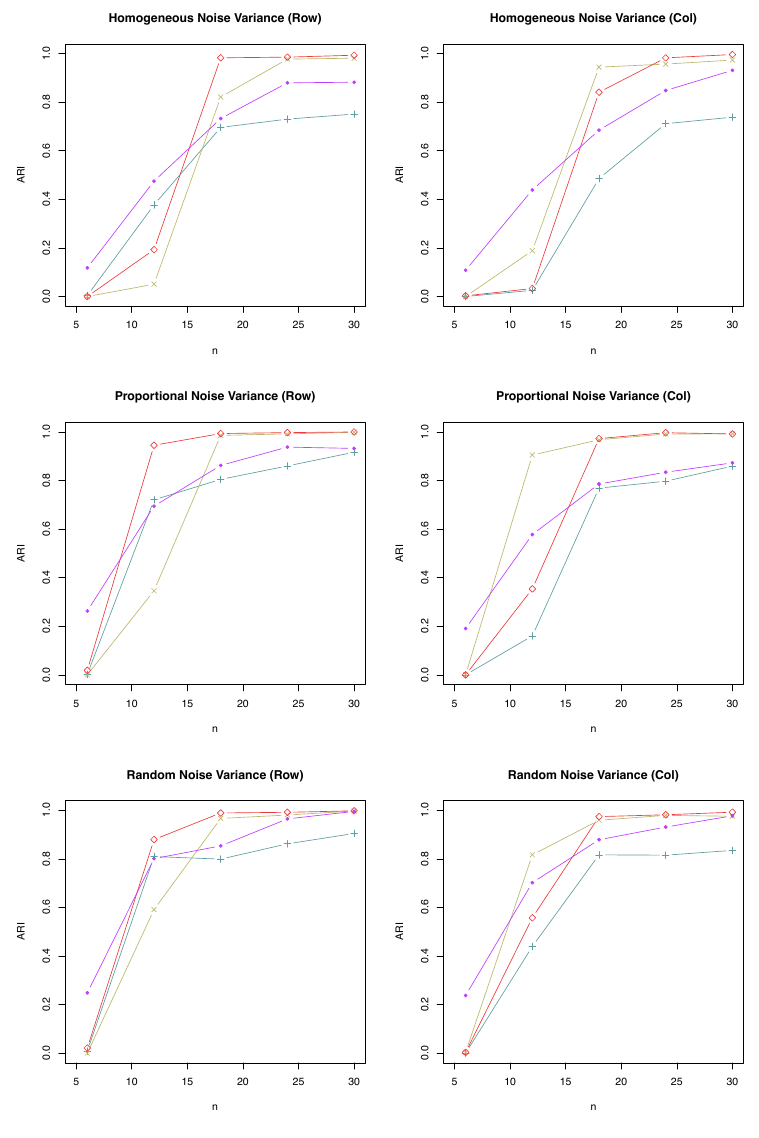}
\end{center}
\caption{A comparison of the adjusted Rand index (ARI) for row clustering (left) and column clustering (right). Our hierarchical clustering Algorithm \ref{alg_1} with $W_I$ (\textsc{naive cod}, $\color{teal}{-+-}$), \textsc{1-step cod} ($\color{olive}{-\times-}$), %with \ref{alg_2} 
    \textsc{2-step cod} ($\color{purple}{-\diamond-}$), %\ref{alg_3}
    and the competing method, \textsc{vanilla hierarchical} ($\color{violet}{-\bullet-}$).}
\label{fig_sim_total}
\end{figure}

The results for the entire main simulations including the homogeneous and random noise variance settings are shown in Figure \ref{fig_sim_total}. The implications from the different settings are very similar and will be omitted here for brevity. 

\subsection{Comparison with \textsc{DEEM}, \textsc{TGMM} and \text{TEEM}}\label{additional_simulations}

In this section, we consider for comparison the following clustering methods: (1) \texttt{DEEM}; (2) \texttt{TGMM}; (3) \texttt{TEMM}; (4) Algorithm \ref{alg_1} with the  initial weight $W_I$ ($\NAIVE$ $\COD$); (5) Algorithm \ref{alg_2} ($\ONESTEPCOD$); and (6) Algorithm \ref{alg_3} ($\TWOSTEPCOD$ in Section \ref{two-step} of the Supplementary Material). We note that the  data-driven cross-validation scheme is used to choose the tuning parameter in our $\COD$-based algorithms. The competing methods we used in our simulations came from recent model based tensor clustering papers, \cite{mai2022doubly} and \cite{deng2022tensor}. Their \texttt{DEEM} and \texttt{TGMM} methods utilize the tensor normal mixture model and their \texttt{TEMM} method utilizes the tensor envelope mixture model. Together with some variants of the EM algorithm, the authors exploited not only the mean information, but also the covariance structure in each direction of the tensor. These methods were implemented in our simulations using their designated functions within the R package \texttt{TensorClustering}.

Since the goal of \texttt{DEEM}, \texttt{TGMM} and \texttt{TEMM} is to cluster over $n$ i.i.d copies of tensors rather than the features, we need to adapt their methods for row and column clustering in our simulations. In particular, we implement row clustering by regarding the $p$ rows in our setting to be the direction of the observations in their tensor setting, and regard our $q \times n$ matrix slices to be the tensors in their tensor setting. Note that this violates their tensor normal distributional assumptions as the $p$ ``copies" of $q \times n$ matrix slices are no longer i.i.d. The column clustering is performed similarly. The simulation results for the random noise variance setting are shown in Figure \ref{fig_sim3}. The results for the other two noise variance settings can be found in Section \ref{app_sim_main} of the Supplementary Material.

It must be noted that since the distributional assumptions for the \texttt{DEEM}, \texttt{TGMM} and \texttt{TEMM} methods are violated, these methods fail to converge in some of the runs, especially when $n$ is large. One possible reason is that when $n$ is larger, the i.i.d. assumption is more severely violated and the computation becomes indeterminate in the process. In order to fairly compare our method with \texttt{DEEM}, \texttt{TGMM} and \texttt{TEMM}, we report not only the average ARI among the successful runs of their methods, but also the percentage of the successful runs out of the total number of runs. The latter is called the ``successful implementation rate".

%, sensitivity (true positive rate), and specificity (false positive rate) \citep{parikh2008understanding}. To compute sensitivity and specificity, we follow the pairwise approach in \cite{wiwie2015comparing}. 

\subsubsection{Balanced $p$ and $q$} \label{balanced}
We consider the following data generating process. We fix $p=q=100$, $K_1=K_2=10$ and a moderately unbalanced row and column cluster size structure (both having cluster sizes of $3$, $6$, $6$, $8$, $10$, $10$, $12$, $12$, $14$, $19$, respectively). This gives us the membership matrices $A$ and $B$. The latent variable $Z$ is generated from the matrix normal distribution $\mathrm{MN}(0_{K_1 \times K_2},U_{K_1 \times K_1},V_{K_2 \times K_2})$, where $U_{jk}=(-0.4)^{|j-k|}$ and $V_{jk}=0.3^{|j-k|}$. We further generate $\Gamma_{ij}\sim N(0,\sigma_{ij}^2)$, where we consider the following three settings for the noise variance: (1) homogeneous noise variances $\sigma_{ij}^2=15$; (2) heterogeneous noise variances proportional to the corresponding row and column cluster sizes: $\sigma_{ij}^2 = \frac{15pq\cdot v_{ij}}{\sum_{i,j}v_{ij}},  v_{ij}=\frac{m_p^{(i)}\cdot m_q^{(j)}}{\sqrt{\frac{pq}{K_1K_2}}}$; and (3) heterogeneous noise variances randomly generated from the Uniform distribution: $\sigma_{ij}^2 = \frac{15pq\cdot u_{ij}^{h}}{\sum_{i,j}u_{ij}^{h}}, u_{ij} \sim \mathrm{Unif}(0,1)$ where $h$ determines the level of heterogeneity. They will be referred to as the ``homogeneous", ``proportional" and ``random" cases, respectively.

The simulation results show that in all three noise variance settings the successful implementation rates of \texttt{DEEM}, \texttt{TGMM} and \texttt{TEMM} decrease drastically as $n$ increases. In contrast, our $\COD$ methods never fail to converge in any of the settings. When the competing methods converge, the ARI is decent, especially when $n \leq 12$ where they outperform the $\COD$ methods in the homogeneous and random noise variance settings. However, as $n$ grows, our $\COD$ methods become more reliable and more accurate.

In most cases, our iterative algorithms $\ONESTEPCOD$ and $\TWOSTEPCOD$ improve the performance of $\NAIVE$ $\COD$, which is consistent with our theoretical analysis. In particular, $\ONESTEPCOD$ and $\TWOSTEPCOD$ can achieve an ARI value close to 1 when $n=18$, whereas the $\NAIVE$ $\COD$ may require $n$ much larger than $30$ to attain the same level of accuracy. The phenomenon holds for both row and column clustering.  We also find that when $n$ is moderate (e.g., $n\geq 18$), $\ONESTEPCOD$ and $\TWOSTEPCOD$ have very similar performances. Due to the extra computational cost of $\TWOSTEPCOD$, we generally recommend $\ONESTEPCOD$ for practical use if $n$ is moderate or large. However if $n$ is small, $\TWOSTEPCOD$ may outperform $\ONESTEPCOD$. 

\begin{figure}
\begin{center}
\includegraphics[width=1\linewidth]{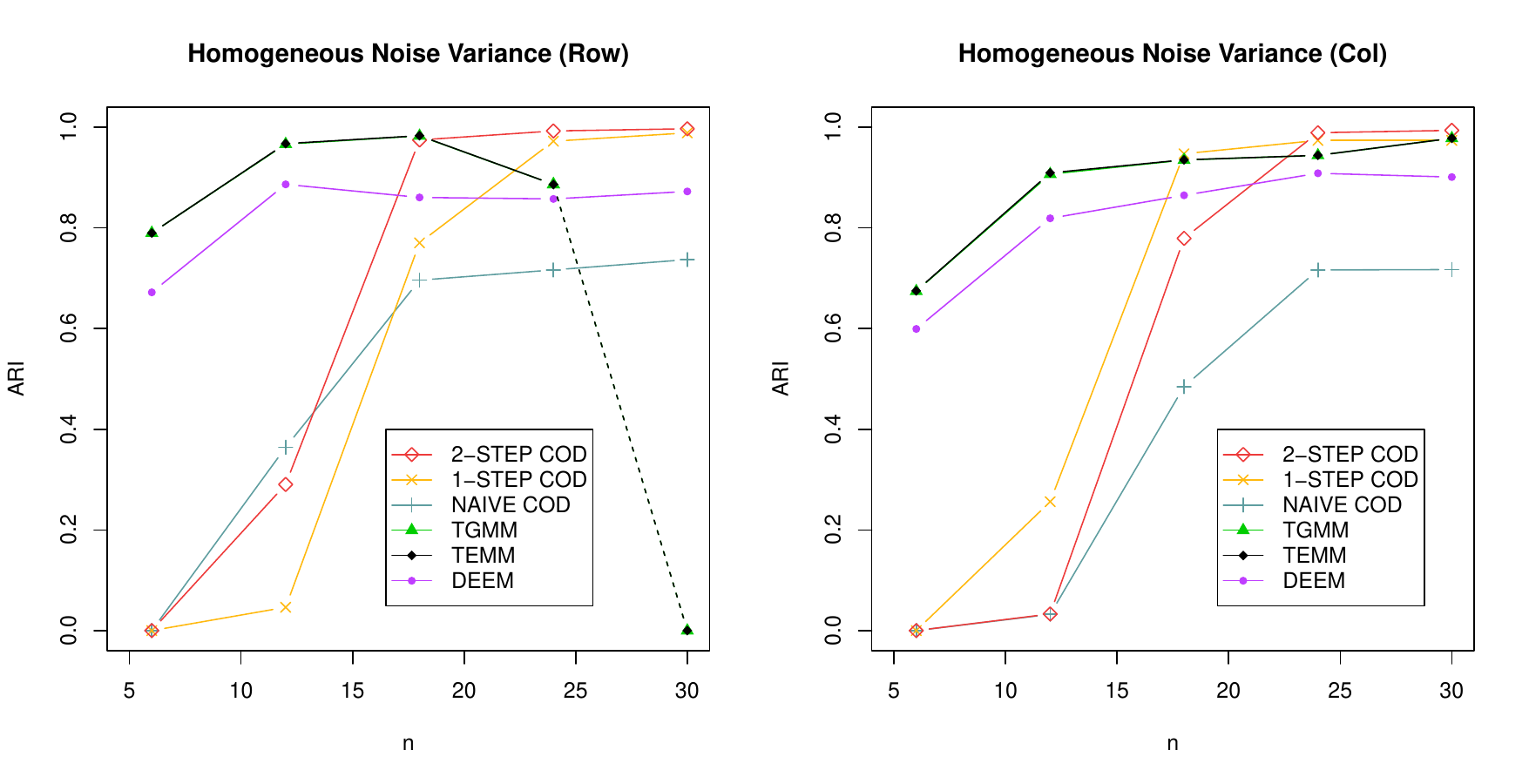}
\includegraphics[width=1\linewidth]{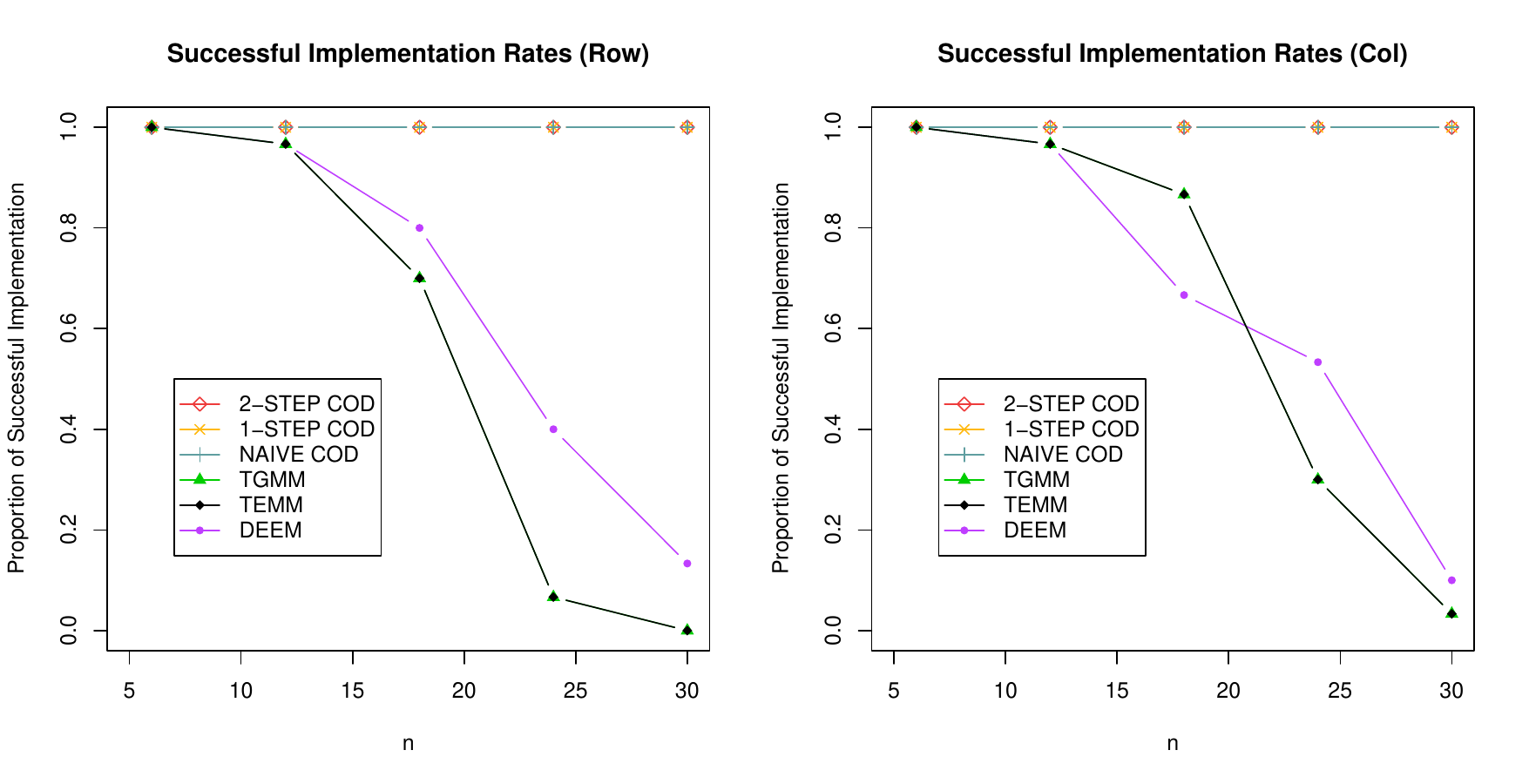}
\end{center}
\caption{The ARI and the successful implementation rates of our hierarchical clustering Algorithm \ref{alg_1} with $W_I$ ($\NAIVE$ $\COD$), the one-step Algorithm \ref{alg_2} ($\ONESTEPCOD$), the two-step Algorithm \ref{alg_3} ($\TWOSTEPCOD$), and competing model-based tensor clustering methods \texttt{DEEM} [\cite{mai2022doubly}], \texttt{TGMM} and \texttt{TEMM} [\cite{deng2022tensor}] under the homogeneous noise variance setting. Here, $p=q=100$, $K_1=K_2=10$, and we have moderately unbalanced cluster sizes of $(3, 6, 6, 8, 10, 10, 12, 12, 14, 19)$ for the rows and the columns each. Successful implementation rates are especially emphasized as the competing methods (\texttt{DEEM}, \texttt{TGMM}, \texttt{TEMM}) tended to fail during execution for some of the runs. The ARI rates for these methods were averaged over the successful runs, and when there were no successes, the ARI was reported as 0 with a dashed line.}
\label{fig_sim}
\end{figure}

\begin{figure}
\begin{center}
\includegraphics[width=1\linewidth]{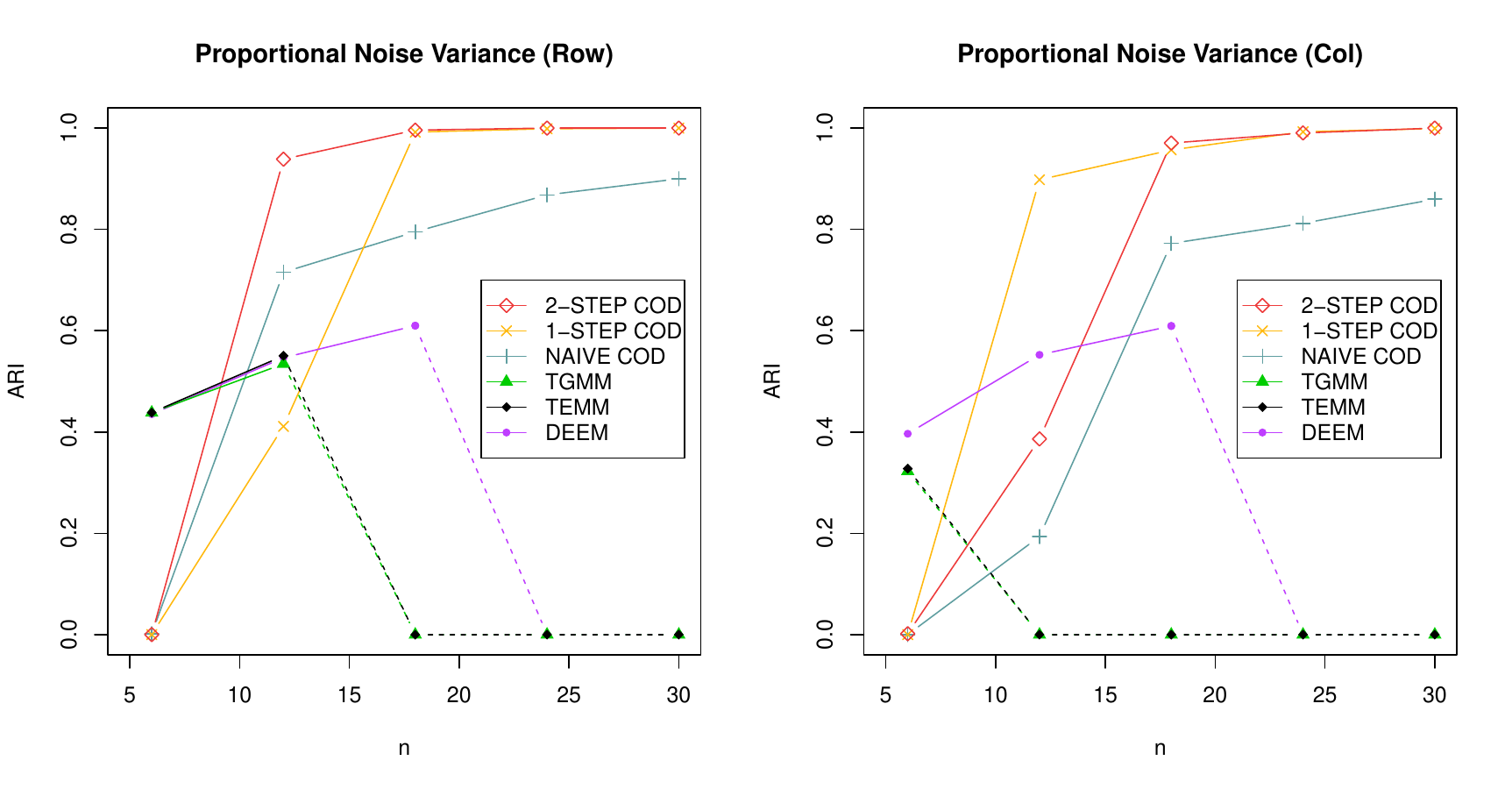}
\includegraphics[width=1\linewidth]{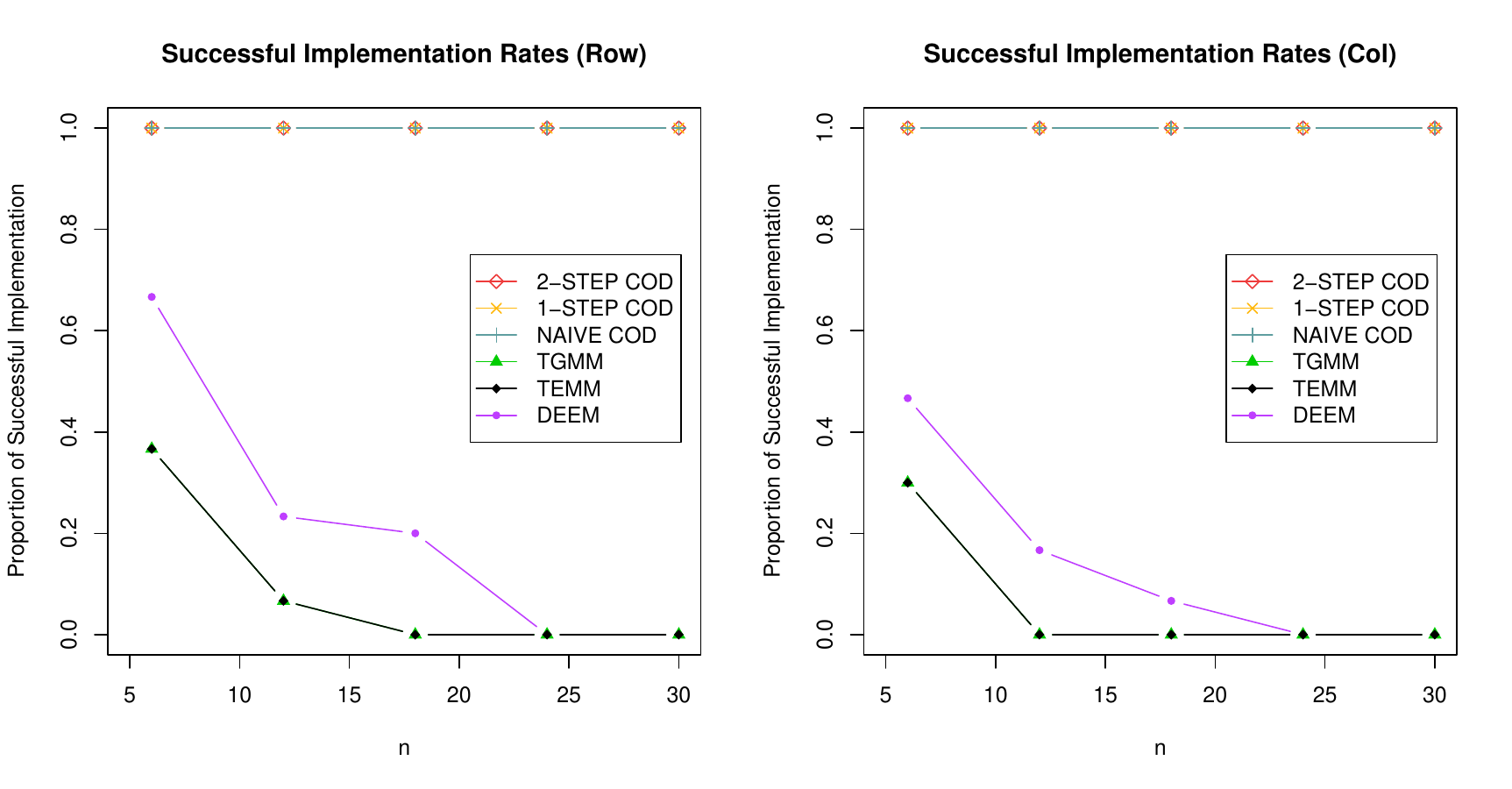}
\end{center}
\caption{The ARI and the successful implementation rates of our hierarchical clustering Algorithm \ref{alg_1} with $W_I$ ($\NAIVE$ $\COD$), the one-step Algorithm \ref{alg_2} ($\ONESTEPCOD$), the two-step Algorithm \ref{alg_3} ($\TWOSTEPCOD$), and competing model-based tensor clustering methods \texttt{DEEM} [\cite{mai2022doubly}], \texttt{TGMM} and \texttt{TEMM} [\cite{deng2022tensor}] under the proportional noise variance setting. The other settings are the same as in Figure \ref{fig_sim}.}
\label{fig_sim2}
\end{figure}

\begin{figure}
\begin{center}
\includegraphics[width=\linewidth]{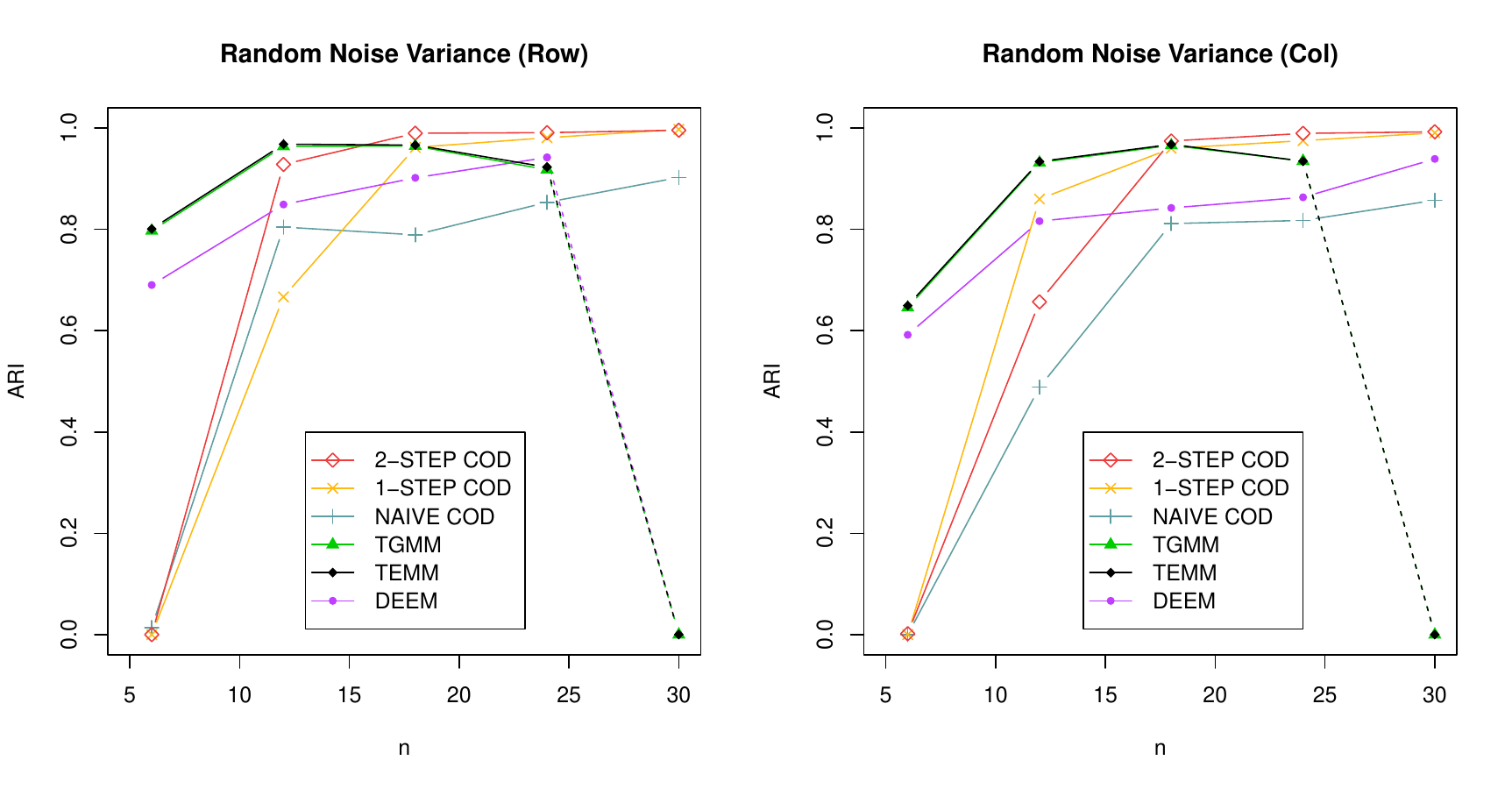}
\includegraphics[width=\linewidth]{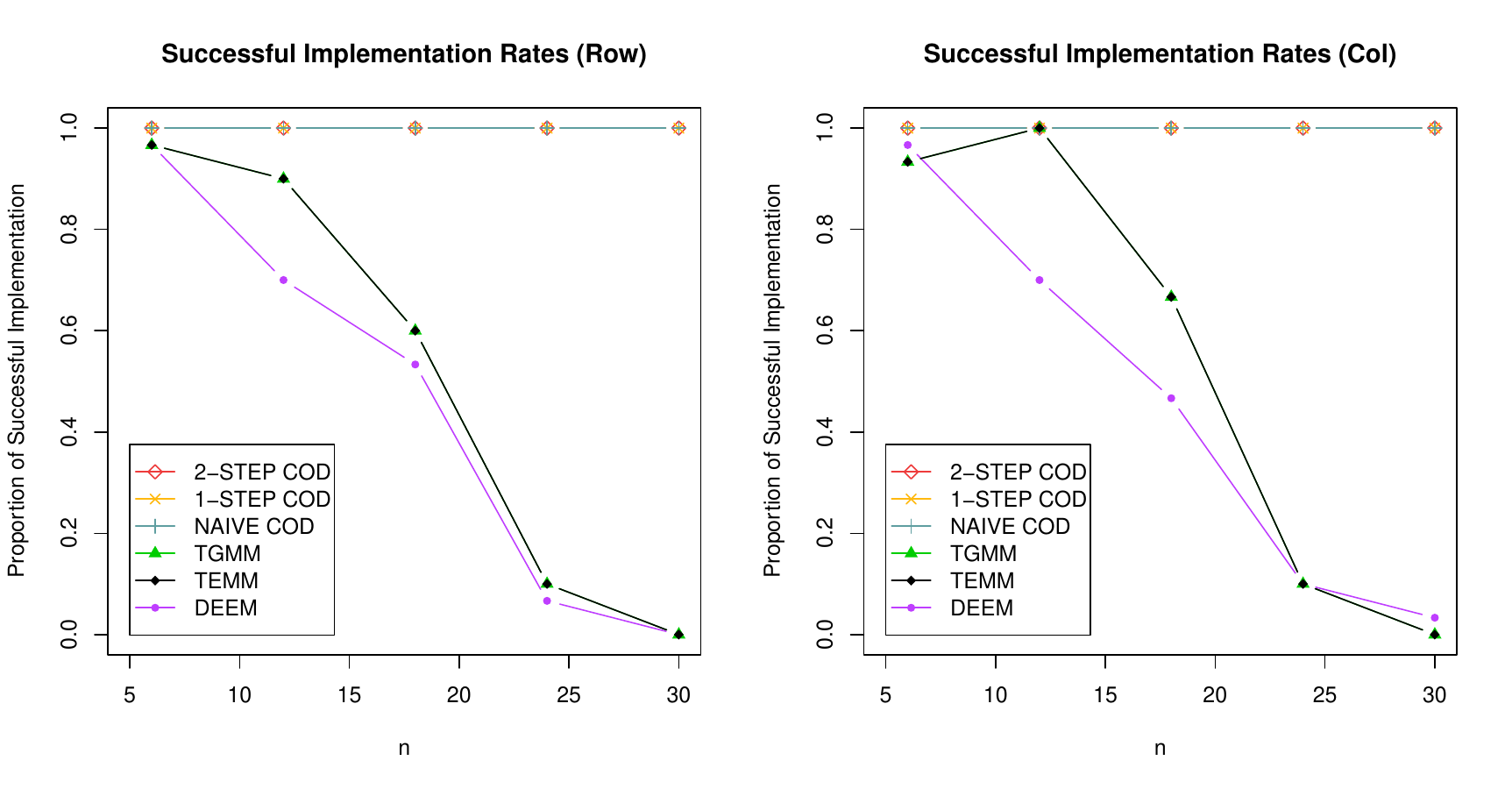}
\end{center}
\caption{The ARI and the successful implementation rates of our hierarchical clustering Algorithm \ref{alg_1} with $W_I$ ($\NAIVE$ $\COD$), the one-step Algorithm \ref{alg_2} ($\ONESTEPCOD$), the two-step Algorithm \ref{alg_3} ($\TWOSTEPCOD$), and competing model-based tensor clustering methods \texttt{DEEM} [\cite{mai2022doubly}], \texttt{TGMM} and \texttt{TEMM} [\cite{deng2022tensor}] under the random noise variance setting. The other settings are the same as in Figure \ref{fig_sim}.}
\label{fig_sim3}
\end{figure} 

\subsubsection{Unbalanced $p$ and $q$} \label{unbalanced}
\begin{figure}
\begin{center}
\includegraphics[width=\linewidth]{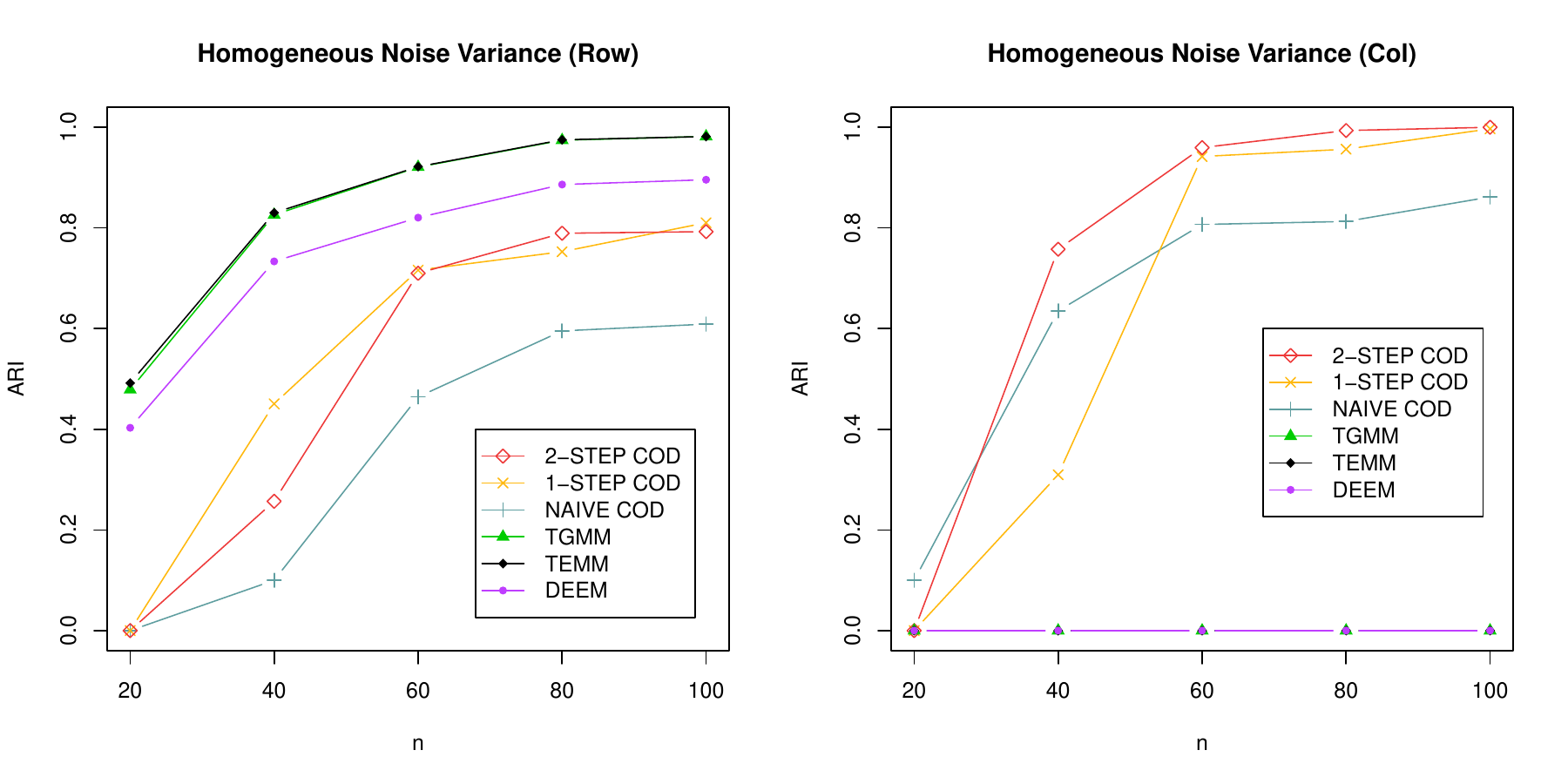}
\includegraphics[width=\linewidth]{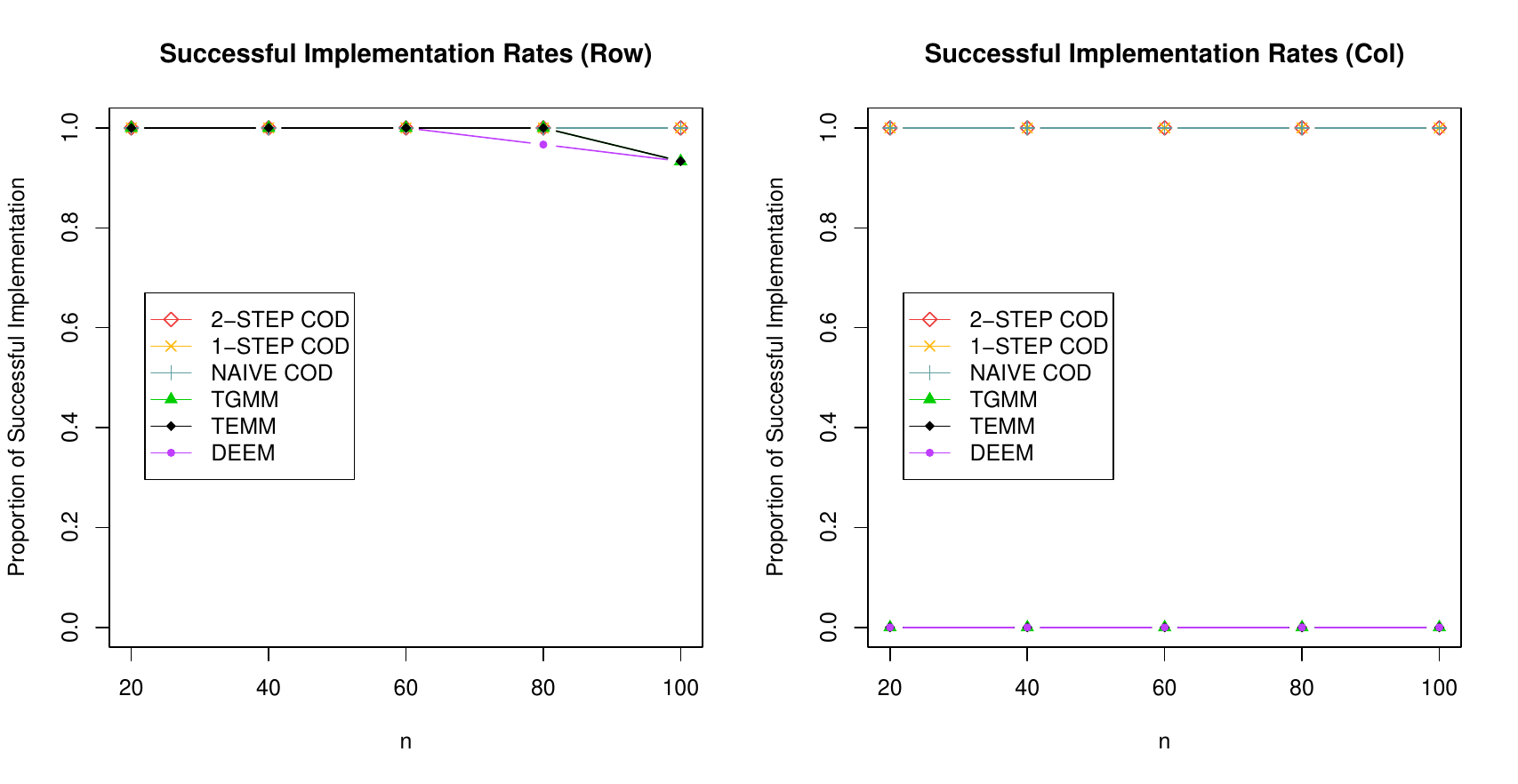}
\end{center}
\caption{The ARI and the successful implementation rates of our hierarchical clustering Algorithm \ref{alg_1} with $W_I$ ($\NAIVE$ $\COD$), the one-step Algorithm \ref{alg_2} ($\ONESTEPCOD$), the two-step Algorithm \ref{alg_3} ($\TWOSTEPCOD$), and competing model-based tensor clustering methods \texttt{DEEM} [\cite{mai2022doubly}], \texttt{TGMM} and \texttt{TEMM} [\cite{deng2022tensor}] under the homogeneous noise variance setting. Here, $p=100$, $q=15$, $K_1=10$, $K_2=5$, where the decreased $q$ is the notable difference with the main simulation results. We have moderately unbalanced cluster sizes of $(3,6,6,8,10,10,12,12,14,19)$ and $(2,2,2,3,6)$ for the rows and the columns, respectively. Successful implementation rates are especially emphasized as the competing methods (\texttt{DEEM}, \texttt{TGMM}, \texttt{TEMM}) tended to fail during execution for some of the runs. The ARI rates for these methods were averaged over the successful runs, and when there were no successes, the ARI was reported as 0 with a dashed line.}
\label{fig:extra2}
\end{figure}

\begin{figure}
\begin{center}
\includegraphics[width=\linewidth]{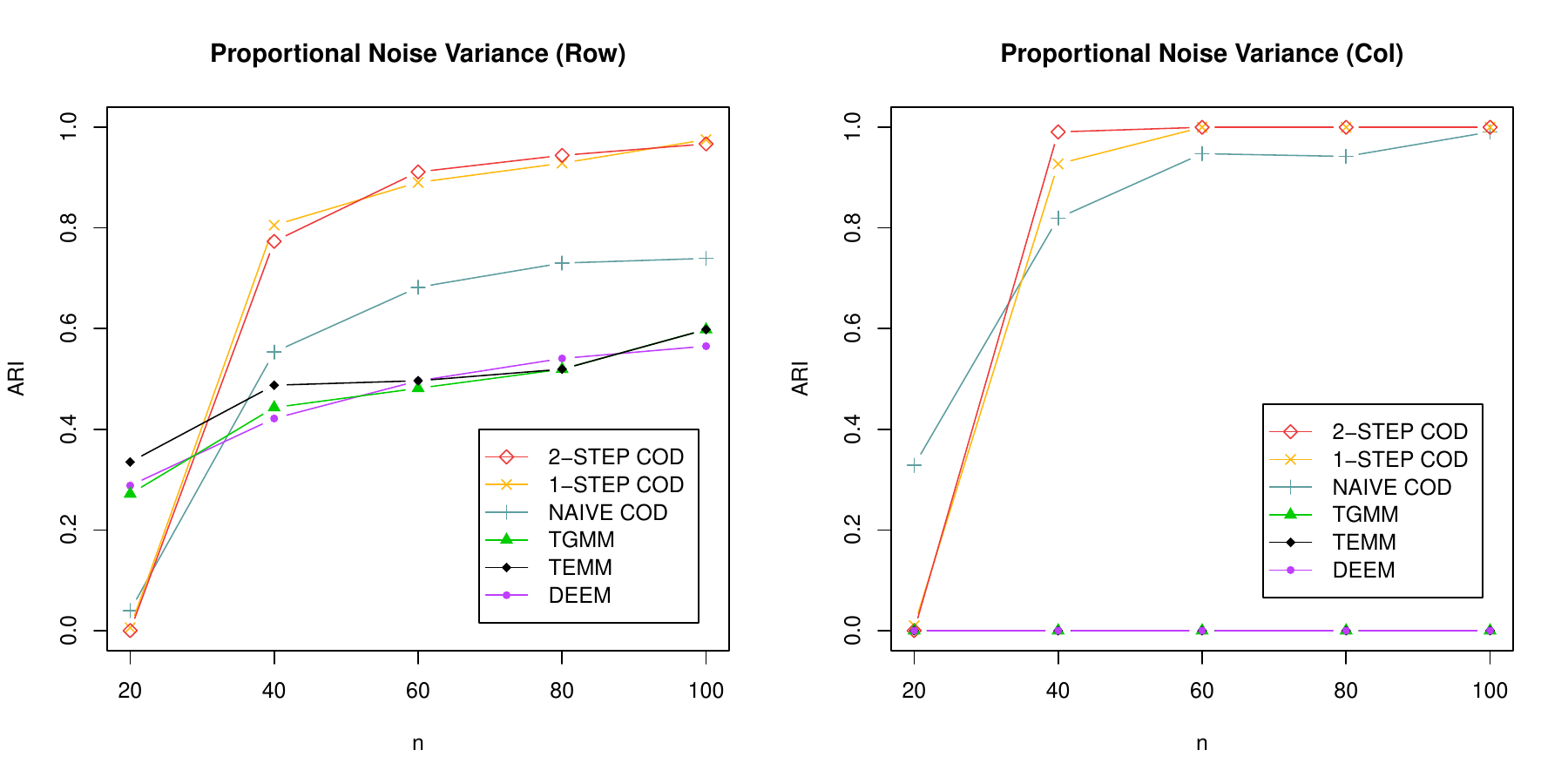}
\includegraphics[width=\linewidth]{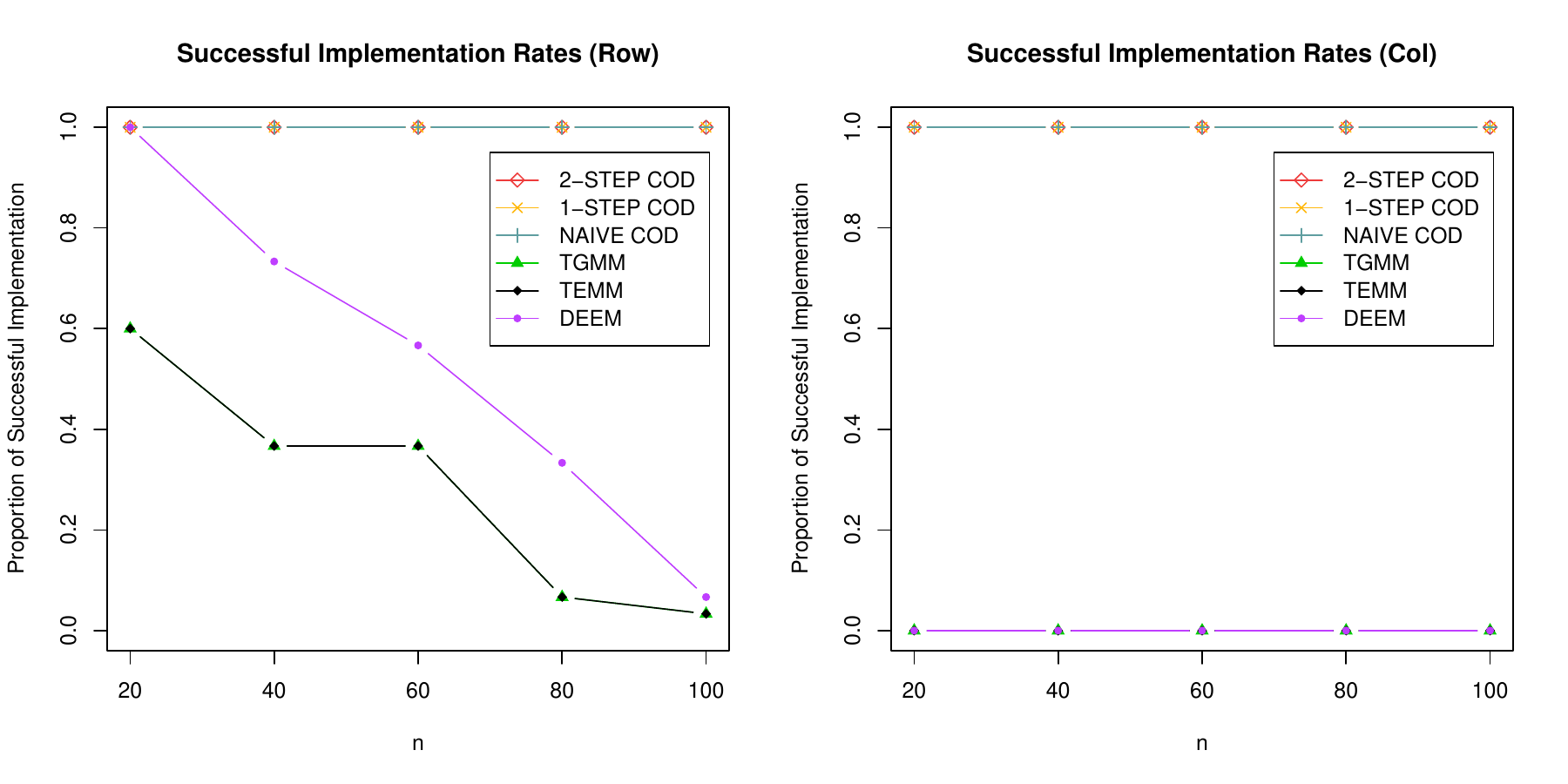}
\end{center}
\caption{The ARI and the successful implementation rates of our hierarchical clustering Algorithm \ref{alg_1} with $W_I$ ($\NAIVE$ $\COD$), the one-step Algorithm \ref{alg_2} ($\ONESTEPCOD$), the two-step Algorithm \ref{alg_3} ($\TWOSTEPCOD$), and competing model-based tensor clustering methods \texttt{DEEM} [\cite{mai2022doubly}], \texttt{TGMM} and \texttt{TEMM} [\cite{deng2022tensor}] under the proportional noise variance setting. The other settings are the same as in Figure \ref{fig:extra2}.}
\label{fig:extra2.1}
\end{figure}

\begin{figure}
\begin{center}
\includegraphics[width=\linewidth]{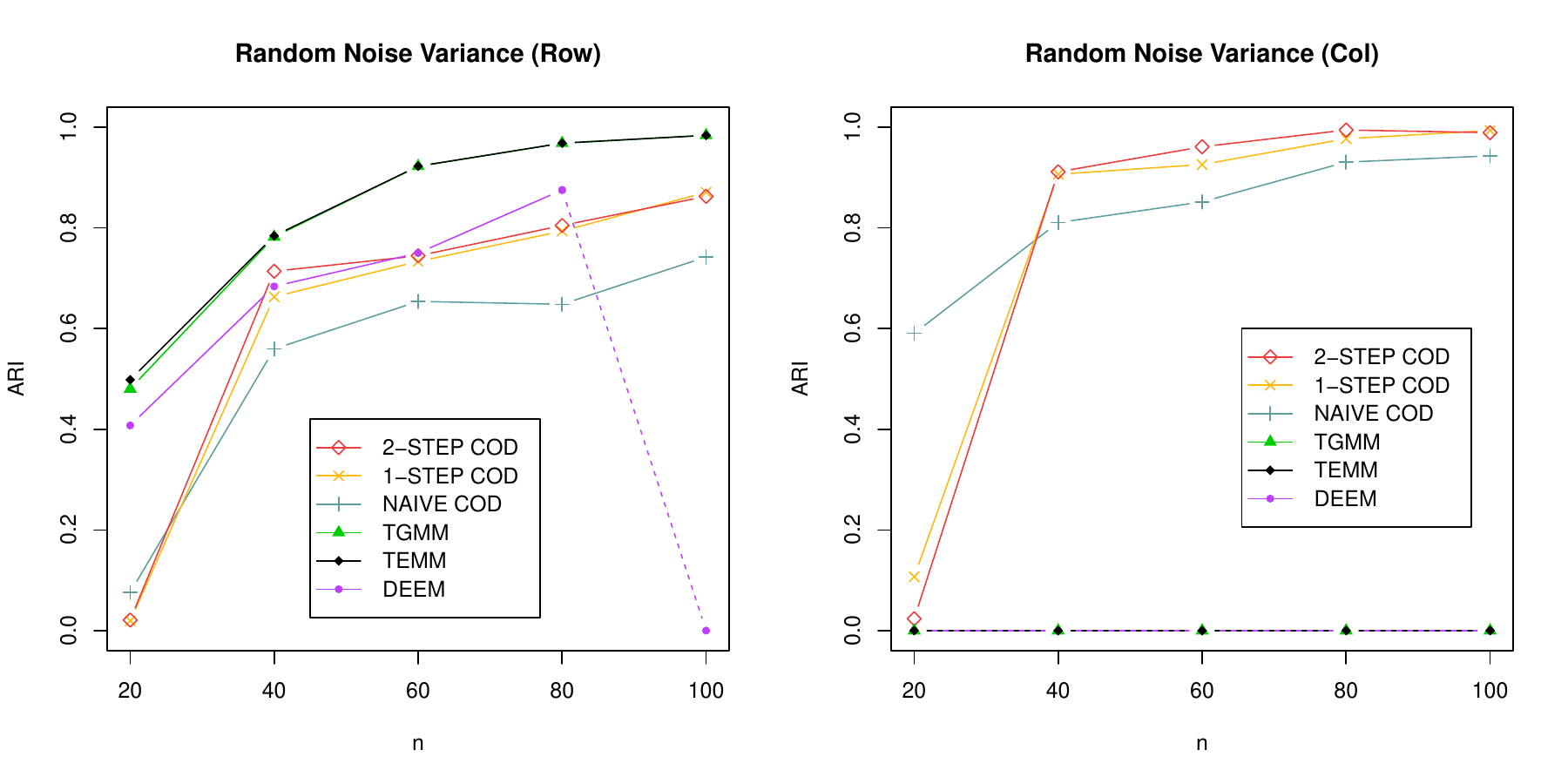}
\includegraphics[width=\linewidth]{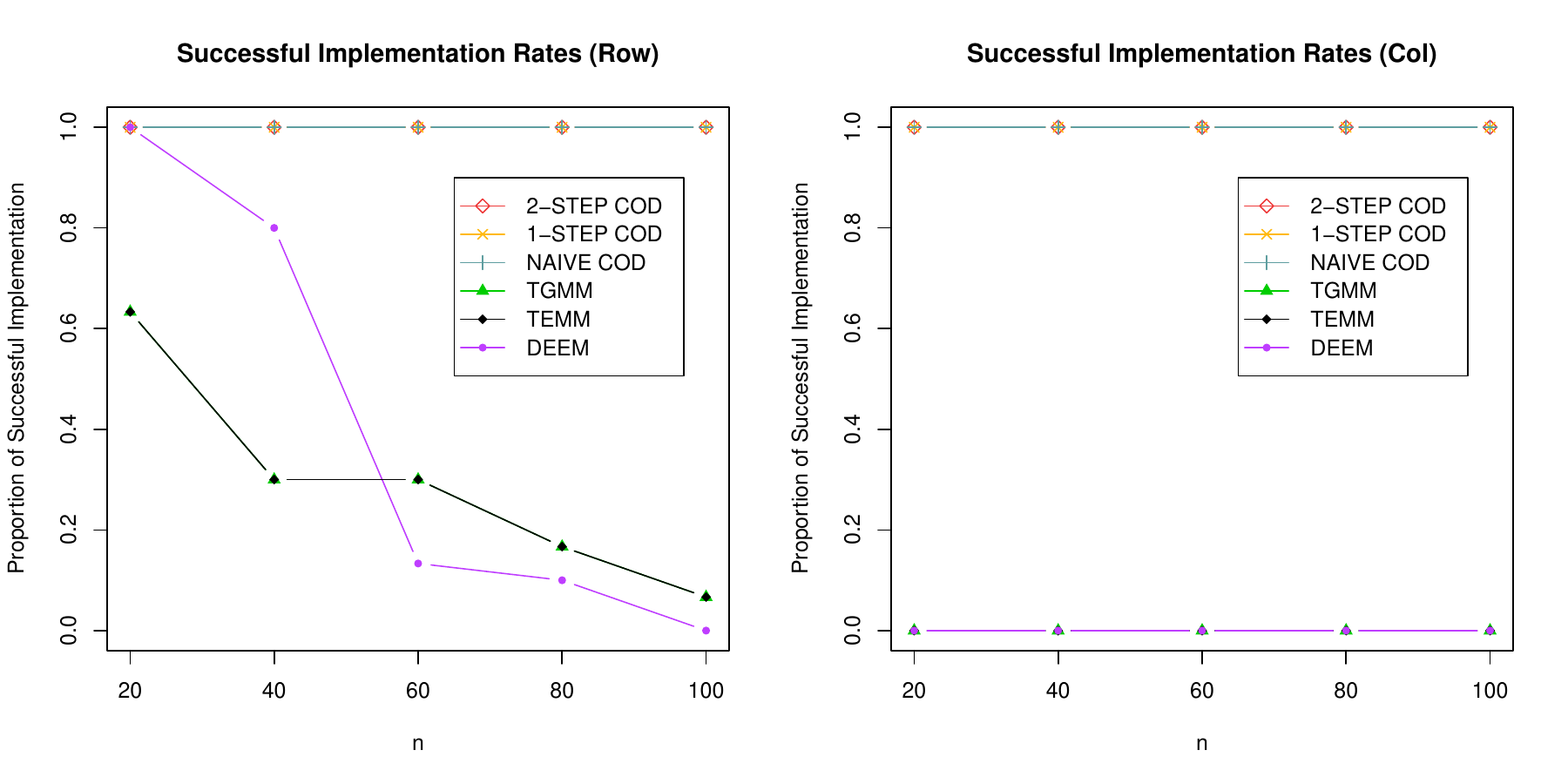}
\end{center}
\caption{The ARI and the successful implementation rates of our hierarchical clustering Algorithm \ref{alg_1} with $W_I$ ($\NAIVE$ $\COD$), the one-step Algorithm \ref{alg_2} ($\ONESTEPCOD$), the two-step Algorithm \ref{alg_3} ($\TWOSTEPCOD$), and competing model-based tensor clustering methods \texttt{DEEM} [\cite{mai2022doubly}], \texttt{TGMM} and \texttt{TEMM} [\cite{deng2022tensor}] under the random noise variance setting. The other settings are the same as in Figure \ref{fig:extra2}.}
\label{fig:extra2.3}
\end{figure}

Here we present the unbalanced setting of $p=100$, $q=15$, $K_1=10$, $K_2=5$ (with all of the other settings being  the same as in \ref{balanced}) shown in Figures \ref{fig:extra2}, \ref{fig:extra2.1} and \ref{fig:extra2.3}. In this case, we have long, skinny matrices that are ``unbalanced". We find that the $\COD$ methods perform drastically better in column clustering compared to the tensor model based methods. In all three noise variance settings, the competing methods have 0 successful implementation rates across all $n$ values in column clustering, whereas our $\COD$ methods perform significantly better. We believe this occurs because in the tensor normal setting, having too small of a $q$ value is equivalent to having too few ``i.i.d. observations", which causes their algorithms to diverge.  
%In any statistical setting, having too few observations causes the problem to be much more difficult. (Note that in row clustering, however, in some noise variance settings the competing methods perform better than the COD methods for the same reason - because $p$ is relatively large.) 
In contrast, for the $\COD$ algorithms, since the threshold for perfect column clustering recovery is on the order of $\sqrt{\frac{\log q}{nK_1}}$, if $q$ is relatively small and $K_1$ is relatively large, this threshold decreases compared to the balanced matrix setting and the condition for perfect column clustering recovery becomes more relaxed. %(Note that in row clustering, however, in some noise variance settings the competing methods perform better than the COD methods for the same reason - because the threshold order of $\sqrt{\frac{\log p}{nK_2}}$ is larger compared to the more balanced case. However, the lack in performance of COD in the row clustering case is very mild compared to the lack of performance of the competing methods in the column clustering case.) 
Thus, our $\COD$ based methods typically yield more accurate clustering results compared to the competing tensor model based methods, especially when the number of rows and columns are unbalanced.

\section{Real Data Analysis with a Competing Method} \label{real_data_deem}
\begin{figure}
\begin{center}
\includegraphics[width=0.8\linewidth]{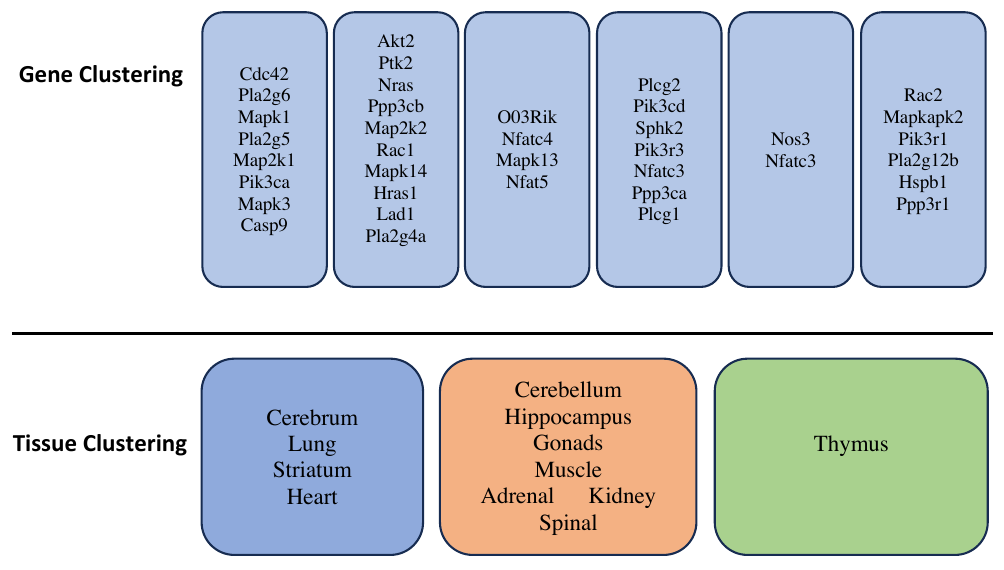}
\end{center}
\caption{Gene clusters and tissue clusters obtained from the competing method \texttt{DEEM} (\cite{deng2022tensor}).}
\label{real_data_analysis_DEEM}
\end{figure}

We add the real data analysis results from using the competing method \texttt{DEEM} (\cite{deng2022tensor}) on the same gene-tissue data mentioned in Section \ref{sec_real}. The results are illustrated in Figure \ref{real_data_analysis_DEEM}. Of the gene clustering results from our method shown in Figure \ref{fig:cod}, \texttt{DEEM} was also able to cluster ``O03Rik" and ``Nfat5" together and ``Rac2" and ``Mapkapk2" together. However, \texttt{DEEM} was not able to capture the relationship between ``Nfatc4" and ``Ppp3r1", which was identified with our method. The mechanism behind the relationship is that the ``Ppp3r1" and ``Nfatc4" genes are both related to calcineurin - one having to do with how it is generated, and one having to do with something it activates. Another possible explanation is given in \cite{heit2006calcineurin} which postulates that calcineurin/NFAT signaling is critical in $\beta$-cell growth in the pancreas. Either way, it signals that our method can cluster meaningful genes that are not able to be identified with other competing methods. For the tissues, the competing method \texttt{DEEM} was able to cluster the ``Lung" - ``Heart" pair of vascular tissues and the ``Adrenal" - ``Gonads" pair of steroid responsive tissues that did not appear in our results with $\COD$. However, it was not able to cluster the ``Cerebrum" - ``Cerebellum" - ``Hippocampus" trio of neural tissues, the steroid responsive ``Adrenal" - ``Thymus" pair, or the vascular ``Lung" - ``Kidney" pair that were clustered together with $\COD$. Thus, although the competing method gives reasonably meaningful results that are different from the results using $\COD$, it is apparent that our method can complement existing clustering methods and provide useful information that wasn't available.

\section{Cluster Evaluation Metrics}\label{metrics}
For this section, the true row cluster partition will be denoted simply as $\cG=\{G_1,...,G_{K_1}\},$\\
$~\textrm{where}~~ G_k=\{a: A_{ak}=1\}$, while the estimated row cluster partition will be denoted as $\hat{\cG}=\{\hat{G}_1,...,\hat{G}_{s}\}, ~\textrm{where}~~ \hat{G}_k=\{a: \hat{A}_{ak}=1\}$. 

\subsection{Sensitivity and Specificity}

Then, for any pair $1 \leq j < k \leq p$, we can define
\begin{align*}
    TP_{jk} &= \textbf{1}\{j,k \in G_{a} \text{ and } j,k \in \hat{G}_{b} \text{ for some } 1\leq a \leq K_1, 1\leq b \leq s\}\\
    TN_{jk} &= \textbf{1}\{j,k \not\in G_{a} \text{ and } j,k \not\in \hat{G}_{b} \text{ for some } 1\leq a \leq K_1, 1\leq b \leq s\}\\
    FP_{jk} &= \textbf{1}\{j,k \not\in G_{a} \text{ and } j,k \in \hat{G}_{b} \text{ for some } 1\leq a \leq K_1, 1\leq b \leq s\}\\
    FN_{jk} &= \textbf{1}\{j,k \in G_{a} \text{ and } j,k \not\in \hat{G}_{b} \text{ for some } 1\leq a \leq K_1, 1\leq b \leq s\}
\end{align*}
and define
\begin{align*}
    TP &= \sum_{1 \leq j < k \leq p} TP_{jk}, ~~~~~~ TN = \sum_{1 \leq j < k \leq p} TN_{jk}\\
    FP &= \sum_{1 \leq j < k \leq p} FP_{jk}, ~~~~~~ FN = \sum_{1 \leq j < k \leq p} FN_{jk} 
\end{align*}
Sensitivity (SN) and specificity (SP) can then be defined as follows:
\begin{align*}
    SN = \frac{TP}{TP + FN} ~~~~~~ SP = \frac{TN}{TN + FP}
\end{align*}

\subsection{Adjusted Rand Index (ARI)}
The following is a cross tabulation of two cluster partitions (the true partition and the estimated partition). 
\begin{center}
\begin{tabularx}{0.6\textwidth} { 
  | >{\centering\arraybackslash}X 
  | >{\centering\arraybackslash}X 
  | >{\centering\arraybackslash}X
  | >{\centering\arraybackslash}X
  | >{\centering\arraybackslash}X
  | >{\centering\arraybackslash}X |}
 \hline
  & $\hat{G}_1$ & $\hat{G}_2$ & ... & $\hat{G}_s$ & Sum \\
 \hline
 $G_1$  & $n_{11}$  & $n_{12}$ & ... & $n_{1s}$ & $|G_1|$  \\
 \hline
 $G_2$  & $n_{21}$  & $n_{22}$ & ... & $n_{2s}$ & $|G_2|$  \\
 \hline
 :  & :  & : & ... & : & :  \\
 \hline
 $G_{K_1}$  & $n_{K_1 1}$  & $n_{K_1 2}$ & ... & $n_{K_1 s}$ & $|G_{K_1}|$  \\
 \hline
 Sum  & $|\hat{G}_1|$  & $|\hat{G}_2|$ & ... & $|\hat{G}_s|$ & $p$  \\
\hline
\end{tabularx}
\end{center}
The ARI is then defined as follows:
\begin{align*}
    \text{ARI} = \frac{\sum_{i,j}\binom{n_{ij}}{2} -  \Big[\sum_{i}\binom{|G_i|}{2} \sum_{j}\binom{|\hat{G}_j|}{2}\Big]\Big/\binom{p}{2} }{\frac{1}{2}\Big[\sum_{i}\binom{|G_i|}{2} + \sum_{j}\binom{|\hat{G}_j|}{2}\Big] -  \Big[\sum_{i}\binom{|G_i|}{2} \sum_{j}\binom{|\hat{G}_j|}{2}\Big]\Big/\binom{p}{2}}
\end{align*}
Note that this is the corrected-for-chance version of the Rand index \citep{rand1971objective}. An ARI value of 1 implies a perfect match between the two cluster partitions.

\end{document}